**LPAR-05 Workshop:**

# Empirically Successful Automated Reasoning in Higher-Order Logic (ESHOL)

Christoph Benzmüller, John Harrison, and Carsten Schürmann
(eds.)

$$\exists p.p \supseteq A \cap B$$

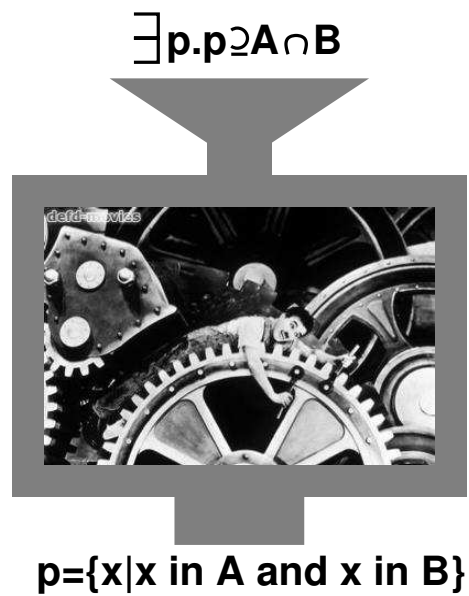

**p={x|x in A and x in B}**

## The ESHOL-05 Workshop

This workshop brings together practioners and researchers who are involved in the everyday aspects of logical systems based on higher-order logic. We hope to create a friendly and highly interactive setting for discussions around the following four topics. Implementation and development of proof assistants based on any notion of impredicativity, automated theorem proving tools for higher-order logic reasoning systems, logical framework technology for the representation of proofs in higher-order logic, formal digital libraries for storing, maintaining and querying databases of proofs.

We envision attendees that are interested in fostering the development and visibility of reasoning systems for higher-order logics. We are particularly interested in a discusssion on the development of a higher-order version of the TPTP and in comparisons of the practical strengths of automated higher-order reasoning systems. Additionally, the workshop includes system demonstrations.

ESHOL is the successor of the ESCAR and ESFOR workshops held at CADE 2005 and IJCAR 2004.

November, 2005

| | |
|---|---|
| Christoph Benzmüller | Saarland University, Germany |
| John Harrison | Intel Corporation, USA |
| Carsten Schürmann | IT University of Copenhagen, Denmark |

### Programme Committee

| | |
|---|---|
| Peter Andrews | Carnegie Mellon University, USA |
| Michael Beeson | San Jose State University, USA |
| Chad Brown | Saarland University, Germany |
| Gilles Dowek | École Polytechnique, France |
| Christoph Kreitz | Potsdam University, Germany |
| Larry Paulson | Cambridge University, UK |
| Frank Pfenning | Carnegie Mellon University, USA |
| Geoff Sutcliffe | University of Miami, USA |
| Volker Sorge | University of Birmingham, UK |
| Freek Wiedijk | Nijmegen University, Netherlands |

## Schedule (December 2nd, 2005)

09:00-10:00   Invited Talk

          **Joe Hurd (Oxford):** First Order Proof for Higher Order Logic Theorem Provers

10:00-10:30   Coffee Break

10:30-12:30   Paper Session I

          **Michael Beeson:** Implicit Typing in Lambda Logic

          **Christoph Benzmüller:** LEO – A Resolution based Higher Order Theorem Prover

          **Christoph Benzmüller, Volker Sorge, Mateja Jamnik and Manfred Kerber:** Combining Proofs of Higher-Order and First-Order Automated Theorem Provers

12:30-13:30   Lunch Break

13:30-14:30   Invited Talk

          **Chad Brown (Saarbrücken):** Benchmarks for Higher-Order Automated Reasoning

14:30-15:30   Paper Session II

          **Jutta Eusterbrock:** Co-Synthesis of New Complex Selection Algorithms and their Human Comprehensible XML Documentation

          **Alwen Tiu, Gopalan Nadathur and Dale Miller:** Mixing Finite Success and Finite Failure in an Automated Prover

15:30-16:00   Coffee Break

16:00-17:00   System Demonstrations

          **Michael Beeson:** Otter-$\lambda$

          **Chad Brown:** TPS

          **Joe Hurd:** Metis

          **Christoph Benzmüller:** LEO

17:00-18:00   Discussion: Higher-Order TPTP – Feasible or Not?
Chair and Panelists: TBA

# Table of Contents

# First Order Proof for Higher Order Logic Theorem Provers (abstract)

Joe Hurd[*]

Computing Laboratory
University of Oxford,
`joe.hurd@comlab.ox.ac.uk`

Interactive theorem provers are useful for modelling computer systems and then verifying properties of them by constructing a formal proof that the properties logically follow from the definition of the system. The expressivity of higher order logic makes it easy to model systems in a natural way, and there are many interactive theorem provers based on higher order logic, including HOL4 [4], Isabelle [12] and PVS [10]. In these theorem provers the system properties to be formally verified are statements of higher order logic, which are presented to the user as goals. The user proves goals by manually selecting tactics that reduce goals to simpler subgoals, until eventually the subgoals are simple enough that tactics can completely prove them. In general the initial goals corresponding to system properties require some higher order reasoning to prove them (typically an induction), but many subgoals require only first order reasoning and are efficiently proved by a standard first order calculus. Using first order provers to support interactive proof in higher order logic theorem provers has been a productive line of research, and the following is a chronological list of such combinations: FAUST in HOL [9]; SEDUCT in LAMBDA [3]; MESON in HOL [5]; 3TAP in KIV [1]; `blast` in Isabelle [11]; Gandalf in HOL [6]; and Bliksem in Coq [2].

There are two barriers to combining first order provers with interactive higher order theorem provers. The first is the incompatibility of the different logics: a method is required to convert a higher order logic goal to a set of first order clauses, and then to lift a refutation of the clauses to a higher order logic proof. Using the idea of an LCF kernel for first order refutations it is possible to make this logical interface into a module, allowing several different interfaces between first and higher order logic to co-exist [7]. The choice of interface to apply to a particular higher order logic goal depends on both the syntactic structure of the goal and which other interfaces have been tried.

The second barrier is an engineering one: the specifics of how to link up the first order prover and extract the information necessary to reconstruct the refutation and translate it to a higher order logic proof. The LCF kernel design of the logical interface makes it simple to convert refutations to a form in which they can be automatically translated to higher order logic proofs, and thus supports experimentation with a full range of first order calculi. Experiments have shown that resolution is more effective than model elimination for higher order logic

---

[*] Supported by a Junior Research Fellowship at Magdalen College, Oxford.

goals [8], and a calculus with specific rules for equality is also important for this application.

All the above ideas are implemented in the Metis proof tactic in the HOL4 theorem prover, which is separately presented as a system description.

# References

1. Wolfgang Ahrendt, Bernhard Beckert, Reiner Hähnle, Wolfram Menzel, Wolfgang Reif, Gerhard Schellhorn, and Peter H. Schmitt. Integration of automated and interactive theorem proving. In W. Bibel and P. Schmitt, editors, *Automated Deduction: A Basis for Applications*, volume II, chapter 4, pages 97–116. Kluwer, 1998.
2. Marc Bezem, Dimitri Hendriks, and Hans de Nivelle. Automated proof construction in type theory using resolution. In David A. McAllester, editor, *Proceedings of the 17th International Conference on Automated Deduction (CADE-17)*, volume 1831 of *Lecture Notes in Computer Science*, pages 148–163, Pittsburgh, PA, USA, June 2000. Springer.
3. H. Busch. First-order automation for higher-order-logic theorem proving. In Tom Melham and Juanito Camilleri, editors, *Higher Order Logic Theorem Proving and Its Applications, 7th International Workshop*, volume 859 of *Lecture Notes in Computer Science*, Valletta, Malta, September 1994. Springer.
4. M. J. C. Gordon and T. F. Melham, editors. *Introduction to HOL (A theorem-proving environment for higher order logic)*. Cambridge University Press, 1993.
5. John Harrison. Optimizing proof search in model elimination. In Michael A. McRobbie and John K. Slaney, editors, *13th International Conference on Automated Deduction (CADE-13)*, volume 1104 of *Lecture Notes in Artificial Intelligence*, pages 313–327, New Brunswick, NJ, USA, July 1996. Springer.
6. Joe Hurd. Integrating Gandalf and HOL. In Yves Bertot, Gilles Dowek, André Hirschowitz, Christine Paulin, and Laurent Théry, editors, *Theorem Proving in Higher Order Logics, 12th International Conference, TPHOLs '99*, volume 1690 of *Lecture Notes in Computer Science*, pages 311–321, Nice, France, September 1999. Springer.
7. Joe Hurd. An LCF-style interface between HOL and first-order logic. In Andrei Voronkov, editor, *Proceedings of the 18th International Conference on Automated Deduction (CADE-18)*, volume 2392 of *Lecture Notes in Artificial Intelligence*, pages 134–138, Copenhagen, Denmark, July 2002. Springer.
8. Joe Hurd. First-order proof tactics in higher-order logic theorem provers. In Myla Archer, Ben Di Vito, and César Muñoz, editors, *Design and Application of Strategies/Tactics in Higher Order Logics*, number NASA/CP-2003-212448 in NASA Technical Reports, pages 56–68, September 2003.
9. R. Kumar, T. Kropf, and K. Schneider. Integrating a first-order automatic prover in the HOL environment. In Myla Archer, Jeffrey J. Joyce, Karl N. Levitt, and Phillip J. Windley, editors, *Proceedings of the 1991 International Workshop on the HOL Theorem Proving System and its Applications (HOL '91), August 1991*, pages 170–176, Davis, CA, USA, 1992. IEEE Computer Society Press.
10. S. Owre, N. Shankar, J. M. Rushby, and D. W. J. Stringer-Calvert. *PVS System Guide*. Computer Science Laboratory, SRI International, Menlo Park, CA, September 1999.

11. L. C. Paulson. A generic tableau prover and its integration with Isabelle. *Journal of Universal Computer Science*, 5(3), March 1999.
12. Lawrence C. Paulson. Isabelle: A generic theorem prover. *Lecture Notes in Computer Science*, 828:xvii + 321, 1994.

# Implicit Typing in Lambda Logic

Michael Beeson[1]

San José State University, San José, Calif.
beeson@cs.sjsu.edu,
www.cs.sjsu.edu/faculty/beeson

**Abstract.** Otter-lambda is a theorem-prover based on an untyped logic with lambda calculus, called Lambda Logic. Otter-lambda is built on Otter, so it uses resolution proof search, supplemented by demodulation and paramodulation for equality reasoning, but it also uses a new algorithm, lambda unification, for instantiating variables for functions or predicates. The basic idea of a typed interpretation of a proof is to "type" the function and predicate symbols by specifying the legal types of their arguments and return values. The idea of "implicit typing" is that if the axioms can be typed in this way then the consequences should be typable too. This is not true in general if unrestricted lambda unification is allowed, but for a restricted form of "type-safe" lambda unification it is true. The main theorem of the paper shows that the ability to type proofs if the axioms can be typed works for the rules of inference used by Otter-lambda, if type-safe lambda unification is used, and if demodulation and paramodulation from or into variables are not allowed. All the interesting proofs obtained with Otter-lambda, except those explicitly involving untypable constructions such as fixed-points, are covered by this theorem.

## 1 Introduction: the no-nilpotents example

We begin with an example. Consider the problem of proving that there are no nilpotent elements in an integral domain. To explain the problem: an integral domain is a ring $R$ in which $xy = 0$ implies $x = 0$ or $y = 0$, i.e. there are no zero divisors. A element $c$ of $R$ is called *nilpotent* if for some positive integer $n$, $c^n$ (i.e., $c$ multiplied by itself $n$ times) is zero. Informally, one proves by induction on $n$ that $c^n$ is not zero. The equation defining exponentiation is $x^{s(n)} = x * x^n$. If $c$ and $c^n$ are both nonzero, then the integral domain axiom implies that $c^{n+1}$ is also nonzero. It is a very simple proof, but it is interesting because it involves two *types* of objects, ring elements and natural numbers, and the proof involves a mix of the algebraic axioms and the number-theoretical axioms (mathematical induction). Since the proof is so simple, we can consider the issues raised by having two types of objects without being distracted by a complicated proof.

How are we to formalize this theorem in first order logic? The traditional way would be to have two unary predicates $R(x)$ and $N(x)$, whose meaning would be "$x$ is a member of the ring $R$" and "$x$ is a natural number", respectively. Then the ring axioms would be "relativized to $R$", which means that instead of saying $x + 0 = 0$, we would say $R(x) \rightarrow x + 0 = 0$, or in clausal form, $-R(x)|x + 0 = 0$. (The vertical bar means "or", and the minus sign means "not".) Similarly, the axiom of induction would be relativized to $N$. The axiom of induction is usually formulated using a symbol $s$ for the successor function, or "next-integer" function. For example, $s(4) = 5$. The specific instance of induction we need for this proof can be expressed by the two (unrelativized) clauses

$$x^o \neq 0 \mid x^{g(x)} = 0 \mid x^n = 0.$$
$$x^o \neq 0 \mid x^{s(g(x))} \neq 0 \mid x^n = 0.$$

To see that this corresponds to induction, think of $g(x)$ as a constant (on which x is not allowed to depend). Then the middle literal of the first clause is $x^c = 0$. That is the induction hypothesis. The middle literal of the second clause is $x^{s(c)} \neq 0$. That is the negated conclusion of the induction step. We have used $o$ instead of 0 for the natural number zero, which might not be the same as the ring element 0.

A traditional course in logic would teach you that to formalize this problem, you need to relativize all the axioms using $R$ and $N$. Just to be explicit, the relativized versions of the induction axioms would be

$$-R(x) \mid -N(n) \mid x^o \neq 0 \mid x^{g(x,n)} = 0 \mid x^n = 0.$$
$$-R(x) \mid -N(n) \mid x^o \neq 0 \mid x^{s(g(x,n))} \neq 0 \mid x^n = 0.$$
$$-R(x) \mid -N(n) \mid N(g(n,x)).$$

and we would need additional axioms such as these:

$$-R(x) \mid -N(n) \mid R(x^n).$$
$$-R(x) \mid -R(y) \mid R(x + y).$$
$$-R(x) \mid -R(y) \mid R(x * y).$$
$$-R(x) \mid x + 0 = 0.$$

and so on for the other ring axioms.

## 2 Implicit typing in first order logic

Now here is the question: when formalizing this problem, do we need to relativize the induction axioms and the ring axioms using $R(x)$ and $N(x)$, or not? Experimentally, if we put the unrelativized axioms into Otter (Otter-$\lambda$ is not needed, since we have explicitly given the prover the required instance of induction), we do find a proof. What does this proof actually prove? Certainly it shows that in any integral domain whose underlying set is the natural numbers, there are no nilpotents, since in that case all the variables range over the same set, and no question of typing arises. We can prove informally that any countable integral domain is isomorphic to one whose underlying set is the natural numbers. But this is not the theorem that we set out to prove, so it may appear that we must use $R(x)$, $N(x)$, and relativization to formalize this problem.

That is, however, not so. The method of "implicit typing" shows that under certain circumstances we can dispense with unary predicates such as $R$ and $N$. One assigns a type to each predicate, function symbol, and constant symbol, telling what the sort of each argument is, and the sort of the value (in case of a function; predicates have Boolean value). Specifically each argument position of each function or predicate symbol is assigned a sort and the symbol is also assigned a "value type" or "return type". For example, in this problem the ring operations $+$ and $*$ have the type of functions taking two $R$ arguments and producing an $R$ value, which we might express as $type(R, +(R, R))$. If we use $N$ for the sort of natural numbers then we need to use a different symbol for addition on natural numbers, say $type(N, \mathrm{plus}(N, N))$, and we need to use a different symbol for 0 in the ring and zero in $N$. The Skolem symbol $g$ in the induction axiom has the type specification $type(N, g(R))$. The exponentiation function has the type specification $type(R, R^N))$.

Constants are considered as 0-ary function symbols, so they get assigned types, for example $type(R, 0)$ and $type(N, o)$. We call a formula or term *correctly typed* if it is built up consistently with these type assignments. Note that variables are not typed; e.g. $x + y$ is correctly typed no matter what variables $x$ and $y$ are. Types as we discuss them here are not quite the same as types in most programming languages, where variables are declared to have a certain type. Here, when a variable occurs in a formula, it inherits a type from the term in which it occurs, and if it occurs again in the same clause, it must have the same type at the other occcurence for the clause to be considered correctly typed. Once all the function symbols, constants, and predicate symbols have been assigned types, one can check (manually) whether the clauses supplied in an input file are correctly typed.

Then one observes that if the rules of inference preserve the typing, and if the axioms are correctly typed, and the prover finds a proof, then every step of the proof can be correctly typed. That means that it could be converted into a proof that used unary predicates for the sorts. Hence, if it assists the proof-finding process to omit these unary predicates, it is all right to do so. This technique was introduced long ago in [4], but McCune says it was already folklore at that time. It implies that the proof Otter finds using an input file without relativization actually is a valid proof of the theorem, rather than just of the special case where the ring elements are the natural numbers.

"Implicit typing" is the name of this technique, in which unary predicates whose function would be to establish typing are omitted. There are two ways to use implicit typing. First, we could just omit the unary predicates, let a theorem-proving program find a proof, and afterwards verify by hand (or by a computer program) that the proof is indeed well-typed. Second, we could verify that the axioms are well-typed, and prove that the inference rules used in the prover lead from correctly typed clauses to correctly typed clauses. Let us explore this second alternative. In order to state and prove a theorem, we first give some definitions:

**Definition 1.** *A type specification is an expression of the form* $type(R, f(U, V))$*, where* $R, U,$ *and* $V$ *are "type symbols". Any first-order terms not containing variables may be used as type symbols. Here 'type' must occur literally, and* $f$ *can be any symbol. The number of arguments of* $f$*, here shown as two, can be any number, including zero.*

The type $R$ is called the *value type* of $f$. The symbol $f$ is called the symbol of the type specification, and the number of arguments of $f$ is the *arity*.

**Definition 2.** *A typing of a term is an assignment of types to the variables occurring in the term and to each subterm of the term. A typing of a literal is similar, but the formula itself must get value type* bool*. A typing of a clause is an assignment that simultaneously types all the literals of the clause. A typing of a term (or literal or clause or set of clauses)* $t$ *is* correct with respect to a list of type specifications $S$ *provided that*

(i) *each occurrence of a variable in* $t$ *is assigned the same type.*

(ii) *each subterm* $r$ *of* $t$ *is typed according to a type specification in* $S$*. That is, if* $r$ *is* $f(u, v)$ *and* $f(u, v)$,$u$*, and* $v$ *are assigned types* $a$*,* $b$*, and* $c$ *respectively, then there is a type specification in* $S$ *of the form* $type(a, f(b, c))$*.*

(iii) *each occurrence of each subterm* $r$ *of* $t$ *in* $t$ *has the same value type.*

In the definition, nothing prevents $S$ from having more than one type specification for the same function symbol and arity. Condition (iii) is needed in such a case.

The phrase, *correctly typed term* $t$, is short for "term $t$ and a correct typing of $t$ with respect to some list of type specifications given by the context".

*Remark.* We do not allow type specifications to contain variables, but of course at the meta-level we can refer to a "typing of the form $i(U, U)$." That covers any specific typing such as $i(N, N)$, etc. For first-order theories, usually constant terms will suffice for naming the types (which are then usually called *sorts* rather than types, as in "multi-sorted logic").

The simplest theorem on implicit typing concerns the inference rule of (binary) resolution.[1]

**Theorem 1.** *Suppose each function symbol and constant occurring in a theory* $T$ *is assigned a unique type specification, in such a way that all the axioms of* $T$ *are correctly typed (with respect to this list of type specifications). Then conclusions reached from* $T$ *by binary resolution (using first-order unification) are also correctly typed.*

*Remark.* This theorem is perhaps implicit in [4]. We give it here mainly to prepare the way for extensions to lambda logic in the next section.

*Proof.* Suppose that literal $P(r)$ resolves with literal $-P(t)$, where $r$ and $t$ are terms; then there is a substitution $\sigma$ such that $r\sigma = t\sigma$, the unifying substitution.

---

[1] In the following theorem, we assume (as is customary with resolution) that after a theory has been brought to clausal form, the variables in distinct clauses are renamed so that no variable occurs in more than one clause.

Here $P$ stands for any atomic formula and $t$ and $r$ might stand for several terms if $P$ has more than one argument position. Since $P(r)$ and $P(t)$ are correctly typed by hypothesis, $r$ and $t$ must have the same value type (if they are not variables). The result of the resolution will be a disjunction of literals $Q\sigma|S\sigma$, where $Q$ and $S$ are the remaining (unresolved) literals in the clauses that originally contained $P(r)$ and $-P(t)$, respectively. Now $Q$ and $S$ are correctly typed by hypothesis, so we just need to show that applying the substitution $\sigma$ to a correctly typed term or literal will produce a correctly typed term or literal. This will be true by induction on the complexity of terms, provided that substitution $\sigma$ assigns to each variable $x$ in its domain, a term $q$ whose value type is the same as the value type of $x$ in the clause in which $x$ occurs. In first-order unification (but not in lambda unification) variables get assigned a value in unification only when the variable occurs as an argument, either of a parent term or a parent literal. That is, a variable cannot occur in the position of a literal. Thus when we are unifying $f(x, u)$ and $f(q, v)$, $x$ will get assigned to $q$, and the type of $x$ and the value type of $q$ must be the same since they are both in the first argument place of $f$. That completes the proof.

Does this theorem apply to the no-nilpotents example? We have to be careful about the type specification of the equality symbol. If we specify $type(\text{bool}, = (R, R))$, then we cannot use the same equality symbol in the axioms for the natural numbers, for example $s(x) \neq 0$ and $x = y|s(x) \neq s(y)$. However, Otter treats any symbol beginning with EQ as an equality; $=$ is a synonym for EQ, but one can also use, for example EQ2. Therefore, if we want to apply the theorem, we need to use two different equality symbols. Of course, we could just use $=$ throughout and verify afterwards that the proof can be correctly typed, as $=$ is never used in the same clause for equality between natural numbers and equality between ring elements; but if we want to be assured in advance that any proof Otter will find will be correctly typable, then we need to use different equality symbols. If we do so, then the theorem does apply.

There are, of course, more inference rules than just binary resolution. Even in this example, the proof uses demodulation. The theorem above can be extended to include the additional rules of inference factoring, paramodulation, and demodulation. For those not familiar with those rules we review their definitions. *Factoring* permits the derivation of a new clause by unifying two literals in the same clause that have the same sign, and applying the resulting substitution to the entire clause. *Paramodulation* is the following: suppose we have already deduced $t = q$ (or $q = t$) and $P[z := r]$, and unification of $t$ and $r$ produces a substitution $\sigma$ such that $t\sigma = r\sigma$; then we can deduce $P[z := q\sigma]$. Paramodulation *from variables* is the case in which $t$ is a variable. Paramodulation *into a variable* is the case in which $r$ is a variable. Demodulation is similar to paramodulation, except that (i) unlike paramodulation, it is unidirectional (i.e., the hypothesis must be $t = q$, not $q = t$), (ii) it is applied only under certain circumstances and using formulas designated in an input file as "demodulators". From the point of view of soundness proofs, it is a special case of paramodulation.

**Theorem 2.** *Suppose each function symbol and constant occurring in a theory $T$ is assigned a unique type specification, in such a way that all the axioms of $T$ are correctly typed (with respect to this list of type specifications). The type specifications of equality symbols must have the form type(bool, = $(X, X)$) for some type $X$. Then conclusions reached from $T$ by binary resolution, hyperresolution, factoring, demodulation, and paramodulation (using first-order unification in applying these rules) are also correctly typed, provided demodulation and paramodulation are not applied to or from variables.*

*Proof.* Conclusions reached by hyperresolution can also be reached by binary resolution, so that part of the theorem follows from the previous theorem. The results on factoring, paramodulation and demodulation follow from the fact that applying a substitution produced by unification preserves correct typings. The lemma that we need is that if $p$ and $r$ unify, then they have the same value type. If neither is a variable, this follows from the assumption that the axioms of $T$ are correctly typed. (If one is a variable, this need not be the case.)

Suppose, for example, that $r = s$ is to be used as a demodulator on term $t$. The demodulator is applied by unifying $r$ with a certain subterm $p$ of $t$. Let $\sigma$ be the substitution that performs this unification, so $p\sigma = r\sigma$. Then $p$ and $r$, since they unify, have the same value type, and hence $p$, $p\sigma$, and $r\sigma$ all have the same value type. The type specification of equality must have the form $type(bool, = (X, X))$ for some type $X$; so $r$ and $s$ have the same value type, so $r\sigma$ and $s\sigma$ have the same value type. Hence $s\sigma$ and $p\sigma$ also have the same value type, and hence the result of replacing $p$ in $t$ by $s\sigma$ (the result of the demodulation) is a correctly typed term.

*Example.* This example will show that one cannot allow "overloading", or multiple type specifications for the same symbol, and still use implicit typing with guaranteed correctness. For example, suppose we want to use $x + y$ both for natural numbers and for integers. Thinking of integers, we write the axiom $x + (-x) = 0$, and thinking of natural numbers we write $1 + x \neq 0$, Resolving these clauses, we find a contradiction upon taking $x = 1$.

*Example.* This example, taken from Euclidean geometry, shows that the theorem cannot be extended to paramodulation from variables. In this example, $EQpt$ stands for equality between points, $EQline$ stands for equality between lines, $I(a, b)$ stands for point $a$ incident to line $b$, and $p_1(u)$ and $p_2(u)$ are two distinct points on line $u$. The types here are boolean, point, and line. Axioms (1) and (2) are correctly typed:

$$EQpt(x, y) | I(x, line(x, y)). \qquad (1)$$

$$EQline(line(p1(u), p2(u)), u). \qquad (2)$$

Paramodulating from the first clause of (1) into (2), we unify $x$ with $line(p_1(u), p_2(u))$, and thus derive

$$EQline(y, u) | I(line(p1(u), p2(u)), line(line(p1(u), p2(u)), y)). \qquad (3)$$

This conclusion is incorrectly typed since $y$ is a point and $u$ is a line.

*Example.* This simpler example illuminates the situation with regard to paramodulation from variables. Consider the three unit clauses $x = a$, $P(b)$, and $-P(c)$. These clauses lead to a contradiction using paramodulation from the variable $x$ and binary resolution. But without paramodulation from variables, no contradiction can be derived. This shows that we have lost first-order refutation completeness, already in the first order case, as the price of implicit typing. But this is good: if equality is between objects of type $A$ and $P$ is a predicate on objects of type $B$, then these clauses are not contradictory. This loss of first-order completeness already occurs in the first-order case, and is not a phenomenon special to lambda logic. *Question*: "but if $b$ and $c$ have the same type, then shouldn't the contradiction be found?" *Answer*: '$b$' and '$c$' are constants in an untyped language, so they do not have types. Contradictions, like all proofs, are syntactic and involve the symbols. What the example shows is that, if many-sorted *models* are considered, there are models of this theory, even though the theory has no first-order models; and the theorem shows that the inference rules in question are sound for multi-sorted models.

## 3   Lambda logic and lambda unification

Lambda logic is the logical system one obtains by adding lambda calculus to first order logic. This system is formulated, and some fundamental metatheorems are proved, in [1]. The appropriate generalization of unification to lambda logic is this notion: two terms are said to be *lambda unified* by substitution $\sigma$ if $t\sigma = s\sigma$ is provable in lambda logic. An algorithm for producing lambda unifying substitutions, called *lambda unification*, is used in the theorem prover Otter-$\lambda$, which is based on lambda logic rather than first-order logic, but is built on the well-known first-order prover Otter [3]. In Otter-$\lambda$, lambda unification is used, instead of only first-order unification, in the inference rules of resolution, factoring, paramodulation, and demodulation.

We do not regard this work as a "combination of first-order logic and higher-order logic". Lambda logic is not higher-order, it is untyped. Lambda unification is not higher-order unification, it is unification in (untyped) lambda logic. While there probably are interesting connections to typed logics, some of the questions about those relationships are open at present, and out of the scope of this paper. Similarly, while there are projects aimed at combining first-order provers and higher-order provers, that approach is quite different from ours. Otter-lambda is a single, integrated prover, not a combination of a first-order prover and a higher-order prover. There is just one database of deduced clauses on which inferences are performed; there is no need to pass data between provers. Whether other provers can find proofs for the examples that Otter-lambda can find proofs for, we do not know and cannot report on. This paper is solely about lambda logic, lambda unification, and Otter-lambda. This subject offers quite enough complications for one paper.

In Otter-$\lambda$ input files, we write $lambda(x,t)$ for $\lambda x.\, t$, and we write $Ap(x,y)$ for $x$ applied to $y$, which is often abbreviated in technical papers to $x(y)$ or even $xy$. In this paper, $Ap$ will always be written explicitly, but we use both $lambda(x, t)$ and $\lambda x. t$.

Our main objective in this section is to define the lambda unification algorithm. As we define it here, this is a non-deterministic algorithm: it can return, in general, many different unifying substitutions for two given input terms. As implemented in Otter-lambda, it returns just one unifier, making some specific choice at each non-deterministic choice point. As for ordinary unification, the input is two terms $t$ and $s$ (this time terms of lambda logic) and the output, if the algorithm succeeds, is a substitution $\sigma$ such that $t\sigma = s\sigma$ is provable in lambda logic.

We first give the relatively simple clauses in the definition. These have to do with first-order unification, alpha-conversion, and beta-reduction.

The rule related to first-order unification just says that we try that first; for example $Ap(x, y)$ unifies with $Ap(a, b)$ directly in a first-order way. However, the usual recursive calls in first-order unification now become recursive calls to lambda unification. In other words: to unify $f(t_1, \ldots, t_n)$ with $g(s_1, \ldots, s_m)$, this clause does not apply unless $f = g$ and $n = m$; in that case we do the following:

```
for i = 1 to n {
    τ = unify(tᵢ, sᵢ);
    if (τ = failure)
        return failure;
    σ = σ ∘ τ;  }
return σ
```

Here the call to `unify` is a recursive call to the algorithm being defined.

The rule related to alpha-conversion says that, if we want to unify $lambda(z, t)$ with $lambda(x, s)$, let $\tau$ be the substitution $z := x$ and then unify $t\tau$ with $s$, rejecting any substitution that assigns a value depending on $x$.[2] If this unification succeeds with substitution $\sigma$, return $\sigma$.

The rule related to beta-reduction says that, to unify $Ap(lambda(z, s), q)$ with $t$, we first beta-reduce and then unify. That is, we unify $s[z := q]$ with $t$ and return the result.

Lambda unification's most interesting instructions tell how to unify $Ap(x, w)$ with a term $t$, where $t$ may contain the variable $x$, and $t$ does not have main symbol $Ap$. Note that the occurs check of first-order unification does not apply in this case. The term $w$, however, may not contain $x$. In this case lambda unification is given by the following non-deterministic algorithm:

1. Pick a *masking subterm* $q$ of $t$. That means a subterm $q$ such that every occurrence of $x$ in $t$ is contained in some occurrence of $q$ in $t$. (So $q$ "masks" the occurrences of $x$; if there are no occurrences of $x$ in $t$, then $q$ can be any subterm of $t$, but see the next step.)

---

[2] Care is called for in this clause, as illustrated by the following example: Unify $lambda(x, y)$ with $lambda(x, f(x))$. The "solution" $y = f(x)$ is wrong, since substituting $y = f(x)$ in $lambda(x, y)$ gives $lambda(z, f(x))$, because the bound variable is renamed to avoid capture.

2. Call lambda unification to unify $w$ with $q$. Let $\sigma$ be the resulting substitution. If this unification fails, or assigns any value other than a variable to $x$, return failure. If it assigns a variable to $x$, say $x := y$ reverse the assignment to $y := x$ so that $x$ remains unassigned.

3. If $q\sigma$ occurs more than once in $t\sigma$, then pick a set $S$ of its occurrences. If $q$ contains $x$ then $S$ must be the set of *all* occurrences of $q\sigma$ in $t$. Let $z$ be a fresh variable and let $r$ be the result of substituting $z$ in $t\sigma$ for each occurrence of $q\sigma$ in the set $S$.

4. Append the substitution $x := \lambda z.\, r$ to $\sigma$ and return the result.

There are two sources of non-determinism in the above, namely in steps 1 and 3. These steps are made deterministic in Otter-$\lambda$ as follows: in step 1, if $x$ occurs in $t$, we pick the largest masking subterm $q$ that occurs as a second argument of $Ap$.[3] If $x$ occurs in $t$, but no masking subterm occurs as a second argument of $Ap$, we pick the smallest masking subterm. If $x$ does not occur in $t$, we pick a constant that occurs in $t$; if there is none, we fail. In step 3, if $q$ does not contain $x$, then an important application of this choice is to proofs by mathematical induction, where the choice of $q$ corresponds to choosing a constant $n$, replacing some of the occurrences of $n$ by a variable, and deciding to prove the theorem by induction on that variable. Therefore the choice of $S$ is determined by heuristics that prove useful in this case. In the future we hope to implement a version of lambda unification that returns multiple unifiers by trying different sets $S$ in step 3. Our proofs in this paper apply to the full non-deterministic lambda unification, as well as to any deterministic versions, unless otherwise specified.

*Example.* Lambda unification can lead to untypable proofs, for example those needed to produce fixed points in lambda calculus. As an example, if we unify $Ap(x, y)$ with $f(Ap(x, y))$, the masking subterm $q$ is $x$ itself; $w$ is $y$ so $\sigma$ is $y := x$; $w\sigma$ is $x$ and $t\sigma$ is $Ap(x, x)$. Thus we get the following result:[4]

$$x := lambda(z, f(Ap(z, z))) \qquad\qquad y := x$$

Type restrictions will be violated if we have specified the typing:

$$type(B, Ap(i(A, B), A)). \qquad\qquad type(B, f(B)).$$

Variable $x$ has type $i(A, B)$, and variable $y$ has type $A$, so the unification of $x$ and $y$ violates type restrictions, since $i(A, B)$ is not the same type as $A$.

**Definition 3.** *We say that a particular lambda unification (of $Ap(X, w)$ with $t$) is* type-safe *(with respect to some explicit or implicit typings) if the masking subterm $q$ selected by lambda unification has the same type (with respect to those*

---

[3] The point of this choice is that, if we want the proof to be implicitly typable, then $q$ should be chosen to have the same type as $w$, and $w$ is a second argument of $Ap$.

[4] The symbol $i$ does not have to be "defined" here; type assignments can be arbitrary terms. But intuitively, $i(A, B)$ could be thought of as the type of functions from type $A$ to type $B$.

*typings) as the term w, and q is a proper subterm of t (unless the two arguments of Ap have the same type). We also require that the value type assigned to Ap(X, w) is the same as the value type assigned to t.*

The example preceding the definition illustrates a lambda unification that is not type-safe for *any* reasonable typing. The masking subterm is $x$; type safety would require $x$ to be assigned the same type as $y$. But $x$ occurs as a first argument of $Ap$ and $y$ as a second argument of $Ap$. Therefore the type specification of $Ap$ would have to be of the form $type(V, Ap(U, U))$; but normally $Ap$ will have a type specification of the form $type(B, Ap(i(A, B), A))$.

*Remark.* A discussion of the relationship, if any, between lambda unification and the higher-type unification algorithms already in the literature is beyond the scope of this paper. The algorithms apply to different systems and have different definitions. Similarly, the exact relationship between lambda logic and various sytems of higher-order logic, if there is any, is beyond the scope of this paper (or any paper of this length).

## 4   Implicit typing in lambda logic

If we consider the no-nilpotents example in lambda logic, we can state the axiom of mathematical induction in full generality, and Otter-lambda can use lambda unification to find the specific instance of induction that is required. (See the examples on the Otter-lambda website.) The proof, obtained without relativizing to unary predicates, is correctly typable. This is not an accident: there are theorems about implicit typing that guarantee it.

We first give an example to show that the situation is not as straightforward as in first-order logic. If we use the axioms of group theory in lambda logic, must we relativize them to a unary predicate $G(x)$? As we have seen above, that is not necessary when doing first-order inference. We could, for example, put in some axioms about natural numbers, and not relativize them to a unary predicate $N(x)$, and as long as our axioms are correctly typed, our proofs will be correctly typed too. There is, however, reason to worry about this when we move to lambda logic.

In lambda calculus, every term has a fixed point. That is, for every term $F$ we can find a term $q$ such that $Ap(F, q) = q$. Another form of the fixed point theorem says that for each term $H$, we can find a term $f$ such that $Ap(f, x) = H(f, x)$. Applying this to the special case when $H(f, x) = c * Ap(f, x)$, where $c$ is a constant and $*$ is the group multiplication, we get $Ap(f, x) = c * Ap(f, x)$. It follows from the axioms of group theory that $c$ is the group identity. On the other hand, in lambda logic it is given as an axiom that there exist two distinct objects, say $c$ and $d$, and since each of $d$ and $c$ must equal the group identity, this leads to a contradiction. Looked at model-theoretically, this means it is impossible, given a lambda model $M$, to define a binary operation on $M$ and an identity element of $M$ that make $M$ into a group.

Since these axioms are contradictory in lambda logic, what is the value of a proof of a theorem from these axioms? One might think that there is none, and that to be able to trust an automatically produced proof from these axioms I would need to check it independently, or reformulate the axioms by relativizing the group axioms to a unary predicate $G$. The point of this paper is that there are good theoretical reasons why I do *not* need to do that. Even though there exists a derivation of a contradiction in lambda logic from these axioms, it is not a well-typed derivation, and since the axioms are well-typed, the theorems in this paper guarantee that deduced conclusions will also be well-typed. In other words, if we attempt to prove that every element is equal to $c$, we will put in the negated goal $x \neq c$, but if we use only "type-safe" lambda unification, as defined below, we will not be able to construct the fixed point needed to derive a contradiction. If, however, we use unrestricted lambda unification, we can derive it. If we put in the (negation of) the untyped fixed-point equation itself, then we can also prove that (even with type-safe lambda unification), but we need a non-well typed axiom in the input file.

First, let us consider how to type the relevant axioms. Writing $G$ for the type of group elements, 1 for the group identity, and $i(G, G)$ for the type of maps from $G$ to $G$, we would have the following type specifications:

$$type(G, 1).$$
$$type(G, *(G, G)).$$
$$type(G, Ap(i(G, G), G)).$$
$$type(i(G, G), lambda(G, G)).$$

In general, of course, we want $type(i(X, Y), lambda(X, Y))$, but the special case shown is enough in this example. According to these type specifications, the axioms are correctly typable, and when Otter-$\lambda$ produces a proof, the proof turns out to also be correctly typable. This is not an accident, as we will see.

In defining type specifications for lambda logic, the following technicality comes up: Normally in predicate logic we tacitly assume that different symbols are used for function symbols and predicate symbols. Thus $P(P(c))$ would not be considered a well-formed formula. In lambda logic we do wish to be able to define propositional functions, as well as functions whose values are other objects, so we allow $Ap$ both as a predicate symbol and a function symbol. However, except for $Ap$, we follow the usual convention that predicate symbols and function symbols use distinct alphabets. This is the reason for clauses (4) and (5) in the following definition.

**Definition 4.** *A list of type specifications $S$ is called* coherent *if*
*(1) for each (predicate or function) symbol $f$ ( except possibly $Ap$ and $lambda$) and arity $n$, it contains at most one type specification of symbol $f$ and arity $n$; the value type of a predicate symbol must be $Prop$ and of a function symbol, must not be $Prop$.*
*(2) $type(i(X, Y), lambda(X, Y))$ belongs to $S$ if and only if*
*$type(Y, Ap(i(X, Y), X))$ belongs to $S$.*
*(3) all type specifications with symbol $Ap$ have the form $type(V, Ap(i(U, V), U))$, for the same type $U$, which is called the "ground type" of $S$.*

*(4) all type specifications with symbol lambda have the form*
*type(i(U, V), lambda(U, V)),*[5] *where U is the ground type of S.*

*(5) There are at most two type specifications in S with symbol Ap; if there are two, then exactly one must have value type Prop.*

Conditions (2) and (3) guarantee that beta-reduction carries correctly typed terms to correctly typed terms. One might wish for a less restrictive condition in (4) and (5), allowing functions of functions, or functions of functions of functions, etc. But this is the condition for which we can prove theorems at the present time, and it covers a number of interesting examples in algebra and number theory.

If $S$ is a coherent list $S$ of type specifications, it makes sense to speak of "the type assigned to a term $t$ by $S$", if there is at least one type specification in $S$ for the main symbol and arity of $t$. Namely, unless the main symbol of $t$ is $Ap$, only one specification in $S$ can apply, and if the main symbol of $t$ is $Ap$, then we apply the specification that does not have value type $Prop$. Similarly, it makes sense to speak of "the type assigned to an atomic formula by $S$". When the main symbol of $t$ is $Ap$, we can speak of "the type assigned to $t$ as a term" or "the type assigned to $t$ as a formula", using the specification that does not or does have $Prop$ for its value type.

**Theorem 3.** *Let $S$ be a coherent list of type specifications. Let $s$ and $t$ be two correctly typed terms or two correctly typed atomic formulas with respect to $S$. Let $\sigma$ be a substitution produced by successful type-safe lambda unification of $s$ and $t$. Then $s\sigma$ and $t\sigma$ are correctly typed, and $S$ assigns the same type to $s$, $t$, and $s\sigma$.*

*Example.* Let $s$ be $Ap(X, w)$ and $t$ be $a + b$. We can unify $s$ and $t$ by the substitution $\sigma$ given by $X := lambda(x, x + b)$ and $w := a$. If $type(0, Ap(i(0, 0), 0))$ and $type(0, +(0, 0))$ then these are correctly typed terms and the types of $s\sigma$ and $a + b$ are both 0. It may be that $Ap$ also has a type specification $type(Prop, Ap(i(0, Prop), 0))$, used when the first argument of $Ap$ defines a propositional function. However, this additional type specification will not lead to mis-typed unifications, since the two type specifications of $Ap$ are coherent.

*Proof.* We proceed by induction on the length of the computation by lambda unification of the substitution $\sigma$.

(i) Suppose $s$ is a term $f(r, q)$ (or with more arguments to $f$), and either $f$ is not $Ap$, or $r$ is neither a variable nor a lambda term. Then $t$ also as the form $f(R, Q)$ for some $R$ and $Q$, and $\sigma$ is the result of unifying $r$ with $R$ to get $r\tau = R\tau$ and then unifying $q\tau$ with $Q\tau$, producing substitution $\rho$ so that $\sigma = \tau \circ \rho$. By the induction hypothesis, $r\tau$ is correctly typed and gets the same type as $r$ and $R\tau$; again by the induction hypothesis, $q\tau\rho$ and $Q\tau\rho$ are correctly typed and get the same type as $q$. Then $s\sigma = f(r\sigma, q\sigma) = f(r\tau\rho, q\tau\rho)$ is also correctly typed.

---

[5] Intuitively, this says that if $z$ has type $X$ and $t$ has type $Y$ then $lambda(z, t)$ has type $i(X, Y)$, the type of functions from $X$ to $Y$.

(ii) The argument in (i) also applies if $s$ is $Ap(r, q)$ and $t$ is $Ap(R, Q)$ and lambda unification succeeds by unifying these terms as if they were first-order terms.

(iii) If $s$ is a constant then $s\sigma$ is $s$ and there is nothing to prove.

(iv) If $s$ is a variable, what must be proved is that $t$ and $s$ have the same value type. A variable must occur as an argument of some term (or atom) and hence the situation really is that we are unifying $P(s, \ldots)$ with some term $q$, where $P$ is either a function symbol or a predicate symbol. If $P$ is not $Ap$, then $q$ must have the form $P(t, \ldots)$, and $t$ and $s$ occur in corresponding argument positions (not necessarily the first as shown). Since these terms or atoms $P(t, \ldots)$ and $P(s, \ldots)$ are correctly typed, and $S$ is coherent, $t$ and $s$ do have the same types. The case when $P$ is $Ap$ will be treated below.

(v) Suppose $s$ is $Ap(r, q)$, where $r = lambda(z, p)$, and $z$ does occur in $p$. Then $s$ beta-reduces to $p[z := q]$, and lambda unification is called recursively to unify $p[z := q]$ with $t$. By induction hypothesis, $t$, $t\sigma$, $p[z := q]$, and $p[z := q]\sigma$ are well-typed and are assigned the same value type, which must be the value type, say $V$, of $p$. Since $S$ is coherent, the type assigned to $lambda(z, p)$ is $i(U, V)$, where $U$ is the "ground type", the type of the second arg of $Ap$. The type of $q$ is $U$ since $q$ occurs as the second arg of $Ap$ in the well-typed term $s$. The type of $s$, which is $Ap(r, q)$, is $V$. We must show that $s\sigma$ is well-typed and assigned the value type $V$. Now $s\sigma$ is $Ap(r\sigma, q\sigma)$. It suffices to show that $q\sigma$ has type $U$ and $r\sigma$ has type $i(U, V)$. We first show that the type of $q\sigma$ is $U$. Since $z$ has type $U$ in $lambda(z, p)$, $q\sigma$ occurs in the same argument positions in $p[z := q]\sigma$ as $z$ does in $p$, and since $z$ does occur at least once in $p$, and $p[z := q]\sigma$ is well-typed, $q\sigma$ must have the same type as $z$, namely $U$. Next we will show that $r\sigma$ has type $i(U, V)$. We have $r\sigma = lambda(z, p)\sigma = lambda(z, p\sigma)$ (since the bound variable $z$ is not in the domain of $\sigma$). We have $p\sigma[z := q\sigma] = p[z := q]\sigma$ and the type of the latter term is $V$ as shown above. The type of $A[z := B]$ is the type of $A$, and moreover $A[z := B]$ is well-typed provided $A$ and $B$ are well-typed and $z$ gets the same type as $B$. That observation applies here with $A = p\sigma$ and $B = q\sigma$, since the type of $z$ is $U$ and the type of $q\sigma$ is $U$. Therefore the type of $p\sigma$ is the same as the type of $p\sigma[z := q\sigma]$, which is the same as $p[z := q]\sigma$, which has type the same as $p[z := q]$, which we showed above to be $V$. Since $r\sigma = lambda(z, p\sigma)$, and $z$ has type $U$, $r\sigma$ has type $i(U, V)$, which was what had to be proved.

(vi) There are two cases not yet treated: when $s$ is $Ap(X, w)$, and when $s$ is a variable $X$ occurring in the context $Ap(X, w)$. We will treat these cases simultaneously. As described in the previous section, the algorithm will (1) select a masking subterm $q\sigma$ of $t\sigma$ (2) unify $w$ and $q$ with result $\sigma$ (failing if this fails), (3) create a new variable $z$, and substitute $z$ for some or all occurrences of $q\sigma$ in $t\sigma$, obtaining $r$, and (4) produce the unifying substitution $\sigma$ together with $X := lambda(z, r)$.

Assume that $t$ is a correctly typed term. Then every occurrence of $q$ in $t$ has the same type, by the definition of correctly typed. Since by hypothesis this is type-safe lambda unification, $q$ and $w$ have the same type, call it $U$.

Since $q$ unifies with $w$, by the induction hypothesis $q\sigma$ and $w\sigma$ are correctly typed and get the same types as $q$ and $w$, respectively, namely $U$. If $Ap(X, w)$ has type $Prop$, then the type of $s$ and that of $t$ are the same by hypothesis. Otherwise, both occur as arguments of some function or predicate symbol $P$, in corresponding argument positions, and hence, by the coherence of $S$, they are assigned the same (value) type $V$. Then $X$ has the type $i(U, V)$. We now assign the fresh variable $z$ the type $U$; then $r$ is also correctly typed, and gets the same type $V$ as $s$ and $t$, since it is obtained by substituting $z$ for some occurrences of $q\sigma$ in $t\sigma$. For this last conclusion we need to use the fact that $q$ is a proper subterm of $t$, by the definition of type-safe unification; hence $r$ is not a variable, so the value type of $r$ is well-defined, since $S$ is coherent. Since $S$ is coherent, there is a type specification in $S$ of the form $type(i(U, V), lambda(U, V))$. Thus the term $lambda(z, r)$ can be correctly typed with type $i(U, V)$, the same type as $X$. Hence $X\sigma$ has the same type as $X$, and $s\sigma$ has the same type as $s$. That completes the proof of the theorem.

**Theorem 4 (Implicit Typing for Lambda Logic).** *Let $A$ be a set of clauses, and let $S$ be a coherent set of type specifications such that each clause in $A$ is correctly typable with respect to $S$. Then all conclusions derived from $A$ by binary resolution, hyperresolution, factoring, paramodulation, and demodulation (including beta-reduction), using type-safe lambda unification in these rules of inference, are correctly typable with respect to $S$, provided paramodulation from or into variables are not allowed, and paramodulation into or from terms $Ap(X, w)$ with $X$ a variable is not allowed, and demodulators similarly are not allowed to have variables or $Ap(X, w)$ terms on the left.*

*Remark*. The second restriction on paramodulation is necessary, as shown by the following example. Suppose $Ap$ has a type specification $type(Prop, Ap(i, 0, Prop), 0))$. Without the restriction, we could paramodulate from $x + 0 = x$ into $Ap(X, x)$, unifying $x + 0$ with $Ap(X, x)$ as in the example after Theorem 3, with the substitution $X := lambda(x, x + 0)$. The conclusion of the paramodulation inference would be $x$. That is a mistyped conclusion, since $x$ does not have the type $Prop$, although $Ap$ does have value type $Prop$.

*Proof.* Note that a typing assigns type symbols to variables, and the scope of a variable is the clause in which it occurs, so as usual with resolution, we assume that all the variables are renamed, or indexed with clause numbers, or otherwise made distinct, so that the same variable cannot occur in different clauses. In that case the originally separate correct typings $T[i]$ (each obtained from $S$ by assigning values to varaibles in clause $C[i]$) can be combined (by union of their graphs) into a single typing $T$. We claim that the set of clauses $A$ is correctly typed with respect to this typing $T$. To prove this correctness we need to prove:

(i) *each occurrence of a variable in $A$ is assigned the same type by $T$*. This follows from the correctness of $C[i]$, since because the variables have been renamed, all occurrences of any given variable are contained in a single clause $C[i]$.

(ii) *If $r$ is $f(u, v)$, and $r$ occurs in $A$, and $f(u, v)$, $u$, and $v$ are assigned types $a$, $b$, $c$ respectively, then there is a type specification in $S$ of the form $type(a, f(b, c))$.*

If the term $r$ occurs in $A$, then $r$ occurs in some $C[i]$, so by the correctness of $T[i]$, there is a type specification in $S$ as required.

(iii) *each occurrence of each term $r$ that occurs in $A$ has the same value type.* This follows from the coherence of $S$. The different typings $T[i]$ are not allowed to assign different value types to the same symbol and arity.

Hence $A$ is correctly typed with respect to $T$.

All references to correct typing in the rest of the proof refer to the typing $T$.

We prove by induction on the length of proofs that all proofs from $A$ using the specified rules of inference lead to correctly typed conclusions. The base case of the induction is just the hypothesis that $A$ is correctly typable. For the induction step, we take the rules of inference one at a time. We begin with binary resolution. Suppose the two clauses being resolved are $P|Q$ and $-R|B$, where substitution $\sigma$ is produced by lambda unification and satisfies $P\sigma = R\sigma$. Here $Q$ and $B$ can stand for lists of more than one literal, in other words the rest of the literals in the clause, and the fact that we have shown $P$ and $-R$ as the first literals in the clause is for notational convenience only. By hypothesis, $P|Q$ is correctly typed with respect to $S$, and so is $-R|B$, and by Theorem 3, $P\sigma|Q\sigma$ and $-R\sigma|B\sigma$ are also correctly typed. The result of the inference is $Q\sigma|B\sigma$. But the union of correctly typed terms, literals, or sets of literals (with respect to a coherent set of type specifications) is again correctly typed, by the same argument as in the first part of the proof. In other words, coherence implies that if some subterm $r$ occurs in both $Q\sigma$ and in $B\sigma$ then $r$ gets the same value type in both occurrences. That completes the induction step when the rule of inference is binary resolution.

Hyperresolution and negative hyperresolution can be "simulated" by a sequence of binary resolutions, so the case in which the rule of inference is hyperresolution or negative hyperresolution reduces to the case of binary resolution. The rule of "factoring" permits the derivation of a new clause by unifying two literals in the same clause that have the same sign, and applying the resulting substitution to the entire clause. By Theorem 3, a clause derived in this way is well-typed if its premise is well-typed.

Now consider paramodulation. In that case we have already deduced $t = q$ and $P[z := r]$, and unification of $t$ and $r$ produces a substitution $\sigma$ such that $t\sigma = r\sigma$. The conclusion of the rule is $P[z := q\sigma]$. We have disallowed paramodulation from or into variables in the statement of the theorem; therefore $t$ and $r$ are not variables. Let us write $Type(t)$ for the value type of (any term) $t$. Because $t = q$ is correctly typed, we have $Type(t) = Type(q)$. If neither $t$ nor $q$ is an $Ap$ term, then $Type(t\sigma) = Type(q\sigma)$, since they have the same functor. If one of them is an $Ap$ term, then by hypothesis it is not of the form $Ap(X, w)$, with $X$ a variable. Then by Theorem 3, $Type(t\sigma) = Type(t)$ and $Type(q\sigma) = Type(q) = Type(t) = Type(t\sigma)$. Thus in any case $Type(q\sigma) = Type(t\sigma)$. The value type of $r$ is the same at every occurrence, since $P[z := r]$ is correctly typed. To show that $P[z := q\sigma]$ is correctly typed, it suffices to show that $Type(q\sigma) = Type(r)$, which is the same as the type of $r\sigma$. Since the terms $t$ and $r$ unify, and neither is a variable, their main symbols are the same, since by hypothesis $r$ is not of the form $Ap(X, w)$. Hence $Type(r) = Type(r\sigma) = Type(t\sigma) = Type(q\sigma)$, which is what had to be shown.

Now consider demodulation. In this case we have already deduced $t = q$ and $P[z := t\sigma]$ and we conclude $P[z := q\sigma]$, where the substitution $\sigma$ is produced by lambda unification of $t$ with some subterm $\rho$ of $P[z := \rho]$. Taking $r = t\sigma$, we see that demodulation is a special case of paramodulation, so we have already proved what is required. That completes the proof of the theorem.

*Example: fixed points.* The fixed point argument which shows that the group axioms are contradictory in lambda logic requires a term $Ap(f, Ap(x, x))$. The part of this that is problematic is $Ap(x, x)$. If the type specification for $Ap$ is $type(V, Ap(i(U, V), U))$, then for $Ap(x, x)$ to be correctly typed, we must have $V = U = i(U, U)$. If $U$ and $V$ are type symbols, this can never happen, so the fixed point construction cannot be correctly typed. It follows from the theorem above that this argument cannot be found by Otter-$\lambda$ from a correctly typed input file. In particular, in lagrange3.in we have correctly typed axioms, so we will not get a contradiction from a fixed point argument.

On the other hand, in file lambda4.in, we show that Otter-$\lambda$ can verify the fixed-point construction. The input file contains the negated goal

$$Ap(c, Ap(lambda(x, Ap(c, Ap(x, x))), lambda(x, Ap(c, Ap(x, x)))))$$
$$\neq Ap(lambda(x, Ap(c, Ap(x, x))), lambda(x, Ap(c, Ap(x, x)))).$$

Since this contains the term $Ap(x, x)$, it cannot be correctly typed with respect to any coherent list of type specifications $T$. Otter-$\lambda$ does find a proof using this input file, which is consistent with our argument above that fixed-point constructions will not occur in proofs from correctly typable input files. The fact that the input file cannot be correctly typed, which we just observed directly, can also be seen as a corollary of the theorem, since Otter-$\lambda$ finds a proof. The fact that the theoretical result agrees with the results of running the program is a good thing.

*Remarks.* (1) The (unrelativized) axioms of group theory are contradictory in lambda logic, but if we put in only correctly-typed axioms, Otter-$\lambda$ will find only correctly typed proofs, which will be valid in the finite type structure based on any group, and hence will not be proofs of a contradiction.

(2) We already knew that resolution plus factoring plus paramodulation from non-variables is not refutation-complete, even for first-order logic; and we remarked when pointing that out that this permits typed models of some theories that are inconsistent when every object must have the same type. Here is another illustration of that phenomenon in the context of lambda logic.

(3) Of course Otter-lambda *can* find the fixed-point proof that gives the contradiction; but to make it do so, we need to put in some non-well-typed axiom, such as the negation of the fixed-point equation.

## 5 Enforcing type-safety

The theorems above are formulated in the abstract, rather than being theorems about a particular implementation of a particular theorem-prover. As a practical matter, we wish to formulate a theorem that does apply to Otter-$\lambda$ and covers the examples posted on the Otter-$\lambda$ website, some of which have been mentioned here. Otter-$\lambda$ never uses paramodulation into or from variables, so that hypothesis of the above theorems is always satisfied. But Otter-$\lambda$ does not always use only type-safe lambda unification; nor would we want it to do so, since it can find some untyped proofs of interest, e.g. fixed points, Russell's paradox, etc. Once Otter-$\lambda$ finds a correctly typable proof, we can check by hand (and could easily check by machine) that it is correctly typable. Nevertheless it is of interest to be able to set a flag in the input file that enforces type-safe unification. In Otter-$\lambda$, if you put `set(types)` in the input file, then only certain lambda unifications will be performed, and those unifications will always be type-safe.

Specfically, *restricted* lambda unification means that, when selecting a masking subterm, only a second argument of $Ap$ or a constant will be chosen. This is the restriction imposed by the flag `set(types)`. We now prove that this enforces type safety under certain conditions.

**Theorem 5 (Type safety of restricted lambda unification).** *Suppose that a given set of axioms admits a coherent type specification in which there is no typing of the form $Ap(U, U)$, and all constants receive type $U$. Then all deductions from the given axioms by binary resolution, factoring, hyperresolution, demodulation (including beta-reduction) paramodulation (except into or from variables and $Ap$ terms), lead to correctly typable conclusions, provided that restricted lambda unification is used in those rules of inference.*

*Proof.* It suffices to show that lambda unifications will be type-safe under these hypotheses. The unification of $Ap(x, w)$ with $t$ is type-safe (by definition) if in step (1) of the definition of lambda unification, the masking subterm $q$ of $t$ has the same type as $w$. Now $q$ is either a constant or term containing $x$ that appears as a second argument of $Ap$, since those are the "restrictions" in restricted lambda unification. If $q$ is a variable then it must be $x$, and must occur as a second argument of $Ap$; but $x$ occurs as a first argument of $Ap$, and all second arguments of $Ap$ get the same type, so there must be a typing of the form $type(T, Ap(U, U))$. But such a typing is not allowed, by hypothesis. Therefore $q$ is not a variable. Then if $q$ contains $x$, it must occur as a second argument of $Ap$, as does $w$; hence by hypothesis $w$ and $q$ get the same type. Hence we may assume $q$ is a constant. But by hypothesis, all constants get the same type as the second arguments of $Ap$. That completes the proof.

## 6 Some examples covered by Theorem 5

It remains to substantiate the claims made in the abstract and introduction, that the theorems in this paper justify the use of implicit typing in Otter-$\lambda$ for the various examples mentioned. The first theorems apply in generality to any partial implementation of non-deterministic lambda unification, used in combination with resolution and paramodulation, but disallowing paramodulation into and from variables. Only Theorem 5 applies to Otter-lambda specifically, when the *set(types)* command is in the input file. We will now check explicitly that interesting examples are covered by this theorem.

Let us start with the "no nilpotents" example. It appears *prima facie* not to meet the hypotheses of Theorem 5, since that theorem requires that all constants have the same type as the second argument of $Ap$. In this example the type of $Ap$ is the one needed for mathematical induction: $type(Prop, Ap(i(N, Prop), N))$, so the type of the second arg of $Ap$ is $N$; but the axioms include a constant $o$ for the zero of the ring. This is not a serious problem: we can simply replace $o$ in the axioms by $zero(0)$, where $zero$ is a new function symbol with the type specification $type(R, zero(N))$. (The name $zero$ is immaterial; this is just some function symbol.) The term $zero(0)$ is not a constant, so it won't be selected as a masking term (where it would interfere with the proof of Theorem 5). But it will be treated essentially as a constant elsewhere in the inference process; and if we were worried about that, we could use a weight template to ensure that it has the same weight as a constant and hence will be treated *exactly* as a constant. On the logical side we have the following lemma to justify the claim:

**Lemma 1.** *Let $T$ be a theory with at least one constant $c$. Let $T^*$ be obtained from $T$ by adding a new function symbol $f$, but no new axioms. Then (i) $T^*$ plus the axioms $c = f(x)$ is conservative over $T$.*

*(ii) If $T$ contains another constant $b$ and we let $A^o$ be the result of replacing $c$ by $f(b)$ in $A$, then $T$ proves $A$ if and only if $T^*$ proves $A^o$.*

*(iii) There is an algorithm for transforming any proof of $A^o$ in $T^*$ to a proof of $A$ in $T$.*

*Proof.* (i) Every model of $T$ can be expanded to a model of $T^*$ plus $c = f(x)$ by interpreting $f$ as the constant function whose value is the interpretation of $c$. The completeness theorem then yields the stated conservative extension result.

(ii) $A^o$ is equivalent to $A$ in $T^*$ plus $c = f(x)$, so by (i), $A^o$ is provable in $T^*$ plus $c = f(x)$ if and only if $T$ proves $A$. In particular, if $T^*$ proves $A^o$ then $T$ proves $A$. Conversely, if $T$ proves $A$ and we just replace $c$ with $f(b)$ in the proof, we get a proof of $A^o$ in $T^*$.

(iii) The algorithm is fairly obvious: simply replace every term $f(t)$ in the proof with $c$. (Not just terms $f(b)$ but any term with functor $f$ is replaced by $c$.) Terms that unify before this replacement will still unify after the replacement, so resolution proof steps will remain valid. The axioms of $T^*$ plus $c = f(x)$ are converted to axioms of $T$ plus $c = c$. Paramodulation steps remain paramodulation steps and demodulations remain demodulations. Since no variables are introduced, paramodulations that were not from or into variables are still not from or into variables. That completes the proof of the lemma.

This lemma shows us that logically, the formulation of the no-nilpotents problem with $zero(0)$ for the ring zero is equivalent to the original formulation with a constant $o$ for the ring zero; and Theorem 5 directly applies to the formulation with $o(0)$. In practice, if we run Otter-lambda with $o$ replaced by $zero(0)$ in the input file, we find the same proof as before, but with $o$ replaced by $zero(0)$. In essence this amounts to the observation that $o$ was never used as a masking term in lambda unification in the original proof. Technically we should run the input file with $zero(0)$ first. Theorem 5 guarantees that if we find a proof, it will be well-typed. The lemma guarantees that we can convert it into a proof of the original formulation using a text editor to replace all terms with functor $zero$ by the original constant $o$.

*Remark.* Of course there is little difference between $0$ and $zero(0)$, and of course we could allow the user of Otter-lambda to specify which constants have "ground type" and which do not, and only use constants of "ground type" in lambda-unification. In effect that is what this theorem allows us to do, without checking the types of constants at run time: just rename all the "non-ground" constants by wrapping them in one extra function symbol.

We conclude with another example. Bernoulli's inequality is

$$(1 + nx) < (1 + x)^n \qquad \text{if } x > -1 \text{ and } n > 0 \text{ is an integer.}$$

Otter-lambda, in a version that calls on MathXpert [2] for "external simplification", is able to prove this inequality by induction on $n$, being given only the clausal form of Peano's induction axiom, with a variable for the induction predicate. The interest of the example in the present context is the fact that three types are involved: real numbers, positive integers, and propositions. The propositional functions all have $N$, the non-negative integers, for the ground type, but the types are not disjoint: $N$ is a subset of the reals $R$. Moreover, the left-hand side of the inequality uses $n$ in multiplication, so if multiplication is typed to take two real arguments, the inequality as it stands will not be well-typed.

The solution is to introduce a symbol for an injection map $i : N \rightarrow R$. The inequality then becomes

$$(i(1) + i(n)x) < (i(1) + x)^n$$

This formulation is well-typed, if we type $i$ as a function from $N$ to $R$. Again, in the definition of exponentiation we have to use $0$ for the natural number zero, and $zero(0)$ for the real number zero, so that all the constants will have type $N$. If that is done, Theorem 5 applies, and we can be assured that the inference steps performed by Otter-lambda proper will lead from well-typed formulas to well-typed formulas. However, the theorem does not cover the external simplification steps performed by MathXpert. To ensure that these do not lead to mis-typed conclusions, we have to discard any results containing a minus sign or division sign, as that might lead out of the domain of integers. Problems involving embedded subtypes also arise even in typed theorem provers or proof checkers, so it is interesting that those problems are easily solved in lambda logic. The interested reader can find the input and output files for this and other examples on the Otter-lambda website.

## References

1. Beeson, M., Lambda Logic, in Basin, David; Rusinowitch, Michael (eds.) Automated Reasoning: Second International Joint Conference, IJCAR 2004, Cork, Ireland, July 4-8, 2004, Proceedings. Lecture Notes in Artificial Intelligence 3097, pp. 460-474, Springer (2004).
2. MathXpert Calculus Assistant, software available from (and described at) www.HelpWithMath.com.
3. McCune, W., Otter 3.0 Reference Manual and Guide, Argonne National Laboratory Tech. Report ANL-94/6, 1994.
4. Wick, C., and McCune, W., Automated reasoning about elementary point-set topology, *J. Automated Reasoning* **5(2)** 239–255, 1989.

# System Description: Leo − A Resolution based Higher-Order Theorem Prover

Christoph Benzmüller

Fachbereich Informatik, Universität des Saarlandes
66041 Saarbrücken, Germany (`www.ags.uni-sb.de/~chris`)

**Abstract.** We present Leo, a resolution based theorem prover for classical higher-order logic. It can be employed as both an fully automated theorem prover and an interactive theorem prover. Leo has been implemented as part of the $\Omega$mega environment [23] and has been integrated with the $\Omega$mega proof assistant. Higher-order resolution proofs developed with Leo can be displayed and communicated to the user via $\Omega$mega's graphical user interface Loui. The Leo system has recently been successfully coupled with a first-order resolution theorem prover (Bliksem).

## 1 Introduction

Many of today's proof assistants such as Isabelle [22, 20], Pvs [21], Hol [12], Hol-Light [13], and Tps [2, 3] employ classical higher-order logic (also known as Church's simple type theory) as representation and reasoning framework. One important motivation for the development of automated higher-order proof tools thus is to relieve the user of tedious interactions within these proof assistants by automating less ambitious (sub)problems.

In this paper we present Leo, an automated resolution based theorem prover for classical higher-order logic. Leo is based on extensional higher-order resolution which, extending Huet's constrained resolution [14, 15], proposes a goal directed, rule based solution for extensionality reasoning [6, 4, 5]. The main motivation for Leo is to serve as an automated subsystem in the mathematics assistance system $\Omega$mega [23]. Additionally, Leo was intended to serve as a standalone automated higher-order resolution prover and to support the illustration and tutoring of extensional higher-order resolution. A previous system description of Leo has been published in [7]. Novel in this system description is the section on Leo's interaction facilities and its graphical user interface. We also provide more details on Leo's automation and point to some recent extensions.

This system description is structured as follows: In Sections 2 and 3 we briefly address Leo's connection with $\Omega$mega and Leo's calculus. Section 4 presents the interactive theorem prover Leo and Section 5 the automated theorem prover Leo. In Section 6 we illustrate how Leo's resolution proofs can be inspected with $\Omega$mega's graphical user interface Loui. Some experiments with Leo are mentioned in Section 7 and Section 8 concludes the paper.

## 2 Leo is a Subsystem of $\Omega$mega

Leo has been realized as a part of the $\Omega$mega framework. This framework (and thus also Leo) is implemented in Clos [25], an object-oriented extension of Lisp. Leo is mainly dependent on $\Omega$mega's term datastructure package Keim. The Keim package provides many useful data structures (e.g., higher-order terms, literals, clauses, and substitutions) and basic algorithms (e.g., application of substitution, subterm replacement, copying, and renaming). Thus, the usage of Keim allows for a rather quick implementation of new higher or first-order theorem proving systems. In addition to the code provided by the Keim-package, Leo consists of approximately 7000 lines of Lisp code.

Amongst the benefits of the realization of Leo within the $\Omega$mega framework are:

- Employing Keim supported a quick implementation of Leo. Important infrastructure could be directly reused or had to be only slightly adapted or extended.
- An integration of Leo with the proof assistant and proof planner $\Omega$mega was easily possible.
- Leo can easily be combined with other external systems already integrated with $\Omega$mega. Combinations of reasoning systems are particularly well supported in $\Omega$mega with the help of the agent based reasoning framework Oants [8].
- Leo may retrieve and store theorems and definitions via $\Omega$mega from Mbase, which is a structured repository of mathematical knowledge.
- Leo employs $\Omega$mega's input language Post.

There are also drawbacks of Leo's realization as a part of $\Omega$mega:

- Leo's latest version is only available in combination with the $\Omega$mega package. Installation of $\Omega$mega, however, is very complicated. Consequently, there is a conflict with the objective of providing a lean standalone theorem prover.
- The Keim datastructures are neither very efficient nor are they optimized or easily optimizable with respect to Leo.

$\Omega$mega and with it Leo can be download from http://www.ags.uni-sb.de/~omega.

## 3 Leo Implements Extensional Higher-Order Resolution

Leo is based on a calculus for extensional higher-order resolution. This calculus is described in [6, 4] and more recently in [5].

Extensionality treatment in this calculus is based on goal directed extensionality rules which closely connect the overall refutation search with unification by allowing for mutually recursive calls. This suitably extends the higher-order $E$-unification and $E$-resolution idea, as it turns the unification mechanism into a most general, dynamic theory unification mechanism. Unification may now itself employ a Henkin complete higher-order theorem prover as a subordinated reasoning system and the theory under consideration (which is defined by the sum of all clauses in the actual search space) dynamically changes.

In order to illustrate LEO's extensional higher-order resolution approach we discuss the TPTP (v3.0.1 as of 20 January 2005, see http://www.tptp.org) example SET171+3. This problem addresses distributivity of union over intersection[1]

$$\forall A_{o\alpha}, B_{o\alpha}, C_{o\alpha} \bullet A \cup (B \cap C) = (A \cup B) \cap (A \cup C)$$

In a higher-order context we can define the relevant set operations as follows

$$\cup := \lambda Y_{o\alpha}, Z_{o\alpha} \bullet (\lambda x_{\alpha} \bullet x \in Y \vee x \in Z)$$

$$\cap := \lambda Y_{o\alpha}, Z_{o\alpha} \bullet (\lambda x_{\alpha} \bullet x \in Y \wedge x \in Z)$$

$$\in := \lambda Z_{\alpha}, X_{o\alpha} \bullet (X\ Z)$$

After recursively expanding the definitions in the input problem, i.e., completely reducing it to first principles, LEO turns it into a negated unit clause. Unlike in standard first-order resolution, clause normalization is not a pre-process in LEO but part of the calculus. Internalized clause normalization is an important aspect of extensional higher-order resolution in order to support the (recursive) calls from higher-order unification to the higher-order reasoning process. Thus, LEO internally provides rules to deal with non-normal form clauses and this is why it is sufficient to first turn the input problem into a single, usually non-normal, negated unit clause. Then LEO's internalized clause normalization process can take care of subsequent normalization.

In our concrete case, this normalization process leads to the following unit clause consisting of a (syntactically not solvable) unification constraint (here $B_{o\alpha}, C_{o\alpha}, D_{o\alpha}$ are Skolem constants and $Bx$ is obtained from expansion of $x \in B$):

$$[(\lambda x_{\alpha} \bullet Bx \vee (Cx \wedge Dx)) =^? (\lambda x_{\alpha} \bullet (Bx \vee Cx) \wedge (Bx \vee Dx))]$$

Note that negated primitive equations are generally automatically converted by LEO into unification constraints. This is why $[(\lambda x_{\alpha} \bullet Bx \vee (Cx \wedge Dx)) =^? (\lambda x_{\alpha} \bullet (Bx \vee Cx) \wedge (Bx \vee Dx))]$ is generated, and not $[(\lambda x_{\alpha} \bullet Bx \vee (Cx \wedge Dx)) = (\lambda x_{\alpha} \bullet (Bx \vee Cx) \wedge (Bx \vee Dx))]^F$. Observe that we write $[.]^T$ and $[.]^F$ for positive and negative literals, respectively. This unification constraint has no 'syntactic' pre-unifier. It is solvable 'semantically' though with the help of the extensionality principles. Thus, Leo applies its goal directed functional and Boolean extensionality rules which replace this unification constraint (in an intermediate step) by the negative literal (where $x$ is a Skolem constant):

$$[(Bx \vee (Cx \wedge Dx)) \Leftrightarrow ((Bx \vee Cx) \wedge (Bx \vee Dx))]^F$$

---

[1] We use Church's notation $o\iota$, which stands for the functional type $\iota \to o$. The reader is referred to [1] for a more detailed introduction. In the remainder, $o$ will denote the type of truth values and $\iota$ defines the type of individuals. Other small Greek letters will denote arbitrary types. Thus, $X_{o\alpha}$ (resp. its $\eta$-long form $\lambda y_{o} \bullet Xy$) is actually a characteristic function denoting the set of elements of type $\alpha$ for which the predicate associated with $X$ holds. We use the square dot '$\bullet$' as an abbreviation for a pair of brackets, where '$\bullet$' stands for the left one with its partner as far to the right as is consistent with the bracketing already present in the formula.

This intermediate unit clause is not in normal form and subsequent normalization generates 12 clauses including the following four:

$$[Bx]^F \qquad\qquad [Bx]^T \vee [Cx]^T \qquad\qquad [Bx]^T \vee [Dx]^T \qquad\qquad [Cx]^F \vee [Dx]^F$$

This set is essentially of propositional logic character and trivially refutable by Leo. For the complete proof of the problem Leo needs less than one second (on a notebook with an Intel Pentium M processor 1.60GHz and 2 MB cache) and a total of 36 clauses is generated.

## 4 Leo is an Interactive Theorem Prover

Leo is an interactive theorem prover based on extensional higher-order resolution. The motivation for this is twofold. Firstly, the provided interaction features support interaction with the automation of this calculus. For this an automated proof attempt may be interrupted at any time and the system developer can then employ the interaction facilities to investigate the proof state and to perform some steps by hand. He may then again proceed with the automated proof search. Secondly, the interaction facilities can be employed for tutoring higher-order resolution in the classroom and they have in fact been employed for this purpose.

We illustrate below an interactive session with Leo within the ΩMEGA system. For this we assume that the ΩMEGA system has already been started. Interaction within ΩMEGA is supported in two ways. We may use the graphical user interface Loui, or ΩMEGA's older Emacs based command line interface. Both interfaces can be started and used simultaneously. In this section we describe interaction only within the Emacs based interface. All commands could alternatively also be invoked via Loui.

After starting ΩMEGA we are offered the following command line prompt in the Emacs interface:

```
OMEGA:
```

We assume that we have added the following problem, formalized in Post, to ΩMEGA's structured knowledge bass.

$$\exists Q_{o(o\iota)(o\iota)} \bullet p_{oo}((a_{o\iota}m_\iota) \wedge ((b_{o\iota}m_\iota) \vee (c_{o\iota}m_\iota))) \Rightarrow p_{oo}((a_{o\iota}m_\iota) \wedge (Q_{o(o\iota)(o\iota)}c_{o\iota}b_{o\iota}))$$

The Post representation of this problem is

```
(th~defproblem little
        (in base)
        (constants (p (o o)) (a (o i)) (b (o i)) (c (o i)) (m i))
        (conclusion
         (EXISTS (lam (Q (o (o i) (o i)))
                  (IMPLIES (p (AND (a m) (OR (b m) (c m))))
                           (p (AND (a m) (Q c b)))))))))
```

All problems have names and the name chosen for our problem is `little`. The **in**-slot in this problem definition specifies MBASE theories from which further knowledge, e.g., definitions and lemmata, is inherited. Examples of some MBASE theories are `typed-set`, `relation`, `function`, and `group`. Our very simple example problem is defined in the knowledge repository for theory `base`, and no further information is included. The assertion we want to prove is specified in the **conclusion**-slot and the typed constant symbols which occur in the assertion are declared in the **constants**-slot.

We now load all problems defined in theory `base` and then call the $\Omega$MEGA command `show-problems` to display the names of all problems. '[...]' indicates that some less interesting output information from LEO has been deleted here for presentation purposes.

```
OMEGA: load-theory-problems base
[...]

OMEGA: show-problems
[...]
EMBEDDED
little
less-little
TEST
[...]
```

Next we initialize the $\Omega$MEGA proof assistant with the proof problem `little` and display $\Omega$MEGA's central proof object after initialization.

```
OMEGA: prove little
Changing to proof plan LITTLE-1

OMEGA: show-pds
              ...
LITTLE ()         ! (EXISTS [Q:(O (O I) (O I))]              OPEN
                            (IMPLIES
                             (P (AND (A M) (OR (B M) (C M))))
                             (P (AND (A M) (Q C B)))))

OMEGA:
```

Then we initialize LEO with this problem. Here we choose the default settings (suggested in the '[]'-brackets) as input parameters for LEO. Each tactic (here EXT-INPUT-RECURSIVE) determines a specific flag setting of LEO. This flag setting is displayed after initialization.

```
OMEGA: leo-initialize
NODE (NDLINE) Node to prove with LEO: [LITTLE]
TACTIC (STRING) The tactic to be used by LEO: [EXT-INPUT-RECURSIVE]
THEORY-LIST (SYMBOL-LIST) Theories whose definitions will be expanded: [()]
Expanding the Definitions....
Initializing LEO....
```

```
Applying Clause Normalization....
============== variable settings ================
  Value(LEO*F-FO-ATP-RESOURCE) = 0
  Value(LEO*F-COOPERATE-WITH-FO-ATP) = NIL
  Value(LEO*F-TACTIC) = EXT-INPUT-RECURSIVE
  Value(LEO*F-VERBOSE-HALF) = NIL
  Value(LEO*F-VERBOSE) = NIL
  Value(LEO*F-WEIGHT-AGE-INT) = 4
  Value(LEO*F-SOS-TYPE) = TOSET
  Value(LEO*F-USABLE-TYPE) = INDEX
  Value(LEO*F-CLAUSE-LENGTH-RESTRICTION) = NIL
  Value(LEO*F-SAVE-FO-CLAUSES) = T
  Value(LEO*F-SUBSUM-MATCH-RESSOURCE) = NIL
  Value(LEO*F-SOS-SUBSUMTION) = NIL
  Value(LEO*F-BACKWARD-SUBSUMTION) = T
  Value(LEO*F-FORWARD-SUBSUMTION) = T
  Value(LEO*F-PARAMODULATION) = NIL
  Value(LEO*F-REMOVE=EQUIV-NEG) = NIL
  Value(LEO*F-REMOVE=EQUIV-POS) = NIL
  Value(LEO*F-REMOVE=LEIBNIZ-NEG) = NIL
  Value(LEO*F-REMOVE=LEIBNIZ-POS) = NIL
  Value(LEO*F-EXT-DECOMPOSE-ONLY) = T
  Value(LEO*F-EXT-UNICONSTRCLS-ONLY) = NIL
  Value(LEO*F-EXTENSIONALITY-NUM) = 6
  Value(LEO*F-EXT-INPUT-TREATMENT-RECURSIVE) = T
  Value(LEO*F-EXT-INPUT-TREATMENT) = NIL
  Value(LEO*F-EXTENSIONALITY) = T
  Value(LEO*F-NO-FLEX-UNI) = NIL
  Value(LEO*F-UNI-RESSOURCE) = 5
  Value(LEO*F-PRIM-SUBST-TYPES) = NIL
  Value(LEO*F-PRIMITIVE-SUBSTITUTION) = T
  Value(LEO*F-FACTORIZE) = T
  Value(LEO*F-AUTO-PROOF) = NIL
  Value(LEO*F-MAIN-COUNTER) = 0
========== end variable settings =============

OMEGA:
```

During initialization LEO first recursively expands defined symbols occurring in the assumptions or the assertion with respect to the specified theories (here we have none). Then it negates the assertion, turns it into a negated unit input clause and subsequently normalizes it with its internalized normalization rules. The resulting clauses are put either into LEO's set of support (if they stem from the assertion) or the usable set (if they stem from assumptions; here we have none). LEO clauses have unique names starting with `cl` and followed by an automatically created number. In '( )'-brackets further clause specific information follows and the '{ }'-brackets contain the clause literals. Negative literals have a leading `-` and positive literals a leading `+`. Type information is usually not displayed. All symbols starting with `dc` are free variables.

```
OMEGA: show-clauses
    ================= BEGIN ==============================
The set of support
    cl3(1.5|1):{(-(p (and (a m) (dc2 c b))))}
    cl4(1.5|1):{(+(p (and (a m) (or (b m) (c m)))))}
The set of usable clauses
    ================= END ================================
```

LEO offers a list of commands, for instance, to apply calculus rules, to manipulate and maintain the proof state, or to display information.

```
OMEGA: show-commands leo-interactive
[...]
DECOMPOSE:              Applies decomposition on a clause.
DELETE-CLAUSE:          Deletes a clause from the current environment.
END-REPORT:             Closes the report stream.
EXECUTE-LOG:            Reads a log file stepwise and eventually stores
                        some of its commands in a new log file.
EXIT:                   Leave the current command interpreter top level.
EXT:                    Applies extensionality rule on a clause.
[...]
FACTORIZE:              Applies factorization rule on a clause.
[...]
GUI-PROOF:              Displays the LEO proof of node in the GUI.
[...]
LEO-PROVE:              Prove with default parameter-settings.
NEW-LOG:                Sets the log mode and the log file name to the
                        given path name.
PARA:                   Applies paramodulation rule on two clauses.
PRE-UNIFIERS:           Computes the pre-unifiers of a clause.
PRE-UNIFY:              Applies pre-unification on a clause.
PRIM-SUBST:             Applies primitive substitution rule on a clause.
[...]
READ-LEO-PROBLEM:       Read a file, which contains a POST
                        representation of a problem, and transforms this
                        problem into clause normal form.
[...]
RESOLVE:                Applies resolution rule on two clauses.
SAVE-CLAUSE:            Save a clause for use after termination of LEO
                        under its clausename.
[...]
SET-FLAG:               Sets a global flag
SET-TACTIC:             Sets the tactic.
SHOW-CLAUSE:            Displays a clause, determined by name.
SHOW-CLAUSES:           Displays the two clause sets: the set of support
                        (LEO*G-SOS) and the set of usable clauses
                        (LEO*G-USABLE).
[...]
PROJECT:                Applies projection rule on a clause.
SHOW-DERIVATION:        Displays a linearized derivation of a the clause.
SHOW-FLAGS:             Shows the global flags
SHOW-LEO-PROBLEM:       Displays the given problem in POST.
SHOW-LEO-PROOF:         Displays the linearized LEO proof
SHOW-TACTICS:           Shows the tactics.
SHOW-VARS:              Shows the global vars.
START-REPORT:           Opens the tex and html report streams.
STEP-LOG:               Reads a log file stepwise and eventually stores
                        some of its commands in a new log file.
SUBSUMES:               Determines whether a clause subsumes another clause.
[...]
WRITE-DERIVATION:       Writes the derivation of a clause in a file.
WRITE-LOUIDERIVATION:   Writes the derivation of a clause in LOUI format
                        in a file.
WRITE-LOUIPROOF:        Writes the proof in LOUI format in a file.
WRITE-PROOF:            Writes the proof in a file.

OMEGA:
```

We apply the resolution rule to the clause `c13` and `c14` on literal positions 1 and 1 respectively. This results in the clause `c15` which consists only of a unification constraint. In this display unification constraints are presented as negated equations (on the datastructure level they are distinguished from them as already mentioned in Section 3). Provided that we can solve the unification constraint `c15`, we have found an empty clause and we are done.

```
OMEGA: resolve
```

```
NAME1 (STRING) of a clause: cl3
NAME2 (STRING) of a clause: cl4
POSITION1 (INTEGER) in clause 1: 1
POSITION2 (INTEGER) in clause 2: 1
 Clause 1: cl3(1.5|1):{(-(p (and (a m) (dc2 c b))))}.
 Clause 2: cl4(1.5|1):{(+(p (and (a m) (or (b m) (c m)))))}.
 Res(CL3[1],CL4[1]):
    (cl5(1|2):{(-(= (p (and (a m) (dc37 c b)))
                    (p (and (a m) (or (b m) (c m))))))}).

OMEGA:
```

We ask LEO to compute the pre-unifiers for this unification constraint.

```
OMEGA: pre-unifiers
NAME (STRING) of a clause: cl5
 The clause: cl5(1|2):{(-(= (p (and (a m) (dc37 c b)))
                           (p (and (a m) (or (b m) (c m))))))}.
 pre-unifiers: ({(dc37 --> [lam ?h45 ?h46.(or (?h46 m) (?h45 m))])}
                {(dc37 --> [lam ?h45 ?h46.(or (?h46 m) (c m))])}
                {(dc37 --> [lam ?h45 ?h46.(or (b m) (?h45 m))])}
                {(dc37 --> [lam ?h45 ?h46.(or (b m) (c m))])}).

OMEGA:
```

Pre-unification of clause `cl5` first generates these four pre-unifiers and then subsequently applies them to `cl5`. Instantiation of different pre-unifiers usually leads to different result clauses. In our simple case here, however, we obtain four copies of the empty clause.

```
OMEGA: pre-unify cl5
 The clause: cl5(1|2):{(-(= (p (and (a m) (dc37 c b)))
                           (p (and (a m) (or (b m) (c m))))))}.
 Result: (cl14(0|3):{NIL} cl15(0|3):{NIL} cl16(0|3):{NIL} cl17(0|3):{NIL}).

OMEGA:
```

We now display the derivation of clause `cl14` in the Emacs interface. `cl14` is the clause we obtain by application of the first pre-unifier from above. The employed pre-unifier (see clause `cl17`) is displayed in non-idempotent form. After applying pre-unifiers LEO subsequently normalizes the resulting clauses. This is why we have this superfluous looking normalization step from `cl17` to `cl14`.

```
OMEGA: show-derivation cl14
================= clauses =================
=============
Clause cl2 is #<Justified by ((Input))> :
 cl2(2O|0):{(+(not (exists [lam dc-20735.
  (implies (p (and (a m) (or (b m) (c m))))
           (p (and (a m) (dc-20735 c b))))]))))}
=============
================= proof =================
Clause cl2 is #<Justified by ((Input))> :
 cl2(2O|0):{(+(not (exists [lam dc-20735.
  (implies (p (and (a m) (or (b m) (c m))))
           (p (and (a m) (dc-20735 c b))))]))))}
=============
Clause cl3 is #<Justified by ((CNF)) on (cl2)> :
 cl3(1.5|1):{(-(p (and (a m) (dc-20738 c b))))}
=============
Clause cl2 is #<Justified by ((Input))> :
 cl2(2O|0):{(+(not (exists [lam dc-20735.
```

```
    (implies (p (and (a m) (or (b m) (c m))))
             (p (and (a m) (dc-20735 c b))))])))}
 ============
Clause cl4 is #<Justified by ((CNF)) on (cl2)> :
 cl4(1.5|1):{(+(p (and (a m) (or (b m) (c m)))))}
 ============
Clause cl5 is #<Justified by
 ((RES 1 1) (RENAMING {(dc2 --> dc37)})) on (cl3 cl4)> :
 cl5(1|2):{(-(= (p (and (a m) (dc37 c b)))
               (p (and (a m) (or (b m) (c m))))))}
 ============
Clause cl7 is #<Justified by
 ((UNI {(?h113 --> [lam ?h122 ?h123.m])
        (?h107 --> [lam ?h120 ?h121.m])
        (?h101 --> [lam ?h111 ?h112.(?h111 (?h113 ?h111 ?h112))])
        (?h100 --> [lam ?h105 ?h106.(?h107 (?h105 ?h105 ?h106))])
        (dc37 --> [lam ?h98 ?h99.(or (?h100 ?h98 ?h99)
                                     (?h101 ?h98 ?h99))])})
  (RENAMING {})) on (cl5)> :
 cl7(0|2):{NIL}
 ============
Clause cl14 is #<Justified by ((CNF)) on (cl7)> :
 cl14(0|3):{NIL}
 ============
 ========== clauses in proof: 7 ============

OMEGA:
```

We now slightly modify our example problem and obtain a much harder one.

$$\exists Q_{o(o\iota)(o\iota)}\!\cdot\! p_{oo}((a_{o\iota}m_{\iota}) \wedge ((b_{o\iota}m_{\iota}) \vee (c_{o\iota}m_{\iota}))) \Rightarrow p_{oo}((Q_{o(o\iota)(o\iota)}c_{o\iota}b_{o\iota}) \wedge (a_{o\iota}m_{\iota}))$$

In Post this problem is represented as

```
(th~defproblem less-little
        (in base)
        (constants (p (o o)) (a (o i)) (b (o i)) (c (o i)) (m i))
        (conclusion
         (EXISTS (lam (Q (o (o i) (o i)))
                 (IMPLIES (p (AND (a m) (OR (b m) (c m))))
                          (p (AND (Q c b) (a m))))))))
```

While problem `little` can still be solved by simple higher-order to first-order transformational approaches, this is not easily the case for the modified problem `less-little` since extensionality reasoning is required. The syntactic difference to `little`, however, is small. We only switched the two inner conjuncts in the right hand side of the implication. We now first load the problem, then initialize Leo and then display Leo's initial proof state.

```
OMEGA: prove less-little
Changing to proof plan LESS-LITTLE-1

OMEGA: show-pds
                ...
LESS-LITTLE ()          ! (EXISTS [Q:(O (O I) (O I))]          OPEN
                          (IMPLIES
                           (P (AND (A M) (OR (B M) (C M))))
                           (P (AND (Q C B) (A M)))))

OMEGA: leo-initialize
NODE (NDLINE) Node to prove with LEO: [LESS-LITTLE]
TACTIC (STRING) The tactic to be used by LEO: [EXT-INPUT-RECURSIVE]
```

```
THEORY-LIST (SYMBOL-LIST) Theories whose definitions will be expanded: [()]
Expanding the Definitions....
[...]

OMEGA: show-clauses
      ================= BEGIN =============================
The set of support
     cl3(1.5|1):{(-(p (and (dc-304 c b) (a m))))}
     cl4(1.5|1):{(+(p (and (a m) (or (b m) (c m)))))}
The set of usable clauses
      ================= END ===============================

OMEGA:
```

Again we resolve between clauses `cl3` and `cl4`.

```
OMEGA: resolve cl3 cl4 1 1
Clause 1: cl3(1.5|1):{(-(p (and (dc-304 c b) (a m))))}.
Clause 2: cl4(1.5|1):{(+(p (and (a m) (or (b m) (c m)))))}.
Res(CL3[1],CL4[1]): cl5(1|2):{(-(= (p (and (dc1573 c b) (a m)))
                                   (p (and (a m) (or (b m) (c m)))))))}.

OMEGA:
```

Now the resulting clause `cl5` is not pre-unifiable. Its unification constraint
has no 'syntactic' solution.

```
OMEGA: pre-unifiers cl5
The clause: cl5(1|2):{(-(= (p (and (dc1573 c b) (a m)))
                           (p (and (a m) (or (b m) (c m)))))))}.

pre-unifiers: NIL.

OMEGA: pre-unify cl5
The clause: cl5(1|2):{(-(= (p (and (dc1573 c b) (a m)))
                           (p (and (a m) (or (b m) (c m)))))))}.

Result: NIL.

OMEGA:
```

However, there is a semantic solution which we find by application of LEO's
combined extensionality treatment. This first decomposes the unification con-
straint and then applies Boolean extensionality (beforehand it usually tries to
exhaustively apply functional extensionality, which is not applicable here).

```
OMEGA: decompose cl5
The clause: cl5(1|2):{(-(= (p (and (dc1500 c b) (a m)))
                           (p (and (a m) (or (b m) (c m)))))))}.
Decomposed: (cl6(1|3):{(-(= (and (dc1500 c b) (a m))
                            (and (a m) (or (b m) (c m))))))}).

OMEGA: ext cl6
The clause: cl6(1|3):{(-(= (and (dc1500 c b) (a m))
                           (and (a m) (or (b m) (c m)))))}
Result: (cl14(4.5|4):{(+(a m) +(dc-4682 c b))}
         cl13(6.0|4):{(+(c m) +(b m) +(dc-4681 c b))}
         cl12(1.5|4):{(+(a m))}
         cl11(4.5|4):{(+(c m) +(b m) +(a m))}
         cl10(6.0|4):{(-(b m) -(a m) -(dc-4680 c b))}
         cl9(6.0|4):{(-(c m) -(a m) -(dc-4679 c b))}).

OMEGA:
```

This set of resulting (first-order like) clauses is refutable and LEO can find a
refutation. Unfortunately LEO is very bad at first-order reasoning (since it does
```



not employ optimizations and implementation tricks that are well known in the first-order community) and therefore the refutation of this clause set is not very efficient. This motivates a cooperation with first-order provers; see Sections 5 and 7 for further details.

## 5  LEO is an Automated Theorem Prover

LEO is first and foremost an automated theorem prover for classical higher-order logic. Within $\Omega$MEGA it can be applied to prove (sub)problems automatically. Below we first reinitialize $\Omega$MEGA with the problem `less-little` and then call LEO in its standard flag setting to it.

```
OMEGA: prove less-little
Changing to proof plan LESS-LITTLE-7

OMEGA: show-pds
              ...
LESS-LITTLE ()            ! (EXISTS [Q:(O (O I) (O I))]            OPEN
                             (IMPLIES
                               (P (AND (A M) (OR (B M) (C M))))
                               (P (AND (Q C B) (A M)))))

OMEGA: call-leo-on-node
NODE (NDLINE) Node to prove with LEO: [LESS-LITTLE]
TACTIC (STRING) The tactic to be used by LEO: [EXT-INPUT-RECURSIVE]
SUPPORTS (NDLINE-LIST) The support nodes: [()]
TIME-BOUND (INTEGER) Time bound for proof attempt: [100]
THEORY-LIST (SYMBOL-LIST) Theories whose definitions will be
expanded: [(ALL)]
DEFS-LIST (SYMBOL-LIST) Symbols whose definitions will not be
expanded: [(= DEFINED EQUIV)]
INSERT-FLAG (BOOLEAN) A flag indicating whether a partial result will
be automatically inserted.: [()]
Looking for expandable definitions ....
Initializing LEO....
Applying Clause Normalization....
[...]
Start proving ...
Loop: (100sec left)
 #1 (SOS 2 USABLE 0 EXT-QUEUE 0 FO-LIKE 4)
(99sec left)
 #2 (SOS 1 USABLE 1 EXT-QUEUE 0 FO-LIKE 4)
[...]
(94sec left)
 #29 (SOS 72 USABLE 12 EXT-QUEUE 0 FO-LIKE 98)
(91sec left)
 #30 (SOS 111 USABLE 13 EXT-QUEUE 0 FO-LIKE 138)

Total LEO time: 8446
**** proof found (next clause nr: cl599) *****
; cpu time (non-gc) 7,600 msec user, 0 msec system
; cpu time (gc)     750 msec user, 10 msec system
; cpu time (total)  8,350 msec user, 10 msec system
; real time  8,451 msec
; space allocation:
;  8,909,213 cons cells, 3,568 symbols, 122,908,112 other bytes,
; 0 static bytes

OMEGA:
```



## LEO's Architecture and Main Loop

LEO's basic architecture adapts the set of support approach. The four corner-stones of LEO's architecture (see Fig. 1) are:

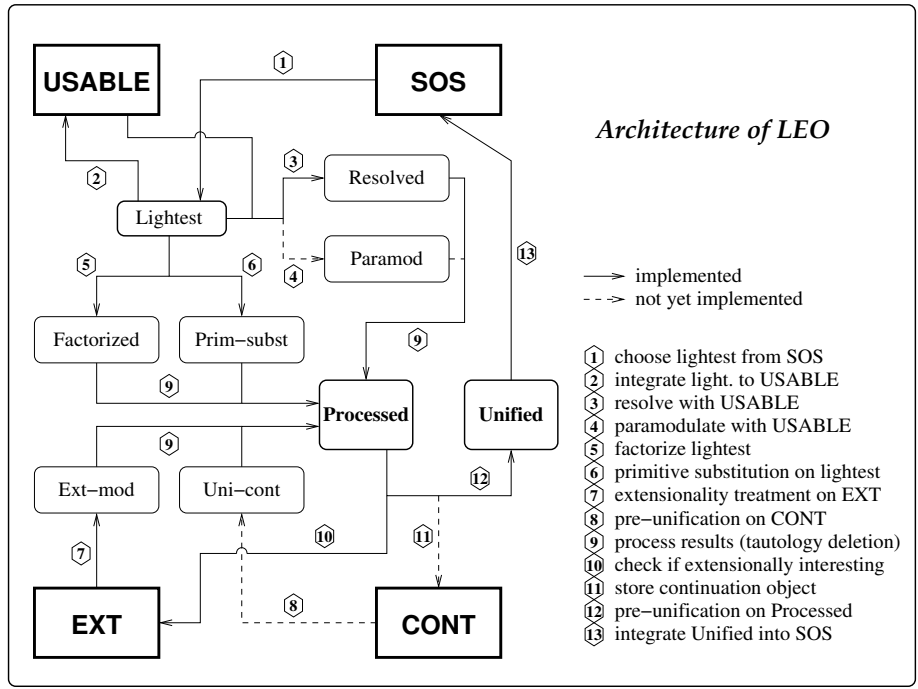

**Fig. 1.** LEO's main architecture (the 'dotted lines' indicates functionalities which are usually disabled or not fully available yet)

**USABLE** The set of all usable clauses, which initially only contains clauses that are assumed to be satisfiable, i.e., the clauses stemming from the assumptions of the theorem to prove.

**SOS** The set of support, which initially only contains the clauses belonging to the negated assertion.

**EXT** The set of all extensionally interesting clauses, i.e., heuristically determined clauses which are stored for extensionality treatment. Initially this set is empty. See Step 12 of the main loop description below.

**CONT** The set of all continuations created by the higher-order pre-unification algorithm when reaching the pre-unification search depth limit. The idea is to support continuations of interrupted pre-unification attempts at a later time. (This store is not activated yet in LEO)



LEO's main loop (see Fig. 1) consists of the steps 1–13 as described below (the **Initialize** step is applied only at the very beginning of the proof attempt and is not part of the main loop). This loop, whose data-flow is graphically illustrated in Figure 1, is executed until an empty clause, i.e., a clause consisting no literals or only of flex-flex-unification constraints[2], is detected.

LEO employs a higher-order subsumption test that is, apart from the technical details, very similar to the ones employed in first-order provers. Instead of first-order matching, the criteria for comparing the single literals is higher-order simplification matching, i.e., matching with respect to the deterministic higher-order simplification rules. It is theoretically possible to develop and employ a much stronger subsumption filter in LEO. This is future work. However, a perfect extensional higher-order subsumption filter, which would be the ideal case, is not feasible since extensional higher-order unification is undecidable.

**Initialize** The specified assumptions and the assertion are pre-clausified, i.e., the assumptions become positive unit (pre-)clauses and the assertion becomes a negative unit (pre-)clause. Definitions are expanded (with respect to the specified theories) and the pre-clauses are normalized. Within the clause normalization process the positive primitive equations are usually replaced by respective Leibniz equations. Negative primitive equations are not expanded but immediately encoded as extensional unification constraints. Furthermore, identical literals are automatically factorized. The assumption clauses are passed to `USABLE` and the assertion clause to `SOS`.

**Step 1 (Choose Lightest)** LEO chooses the lightest (wrt. to a clause ordering), i.e., topmost, clause from SOS. If this clause is a pre-clause, i.e., not in proper clause normal form, then LEO applies clause normalization to it and integrates the resulting proper clauses into `SOS`. Depending on the flag-setting forward and/or backward subsumption is applied; see also step 13).

**Step 2 (Insert to `USABLE`)** LEO inserts the lightest clause into USABLE while employing forward and/or backward subsumption depending on LEO's overall flag-setting.

**Step 3 (Resolve)** The lightest clause is resolved against all clauses in `USABLE` and the results are stored in `RESOLVED`.

**Step 4 (Paramodulate)** Paramodulation is applied between all clauses in `USABLE` and the lightest clause, and the results are stored in `PARAMOD`. (This step is currently not activated in LEO; currently Leibniz equality is globally employed instead of primitive equality.)

**Step 5 (Factorize)** The lightest clause is factorized and the resulting clauses are stored in `FACTORIZED`.

**Step 6 (Primitive Substitution)** LEO applies the primitive substitution principle to the lightest clause. The particular logical connectives to be imitated in this step are specified by a flag. The resulting clauses are stored in `PRIM-SUBST`.

---

[2] A flex-flex-unification constraint has topmost free variables in each of the two terms to be unified. Flex-flex-constraints are always solvable.



**Step 7 (Extensionality Treatment)** The heuristically sorted store `EXT` contains extensionally interesting clauses (i.e., clauses with unification constraints that may have additional pre-unifiers, if the extensionality rules are taken into account). LEO chooses the topmost clause and applies the compound extensionality treatment to all extensionally interesting literals.

**Step 8 (Continue Unification)** The heuristically sorted store `CONT` contains continuations of interrupted higher-order pre-unification attempts from the previous loops (cf. step 10). If the actual unification search depth limit (specified by a flag, whose value can be dynamically increased during proof attempts) allows for a deeper search in the current loop, then the additional search for unifiers will be performed. The resulting instantiated clauses are passed to `UNI-CONT` and the new continuations are sorted and integrated into `CONT`. (This step is currently not activated in LEO)

**Step 9 (Collect Results)** In this step LEO collects all clauses that have been generated within the current loop from the stores `RESOLVED`, `PARAMOD`, `FACTORIZED`, `PRIM-SUBST`, `EXT-MOD`, and `UNI-CONT`, eliminates obvious tautologies, and stores the remaining clauses in `PROCESSED`.

**Step 10 (Pre-Unify)** LEO tries to pre-unify the clauses in `PROCESSED`. Thus, it applies the pre-unification rules exhaustively, thereby spanning a unification tree until it reaches the unification search depth limit specified by a special flag. The unification search depth limit specifies how many subsequent flex-rigid-branchings may at most occur in each path through the unification search tree. The pre-unified, i.e., instantiated, clauses are passed to `UNIFIED`. The main idea of this step is to filter out all those clauses with syntactically non-solvable unification constraints (modulo the allowed search depth limit). But note that there are exceptions, which are determined in steps 11 and 12. That means that not all syntactically non-unifiable clauses are removed from the search space as this would, e.g., also remove the extensionally interesting clauses.

**Step 11 (Store Continuations)** Each time a pre-unification attempt in step 10 is interrupted by reaching the unification search depth limit, a respective continuation is created. This object stores the state of the interrupted unification search process, i.e., it contains the particular unification constraints as given at the point of interruption together with the remaining literals of the clause in focus and some information on the interrupted unification process. Continuations allow the prover to continue the interrupted unification process at any later time. The set of all such continuations is integrated in the sorted store `CONT`. (This step is not activated yet in LEO.)

**Step 12 (Store Extensionally Interesting Clauses)** In the pre-unification process in step 10 LEO analyzes the unification pairs in focus in order to estimate whether this unification constraint and thus this clause is extensionally interesting, i.e., probably solvable with respect to both extensionality principles. All extensionally interesting clauses are passed to `EXT`, which is heuristically sorted. While inserting the clauses into `EXT` forward and/or backward subsumption is applied in order to minimize the number of clauses in this store.



**Step 13 (Integrate to SOS)** In the last step Leo integrates all pre-unified clauses in `UNIFIED` into the sorted store `SOS`. Forward and/or backward subsumption is employed depending on the flag-setting.

## 6 Visualizing Leo Derivations in Loui

$\Omega$mega's graphical user interface Loui [24] usually displays $\Omega$mega proof objects in multiple modalities: a graphical map of the proof tree, a linearized presentation of the proof nodes with their formulae and justifications, and a term browser. Display of type information is optional and is determined by a switch in Loui.

Loui can be used to display proofs of other systems as well. Leo's connection to Loui supports the graphical presentation of extensional higher-order resolution derivations and proofs with the gain for the user that the structure of interactively or automatically created derivations becomes more transparent as is possible in the linearized display shown in Section 4.

Display of external proofs is supported by Loui via a specific interface language. In order to display Leo's resolution proofs a simple mapping of the internal proof state in Leo into this interface language is required. Fig. 2 illustrates the Loui visualization of the automatically generated Leo proof for problem `less-little` from Section 5, and Fig. 3 displays part of the respective interface language representation of this proof.

## 7 Experiments with Leo

Leo has successfully been applied to different higher-order examples. For example, in [4] Leo's performance on simple examples about sets has been investigated. One example is the already addressed TPTP problem SET171+3, i.e., distributivity of union over intersection. Despite their simplicity such examples are often non-trivial for automated first-order theorem provers. More details on this discussion can be found in [9]. Further, proof examples have been investigated in [5].

In [9] a cooperation of Leo with a first-order theorem prover (we used the automated theorem prover Bliksem [11] since this was already well integrated in the OMEGA framework) has been proposed and investigated. Thus, Leo has been slightly extended so that it now constantly accumulates a bag of first-order like clause. First-order like clauses do not contain any 'real' higher-order subterms (such as a $\lambda$-abstraction or embedded equations), and are therefore suitable for treatment by a first-order ATP or even a propositional logic decision procedure after appropriate transformation. We use the transformation mapping as also employed in Tramp [18], which has been previously shown to be sound and is based on [17]. Essentially, it injectively maps expressions such as $P(f(a))$ to expressions such as $@^1_{\text{pred}}(P, @^1_{\text{fun}}(f, a))$, where the $@$ are new first-order operators describing function and predicate application for particular types and arities.



**Fig. 2.** Loui usually displays $\Omega$MEGA proof plans and $\Omega$MEGA natural deduction proofs. In addition it can be employed to display Leo's resolution proofs. Here we display the derivation of clause `cl7` from example problem `less-little`.



```
[...]
 insertNode(grounded none "cl3" ["cl3""cl2"]
  "([NOT (O O)] ([P (O O)] ([AND (O O O)] ([DC-11438 (O (O I) (O I))]
   [C (O I)] [B (O I)]) ([A (O I)] [M I]))))"
  "((CNF))" ["cl2"] false)
 insertNode(grounded none "cl4" ["cl4""cl2"]
  "([P (O O)] ([AND (O O O)] ([A (O I)] [M I]) ([OR (O O O)] ([B (O I)]
   [M I]) ([C (O I)] [M I]))))"
  "((CNF))" ["cl2"] false)

 insertNode(grounded none "cl5" ["cl5""cl4""cl3""cl2"]
  "([NOT (O O)] ([= (O O O)] ([P (O O)] ([AND (O O O)] ([A (O I)] [M I])
   ([OR (O O O)] ([B (O I)] [M I]) ([C (O I)] [M I])))) ([P (O O)]
   ([AND (O O O)] ([dc2762 (O (O I) (O I))] [C (O I)] [B (O I)])
   ([A (O I)] [M I])))))"
  "((RES 1 1) (RENAMING {(dc2705 --> dc2762)}))" ["cl4""cl3"] false)
 insertNode(grounded none "cl6" ["cl6""cl5""cl4""cl3""cl2"]
  "([NOT (O O)] ([= (O O O)] ([AND (O O O)] ([A (O I)] [M I]) ([OR (O O O)]
   ([B (O I)] [M I]) ([C (O I)] [M I]))) ([AND (O O O)] ([dc2762 (O (O I)
   (O I))] [C (O I)] [B (O I)]) ([A (O I)] [M I])))))"
  "((DEC (1)))" ["cl5"] false)
[...]
```

**Fig. 3.** Part of the Loui interface language representation of the Leo proof for problem `less-little` as communicated to Loui.

The injectivity of the mapping guarantees soundness, since it allows each proof step to be mapped back from first-order to higher-order.

Whenever Leo creates a new clause it checks whether this is a first-order like clause, i.e., whether it is in the domain of the employed transformational mapping. If this is the case, a copy of it is passed to the store of first-order like clauses. In each loop of Leo's search procedure a fast first-order prover can now be applied to the set of first-order like clauses to find a refutation. In this case an overall proof has been found. The experiments in [9] show that this is a very promising approach to combining the benefits of higher-order and first-order theorem provers. Whereas this cooperative approach can solve the problem `less-little` only slightly faster than Leo alone, many examples in [9] show that there are often significant improvements possible.

```
OMEGA: prove less-little
Changing to proof plan LESS-LITTLE-10

OMEGA: call-leo-on-node
NODE (NDLINE) Node to prove with LEO: [LESS-LITTLE]
TACTIC (STRING) The tactic to be used
by LEO: [EXT-INPUT-RECURSIVE]fo-atp-cooperation
SUPPORTS (NDLINE-LIST) The support nodes: [()]
TIME-BOUND (INTEGER) Time bound for proof attempt: [100]
THEORY-LIST (SYMBOL-LIST) Theories whose definitions will be
expanded: [(ALL)]
DEFS-LIST (SYMBOL-LIST) Symbols whose definitions will not be
expanded: [(= DEFINED EQUIV)]
INSERT-FLAG (BOOLEAN) A flag indicating whether a partial result
will be automatically inserted.: [()]
Looking for expandable definitions ....
Initializing LEO....
Applying Clause Normalization....
[...]
```



```
Start proving ...
Loop: (100sec left)
 #1 (SOS 2 USABLE 0 EXT-QUEUE 0 FO-LIKE 4)
(99sec left)
 #2 (SOS 1 USABLE 1 EXT-QUEUE 0 FO-LIKE 4)
(98sec left)
[...]
(96sec left)
 #9 (SOS 3 USABLE 4 EXT-QUEUE 0 FO-LIKE 12)
[...]
 Calling bliksem process 22267 with time resource 50sec .
 PARSING BLIKSEM OUTPUT ...
 Bliksem has found a saturation.
[...]
(96sec left)
 #10 (SOS 10 USABLE 5 EXT-QUEUE 0 FO-LIKE 20)
[...]
(94sec left)
 #21 (SOS 39 USABLE 9 EXT-QUEUE 0 FO-LIKE 60)
 Calling bliksem process 22454 with time resource 50sec .
 bliksem Time Resource in seconds:
 PARSING BLIKSEM OUTPUT ...
 Bliksem has found a proof.
Bliksem's time:
; cpu time (non-gc) 0 msec user, 0 msec system
; cpu time (gc)     0 msec user, 0 msec system
; cpu time (total)  0 msec user, 0 msec system
; real time  174 msec
; space allocation:
;  416 cons cells, 0 symbols, 15,720 other bytes, 8936 static bytes
Input Clauses: 75
clauses generated:       37
(94sec left)
 #22 (SOS 53 USABLE 10 EXT-QUEUE 0 FO-LIKE 75)
Total LEO time: 7262
**** proof found (next clause nr: cl323) *****
; cpu time (non-gc) 5,650 msec user, 20 msec system
; cpu time (gc)     490 msec user, 0 msec system
; cpu time (total)  6,140 msec user, 20 msec system
; real time  7,273 msec
; space allocation:
;  5,218,155 cons cells, 1,886 symbols, 82,231,456 other bytes,
;  49536 static bytes
```

# 8  Related Work and Conclusion

There are only very few automated theorem provers available for higher-order logic. Tps [2, 3], which is based on the mating search method, is the oldest and probably still the strongest prover in this category. The extensionality reasoning of Tps has recently been significantly improved by Brown in his PhD thesis [10].

Related to the cooperation approach is the work of Hurd [16] which realizes a generic interface between higher-order logic and first-order theorem provers. It is similar to the solution previously achieved by Tramp [18] in Ωmega. Both approaches pass essentially first-order clauses to first-order theorem provers and then translate their results back into higher-order. More recent related work on the cooperation of Isabelle with the first-order theorem prover Vampire is presented in [19]. Further related work is Otter-λ (see http://mh215a.cs.sjsu.edu/), which extends first-order logic with λ-notation.



Leo has initially been implemented as a demonstrator system for extensional higher-order resolution in the context of the author's PhD thesis [4]. The experiments carried out with Leo so far, in particular, its recent combination with a fast first-order theorem prover, have been very promising, and they motivate further work in this direction. This is particularly true since interactive proof assistants based on higher-order logic are recently gaining increasing attention in formal methods.

During the implementation and later during the experiments many shortcomings of Leo have been identified by the author. These shortcomings are both of theoretical and of practical nature. Altogether this calls for a proper reimplementation of Leo. This reimplementation should ideally be independent of $\Omega$MEGA in order to provide a lean and easy to install and use automated higher-order theorem to the community.

## References


1. P. Andrews. *An Introduction to mathematical logic and Type Theory: To Truth through Proof.* Number 27 in Applied Logic Series. Kluwer, 2002.
2. P. B. Andrews, M. Bishop, and C. E. Brown. System description: TPS: A theorem proving system for type theory. In *Conference on Automated Deduction*, pages 164–169, 2000.
3. P.B. Andrews, M. Bishop, S. Issar, D. Nesmith, F. Pfenning, and H. Xi. TPS: A theorem proving system for classical type theory. *Journal of Automated Reasoning*, 16(3):321–353, 1996.
4. C. Benzmüller. *Equality and Extensionality in Higher-Order Theorem Proving.* PhD thesis, Universität des Saarlandes, Germany, 1999.
5. C. Benzmüller. Comparing approaches to resolution based higher-order theorem proving. *Synthese*, 133(1-2):203–235, 2002.
6. C. Benzmüller and M. Kohlhase. Extensional higher-order resolution. In *Proc. of CADE-15*, number 1421 in LNAI. Springer, 1998.
7. C. Benzmüller and M. Kohlhase. LEO – a higher-order theorem prover. In *Proc. of CADE-15*, number 1421 in LNAI. Springer, 1998.
8. C. Benzmüller and V. Sorge. Oants – An open approach at combining Interactive and Automated Theorem Proving. In *Proc. of Calculemus-2000.* AK Peters, 2001.
9. C. Benzmüller, V. Sorge, M. Jamnik, and M. Kerber. Can a higher-order and a first-order theorem prover cooperate? In F. Baader and A. Voronkov, editors, *Proceedings of the 11th International Conference on Logic for Programming Artificial Intelligence and Reasoning (LPAR)*, volume 3452 of *LNAI*, pages 415–431. Springer, 2005.
10. C. E. Brown. *Set Comprehension in Church's Type Theory.* PhD thesis, Dept. of Mathematical Sciences, Carnegie Mellon University, USA, 2004.
11. H. de Nivelle. *The Bliksem Theorem Prover, Version 1.12.* Max-Planck-Institut, Saarbrücken, Germany, 1999. http://www.mpi-sb.mpg.de/~bliksem/manual.ps.
12. M. Gordon and T. Melham. *Introduction to HOL – A theorem proving environment for higher order logic.* Cambridge University Press, 1993.
13. J. Harrison. The hol light theorem prover.
14. G.P. Huet. *Constrained Resolution: A Complete Method for Higher Order Logic.* PhD thesis, Case Western Reserve University, 1972.





15. G.P. Huet. A mechanization of type theory. In Donald E. Walker and Lewis Norton, editors, *Proc. of the 3rd International Joint Conference on Artificial Intelligence (IJCAI73)*, pages 139–146, 1973.

16. J. Hurd. An LCF-style interface between HOL and first-order logic. In *Automated Deduction — CADE-18*, volume 2392 of *LNAI*, pages 134–138. Springer, 2002.

17. M. Kerber. *On the Representation of Mathematical Concepts and their Translation into First Order Logic*. PhD thesis, Universität Kaiserslautern, Germany, 1992.

18. A. Meier. TRAMP: Transformation of Machine-Found Proofs into Natural Deduction Proofs at the Assertion Level. In *Proc. of CADE-17*, number 1831 in LNAI. Springer, 2000.

19. J. Meng and L. C. Paulson. Experiments on supporting interactive proof using resolution. In *Proc. of IJCAR 2004*, volume 3097 of *LNCS*, pages 372–384. Springer, 2004.

20. T. Nipkow, L.C. Paulson, and M. Wenzel. *Isabelle/HOL: A Proof Assistant for Higher-Order Logic*. Number 2283 in LNCS. Springer, 2002.

21. S. Owre, S. Rajan, J.M. Rushby, N. Shankar, and M. Srivas. PVS: Combining specification, proof checking, and model checking. In R. Alur and T. Henzinger, editors, *Computer-Aided Verification, CAV '96*, number 1102 in LNCS, pages 411–414, New Brunswick, NJ, 1996. Springer.

22. L. Paulson. *Isabelle: A Generic Theorem Prover*. Number 828 in LNCS. Springer, 1994.

23. J. Siekmann, C. Benzmüller, A. Fiedler, A. Meier, I. Normann, and M. Pollet. Proof development in OMEGA: The irrationality of square root of 2. In *Thirty Five Years of Automating Mathematics*. Kluwer, 2003.

24. J. Siekmann, S. Hess, C. Benzmüller, L. Cheikhrouhou, A. Fiedler, H. Horacek, M. Kohlhase, K. Konrad, A. Meier, E. Melis, M. Pollet, and V. Sorge. LOUI: Lovely OMEGA user interface. *Formal Aspects of Computing*, 11:326–342, 1999.

25. G.L. Steele. *Common Lisp: The Language, 2nd edition*. Digital Press, Bedford, Massachusetts, 1990.




# Combining Proofs of Higher-Order and First-Order Automated Theorem Provers


Christoph Benzmüller[1], Volker Sorge[2], Mateja Jamnik[3], and Manfred Kerber[2]

[1]Fachbereich Informatik, Universität des Saarlandes
66041 Saarbrücken, Germany (www.ags.uni-sb.de/~chris)
[2]School of Computer Science, The University of Birmingham
Birmingham B15 2TT, England, UK (www.cs.bham.ac.uk/~vxs)
[3]University of Cambridge Computer Laboratory
Cambridge CB3 0FD, England, UK (www.cl.cam.ac.uk/~mj201)



**Abstract.** $\Omega$ANTS is an agent-oriented environment for combining inference systems. A characteristics of the $\Omega$ANTS approach is that a common proof object is generated by the cooperating systems. This common proof object can be inspected by verification tools to validate the correctness of the proof. $\Omega$ANTS makes use of a two layered blackboard architecture, in which the applicability of inference rules are checked on one (abstract) layer. The lower layer administrates explicit proof objects in a common language. In concrete proofs these proof objects can be quite bit, which can make communication during proof search very inefficient. As a result we had situations in which most of the resources went into the overhead of constructing explicit proof objects and communicating between different components. Therefore we have recently developed an alternative modelling of cooperating systems in $\Omega$ANTS which allows direct communication between related systems during proof search. This has the consequence that proof objects can no longer be directly constructed and thus the correctness-validation in this novel approach is in question. In this paper we present a pragmatic approach how this can rectified.


## 1 Introduction

$\Omega$ANTS is an agent-oriented environment for combining inference rules and inference systems. $\Omega$ANTS was originally conceived to support interactive theorem proving but was later extended to a fully automated proving system [23, 8]. A characteristics of the $\Omega$ANTS approach is that a joint proof object is generated by the cooperating inference rules and inference systems. This joint proof object can be inspected by proof verification tools in combination with proof expansion in order to validate the correctness at a purely logic level. The $\Omega$ANTS blackboard architecture consists of two layers, an abstract upper layer, and a more detailed lower layer. Applicability criteria for inference rules are modelled at the upper layer. The upper layer is supported by computations at the lower layer which models criteria for the instantiation of the parameters of the inference rules.



External systems have been modelled in $\Omega$ANTS as individual inference rules at the upper layer. With this approach, $\Omega$ANTS has been successfully employed in past experiments to check the validity of set equations using higher-order and first-order theorem provers, model generation, and computer algebra [5]. However, this approach was very inefficient for hard examples because of the communication overhead imposed by the need to translate all steps into a common proof data structure.

Therefore, we have recently developed an alternative approach: *the single inference rule approach* of cooperating systems in $\Omega$ANTS which exploits the lower layer of the blackboard architecture. This approach has been successfully applied to the combination of automated higher-order and first-order theorem provers. In particular, it has outperformed state-of-the-art first-order specialist reasoners (including Vampire 7.0) on 45 examples on sets, relations and functions; see [9].

Unfortunately, using a single inference rule approach, we had to sacrifice the generation of joint proof objects and correctness validation in this novel approach. In this paper we present a pragmatic approach to how this can be rectified.

The paper is structured as follows: In Section 2 we motivate and illustrate the cooperation between a higher-order theorem prover (we employ LEO [6]) and a first-order theorem prover (we employ BLIKSEM [12]). In Section 3 we compare the two options for modelling cooperative reasoning systems in $\Omega$ANTS: the initial multiple inference approach and the novel single inference rule approach. In Section 4 we show how a joint proof object can also be obtained for the latter modelling by mapping it back to the former. Section 5 concludes the paper.

## 2 Combining Higher-Order and First-Order ATP

### 2.1 Motivation

When dealing with problems containing higher-order concepts, such as sets, functions, or relations, today's state-of-the-art automated theorem provers (ATPs) still exhibit weaknesses on problems considered relatively simple by humans (cf. [15]). One reason is that the problem formulations use an encoding in a first-order set theory, which makes it particularly challenging when trying to prove theorems from first principles, that is, basic axioms. Therefore, to aid ATPs in finding proofs, problems are often enriched by hand-picked additional lemmata, or axioms of the selected set theory are dropped leaving the theory incomplete. This has recently motivated extensions of state-of-the-art first-order calculi and systems, as for example presented in [15] for the SATURATE system. The extended SATURATE system can solve some problems from the SET domain in the TPTP [25] which VAMPIRE [22] and E-SETHEO's [24] cannot solve.

While it has already been shown in [6, 2] that many problems of this nature can be easily proved from first principles using a concise higher-order representation and the higher-order resolution ATP LEO [6], the combinatorial explosion inherent in LEO's calculus prevents the prover from solving a whole range of



| | |
|---|---|
| SET171+3 | $\forall X_{o\alpha}, Y_{o\alpha}, Z_{o\alpha}.X \cup (Y \cap Z) = (X \cup Y) \cap (X \cup Z)$ |
| SET611+3 | $\forall X_{o\alpha}, Y_{o\alpha}.(X \cap Y = \emptyset) \Leftrightarrow (X \setminus Y = X)$ |
| SET624+3 | $\forall X_{o\alpha}, Y_{o\alpha}, Z_{o\alpha}.\mathrm{Meets}(X, Y \cap Z) \Leftrightarrow \mathrm{Meets}(X, Y) \vee \mathrm{Meets}(X, Z)$ |
| SET646+3 | $\forall x_{\alpha}, y_{\beta}.\mathrm{Subrel}(\mathrm{Pair}(x, y), (\lambda u_{\alpha}.\top) \times (\lambda v_{\beta}.\top))$ |
| SET670+3 | $\forall Z_{o\alpha}, R_{o\beta\alpha}, X_{o\alpha}, Y_{o\beta}.\mathrm{IsRelOn}(R, X, Y) \Rightarrow$ |
| | $\mathrm{IsRelOn}(\mathrm{RestrictRDom}(R, Z), Z, Y)$ |

**Table 1.** Test Problems on Sets and Relations: Examples

possible problems with one universal strategy. Often higher-order problems require only relatively few but essential steps of higher-order reasoning, while the overwhelming part of the reasoning is first-order or even propositional level. This suggests that LEO's performance could be improved when combining it with a first-order ATP to search efficiently for a possible refutation in the subset of those clauses that are essentially first-order.

The advantages of such a combination are not only that many problems can still be efficiently shown from first principles in a general purpose approach, but also that problems can be expressed in a very concise way.

For instance, in [9] we present 45 problems from the SET domain of the TPTP-v3.0.1, together with their entire formalisation in less than two pages, which is difficult to achieve within a framework that does not provide $\lambda$-abstraction. We have used this problem set, which is an extension of the problems considered in [15], to show the effectiveness of our approach (cf. [9]). While many of the considered problems can be proved by LEO alone with some strategy, the combination of LEO with the first-order ATP BLIKSEM [12] is not only able to solve more problems, but also needs only a single strategy to prove them. Several of our problems are considered very challenging by the first-order community and five of them (of which LEO can solve four) have a TPTP rating of 1.00, saying that they cannot be solved by any TPTP prover to date.

Technically, the combination has been realised in the concurrent reasoning system $\Omega$ANTS [23, 8] which enables the cooperation of hybrid reasoning systems to construct a common proof object. In our past experiments, $\Omega$ANTS has been successfully employed to check the validity of set equations using higher-order and first-order ATPs, model generation, and computer algebra [5]. While this enabled a cooperation between LEO and a first-order ATP, the proposed solution could not be classified as a general purpose approach. A major shortcoming was that all communication of partial results had to be conducted via the common proof object, which was very inefficient for hard examples. Thus, the solved examples from set theory were considered too trivial, albeit they were often similar to those still considered challenging in the TPTP in the first-order context.

In [9] we have presented our novel approach to the cooperation between LEO and BLIKSEM inside $\Omega$ANTS by decentralising communication. As has been documented in [9] this leads not only to a higher overall efficiency but also to a general purpose approach based on a single strategy in LEO.



## 2.2 Sets, Relations, and Functions: Higher-Order Logic Encoding

We list some examples of the test problems on sets and relations (and functions) that have been investigated in [9]. These test problems are taken from the TPTP against which we evaluated our system. The problems are given by the identifiers used in the SET domain of the TPTP, and are formalized in a variant of Church's simply typed $\lambda$-calculus with prefix polymorphism. In classical type theory, terms and all their sub-terms are typed. Polymorphism allows the introduction of type variables such that statements can be made for all types. For instance, in problem SET171 in Table 1, the universally quantified variable $X_{o\alpha}$ denotes a mapping from objects of type $\alpha$ to objects of type $o$. We use Church's notation $o\alpha$, which stands for the functional type $\alpha \to o$. The reader is referred to [1] for a more detailed introduction. In the remainder, $o$ will denote the type of truth values, and small Greek letters will denote arbitrary types. Thus, $X_{o\alpha}$ (and its $\eta$-longform $\lambda y_{o\bullet}.Xy$) is actually a characteristic function denoting the set of elements of type $\alpha$, for which the predicate associated with $X$ holds. As further notational convention, we use capital letter variables to denote sets, functions, or relations, while lower case letters denote individuals. Types are usually only given in the first occurrence of a variable and omitted if inferable from the context. Table 1 presents some examples of the test problems investigated in [9].

These test problems employ defined concepts that are specified in a knowledge base of hierarchical theories that Leo has access to. Table 2 gives the concepts necessary for defining the above problems:

| | |
|---|---|
| $\_ \in \_$ | $:= \lambda x_\alpha, A_{o\alpha}.[Ax]$ |
| $\emptyset$ | $:= [\lambda x_\alpha.\bot]$ |
| $\_ \cap \_$ | $:= \lambda A_{o\alpha}, B_{o\alpha}.[\lambda x_\alpha.x \in A \wedge x \in B]$ |
| $\_ \cup \_$ | $:= \lambda A_{o\alpha}, B_{o\alpha}.[\lambda x_\alpha.x \in A \vee x \in B]$ |
| $\_ \setminus \_$ | $:= \lambda A_{o\alpha}, B_{o\alpha}.[\lambda x_\alpha.x \in A \vee x \notin B]$ |
| $\text{Meets}(\_,\_)$ | $:= \lambda A_{o\alpha}, B_{o\alpha}.[\exists x_\alpha.x \in A \wedge x \in B]$ |
| $\text{Pair}(\_,\_)$ | $:= \lambda x_\alpha, y_\beta.[\lambda u_\alpha, v_\beta.u = x \vee v = y]$ |
| $\_ \times \_$ | $:= \lambda A_{o\alpha}, B_{o\beta}.[\lambda u_\alpha, v_\beta.u \in A \wedge v \in B]$ |
| $\text{Subrel}(\_,\_)$ | $:= \lambda R_{o\beta\alpha}, Q_{o\beta\alpha}.[\forall x_\alpha, y_\beta.Rxy \Rightarrow Qxy]$ |
| $\text{IsRelOn}(\_,\_,\_)$ | $:= \lambda R_{o\beta\alpha}, A_{o\alpha}, B_{o\beta}.[\forall x_\alpha, y_\beta.Rxy \Rightarrow x \in A \wedge y \in B]$ |
| $\text{RestrictRDom}(\_,\_)$ | $:= \lambda R_{o\beta\alpha}, A_{o\alpha}.[\lambda x_\alpha, y_\beta.x \in A \wedge Rxy]$ |

**Table 2.** Definitions of Operations on Sets and Relations: Examples

These concepts are defined in terms of $\lambda$-expressions and they may contain other, already specified concepts. For presentation purposes, we use customary mathematical symbols $\cup, \cap$, etc., for some concepts like *union, intersection*, etc., and we also use infix notation. For instance, the definition of union on sets in Table 2 can be easily read in its more common mathematical representation $A \cup B := \{x | x \in A \vee x \in B\}$. Before proving a problem, Leo always expands —



| | | |
|---|---|---|
| Assumptions: | $\forall B, C, x_\bullet[x \in (B \cup C) \Leftrightarrow x \in B \vee x \in C]$ | (1) |
| | $\forall B, C, x_\bullet[x \in (B \cap C) \Leftrightarrow x \in B \wedge x \in C]$ | (2) |
| | $\forall B, C_\bullet[B = C \Leftrightarrow B \subseteq C \wedge C \subseteq B]$ | (3) |
| | $\forall B, C_\bullet[B \cup C = C \cup B]$ | (4) |
| | $\forall B, C_\bullet[B \cap C = C \cap B]$ | (5) |
| | $\forall B, C_\bullet[B \subseteq C \Leftrightarrow \forall x_\bullet x \in B \Rightarrow x \in C]$ | (6) |
| | $\forall B, C_\bullet[B = C \Leftrightarrow \forall x_\bullet x \in B \Leftrightarrow x \in C]$ | (7) |
| Proof Goal: | $\forall B, C, D_\bullet[B \cup (C \cap D) = (B \cup C) \cap (B \cup D)]$ | (8) |

**Table 3.** Problem SET171+3: The First-Order TPTP Encoding.

recursively, if necessary — all occurring concepts. This straightforward expansion to first principles is realised by an automated preprocess in our current approach.

### 2.3 Sets, Relations, and Functions: First-Order Logic Encoding

Let us consider example SET171+3 in its first-order formulation from the TPTP (see Table 3). We can observe that the assumptions provide only a partial axiomatisation of naive set theory. On the other hand, the specification introduces lemmata that are useful for solving the problem. In particular, assumption (7) is trivially derivable from (3) with (6). Obviously, clausal normalisation of this first-order problem description yields a much larger and more difficult set of clauses. Furthermore, definitions of concepts are not directly expanded as in LEO. It is therefore not surprising that most first-order ATPs still fail to prove this problem. In fact, very few TPTP provers were successful in proving SET171+3. Amongst them are MUSCADET 2.4. [21], VAMPIRE 7.0, and SATURATE. The natural deduction system MUSCADET uses special inference rules for sets and needs 0.2 seconds to prove this problem. VAMPIRE needs 108 seconds. The SATURATE system [15] (which extends VAMPIRE with Boolean extensionality rules that are a one-to-one correspondence to LEO's rules for Extensional Higher-Order Paramodulation [3]) can solve the problem in 2.9 seconds while generating 159 clauses. The significance of such comparisons is clearly limited since different systems are optimised to a different degree. One noted difference between the experiments with first-order provers listed above, and the experiments with LEO and LEO-BLIKSEM is that first-order systems often use a case tailored problem representation (e.g., by avoiding some base axioms of the addressed theory), while LEO and LEO-BLIKSEM have a harder task of dealing with a general (not specifically tailored) representation. Thus, the comparison of the performance of LEO and LEO-BLIKSEM with first-order systems as done in [9] is unfair: the higher-order systems attack harder, non-tailored problems. Nevertheless, as we demonstrated by the performance results in [9] the higher-order systems still perform better.





For the experiments with LEO and the cooperation of LEO with the first-order theorem prover BLIKSEM, $\lambda$-abstraction as well as the extensionality treatment inherent in LEO's calculus [4] is used. This enables a theoretically[1] Henkin-complete proof system for set theory. In the above example SET171+3, LEO generally uses the application of functional extensionality to push extensional unification constraints down to base type level, and then eventually applies Boolean extensionality to generate clauses from them. These are typically much simpler and often even *propositional-like* or *first-order-like* (FO-like, for short), that is, they do not contain any 'real' higher-order subterms (such as a $\lambda$-abstraction or embedded equations), and are therefore suitable for treatment by a first-order ATP or even a propositional logic decision procedure.

### 2.4 Solving the Test Problem SET171+3 in LEO

Table 4 illustrates how LEO tackles and solves the test problem SET171+3. First the resolution process is initialised, that is, the definitions occurring in the input problem are expanded, that is, completely reduced to first principles. Then the problem is turned into a negated unit clause. The resulting (not displayed intermediate) clause is not in normal form and therefore LEO first normalizes it with explicit clause normalisation rules (cnf) to reach some proper initial clauses. In

---

[1] For pragmatic reasons, such as efficiency, most of LEO's tactics are incomplete. LEO's philosophy is to rely on a theoretically complete calculus, but to practically provide a set of complimentary strategies so that these cover a broad range of theorems.



our concrete case, this leads to the unit clause (2). Note that negated primitive equations are generally automatically converted by LEO into unification constraints. This is why (2) is automatically converted into (3), which is a syntactically not solvable, but is a semantic unification problem. Observe, that we write $[.]^T$ and $[.]^F$ for positive and negative literals, respectively. LEO then applies its goal directed functional and Boolean extensionality rules which replace the unification constraint (3) by the clauses (4) and (5). Unit clause (5) is again not normal; normalisation, factorisation and subsumption yields the clauses (6)-(9). This set is essentially of propositional logic character and trivially refutable. LEO needs 0.56 seconds for solving the problem and generates a total of 36 clauses.

### 2.5 Solving the Test Problem SET171+3 in LEO-BLIKSEM

As illustrated in Table 4, LEO transforms test problem SET171+3 straightforwardly into a propositional like subproblem. Here the generated clause set (7)–(10) can still be efficiently refuted by LEO. Generally, however, the generated subsets of propositional or first-order like subproblems may quickly become so big that LEO's refutation procedure, which is not optimised for these problem classes, gets stuck. And in LEO's search space generally some further real higher-order clauses have to be taken into account. This observation motivates our cooperative LEO-BLIKSEM proof search approach: while LEO performs its proof search as before, it periodically also passes the detected first-order like clauses (which, of course, include the propositional like clauses) to the first-order specialist reasoner BLIKSEM. We note:

- The generated first-order like clauses in LEO are copied into a special bag which never decreases and usually always increases. That is, the bag of first-order like clauses dynamically changes and eventually becomes refutable (such as clauses (7)–(10) in our example).
- LEO's proof search procedure remains unchanged and LEO still tries to refute such subproblems itself (as before) in a bigger context.
- In addition, specialist reasoners may now support LEO by showing that the bag of first-order like subproblems is refutable.
- Each time the bag of first-order like subproblems is increased by LEO, a new instance of a specialist reasoner is launched (with a resource-bound). This instance runs in parallel to LEO's proof search and may eventually signal success to LEO. If LEO receives such a success signal, it stops its own proof search and reports that a cooperative proof has been found. Alternatively (as before) LEO stops proof search when it finds the proof itself.
- Our cooperative approach can easily be fine-grained by separating the bag of first-order like clauses into even more specialised subclasses, such as propositional logic, guarded fragment, etc. Different specialist reasoners can then be employed to attack these clause sets.
- For the higher-order problems investigated in [9] we further observe:
  - Some problems are immediately mapped by recursive definition expansion (without extensionality reasoning) and normalisation into first-order like problems; an example is SET624+3.



- Some problems are immediately mapped by recursive definition expansion (without extensionality reasoning) and normalisation into the empty clause such that proof search does not even start; an example is SET646+3.
- Some problems require several rounds of extensionality processing within LEO's set-of-support based proof search procedure before the bag of first-order like clauses turns into a refutable set of clauses; an example is SET611+3.

The result of the case study performed in [9] is: The LEO-BLIKSEM cooperation impressively outperforms both state-of-the art first-order specialists (including Vampire 7.0) and the non-cooperative LEO system.

In the next section we describe in more detail how the cooperative proof search approach between LEO and the first-order prover BLIKSEM has been modelled in $\Omega$ANTS.

## 3    $\Omega$ANTS

$\Omega$ANTS was originally conceived to support interactive theorem proving but was later extended to a fully automated proving system [23, 8]. Its basic idea is to compose a *central proof object* by generating, in each proof situation, a ranked list of potentially applicable inference steps. In this process, all inference rules, such as calculus rules or tactics, are uniformly viewed with respect to three sets: premises, conclusions, and additional parameters. The elements of these three sets are called *arguments* of the inference rule and they usually depend on each other. An inference rule is applicable if at least some of its arguments can be instantiated with respect to the given proof context. The task of the $\Omega$ANTS architecture is now to determine the applicability of inference rules by computing instantiations for their arguments.

The architecture consists of two layers. On the lower layer, possible instantiations of the arguments of individual inference rules are computed. In particular, each inference rule is associated with its own blackboard and concurrent processes, one for each argument of the inference rule. The role of every process is to compute possible instantiations for its designated argument of the inference rule, and to record these on the blackboard. The computations are carried out with respect to the given proof context and by exploiting information already present on the blackboard, that is, argument instantiations computed by other processes. On the upper layer, the information from the lower layer is used for computing and heuristically ranking the inference rules that are applicable in the current proof state. The most promising rule is then applied to the central proof object and the data on the blackboards is cleared for the next round of computations.

$\Omega$ANTS employs resource reasoning to guide search.[2] This enables the controlled integration (e.g., by specifying time-outs) of full-fledged external reason-

---

[2] $\Omega$ANTS provides facilities to define and modify the processes at run-time. But notice that we do not use these advanced features in the case study presented in this paper.



ing systems such as automated theorem provers, computer algebra systems, or model generators into the architecture.

## 3.1 Cooperation via multiple inference rules

The use of the external systems is modelled by inference rules, usually one for each system. Their corresponding computations are encapsulated in one of the independent processes in the architecture. For example, an inference rule modelling the standard application LEO has its conclusion argument set to be an open higher-order (HO) goal.

$$\frac{}{\textbf{HO-goal}} \text{ LEO } \textbf{(LEO-parameters)}$$

A process can then place an open goal on the blackboard, where it is picked up by a process that applies the LEO prover to it. Any computed proof from the external system is again written to the blackboard from where it is subsequently inserted into the proof object when the inference rule is applied. While this setup enables proof construction by a collaborative effort of diverse reasoning systems, the cooperation can only be achieved via the central proof object. This means that all partial results have to be translated back and forth between the syntaxes of the integrated systems and the language of the proof object. For modelling the cooperation of LEO with a first-order reasoner we have first experimented with the following multiple inference rule modelling (see also [5]):

$$\frac{\textbf{Neg-Conj-of-FO-clauses}}{\textbf{HO-goal}} \text{ LEO-with-partial-result} \textbf{(LEO-parameters)}$$

$$\frac{}{\textbf{FO-goal}} \text{ BLIKSEM } \textbf{(Bliksem-parameters)}$$

The first rule models a process that picks up higher-order proof problem from the blackboard, passes it to LEO which starts its proof search, and then returns the negated conjunction of generated first-order clauses back (e.g. the negated conjunction of the clauses (7)–(10) in our previous example). For each modified bag of first-order like clauses in LEO this rule may suggest a novel reduction of the original higher-order goal to a first-order criterion.

Since there are many types of integrated systems, the language of the proof object maintained in $\Omega$ANTS — a higher-order language even richer than LEO's, together with a natural deduction calculus — is expressive but also cumbersome. This leads not only to a large communication overhead, but also means that complex proof objects have to be created, even if the reasoning of all systems involved is clause-based. Large clause sets need to be transformed into large single formulae to represent them in the proof object; the support for this in $\Omega$ANTS to date is inefficient. Consequently, the cooperation between external systems is typically rather inefficient [5].



### 3.2 Cooperation via a single inference rule

In order to overcome the problem of the communication bottleneck described above, we devised a new method for the cooperation between a higher-order and a first-order theorem prover within $\Omega$ANTS. Rather than modelling each theorem prover as a separate inference rule (and hence needing to translate the communication via the language of the central proof object), we model the cooperation between a higher-order (concretely, LEO) and a first-order theorem prover (in our case study BLIKSEM) in $\Omega$ANTS as a single inference rule.

$$\frac{}{\textbf{HO-goal}} \text{ LEO-BLIKSEM } \left( \begin{array}{l} \textbf{Leo-partial-proof, FO-clauses, FO-proof, Leo-} \\ \textbf{parameters, Bliksem-parameters} \end{array} \right)$$

The communication between the two theorem provers is carried out directly by the parameters of the inference rule and not via the central proof object. This avoids translating clause sets into single formulae and back.

Concretely, the single inference rule modelling the cooperation between LEO and a first-order theorem prover needs the following arguments to be applicable: (1) an open higher-order proof goal, (2) a partial LEO proof, (3) a set of FO-like clauses in the partial proof, (4) a first-order refutation proof for the set of FO-like clauses, and (5) and (6) the usual flag-parameters for the theorem provers LEO and BLIKSEM. Each of these arguments is computed, that is, its instantiation is found, by an independent process. The first process finds open goals in the central proof object and posts them on the blackboard associated with the new rule. The second process starts an instance of the LEO theorem prover for each new open goal on the blackboard. Each LEO instance maintains its own set of FO-like clauses. The third process monitors these clauses, and as soon as it detects a change in this set, that is, if new FO-like clauses are added by LEO, it writes the entire set of clauses to the blackboard. Once FO-like clauses are posted, the fourth process first translates each of the clauses directly into a corresponding one in the format of the first-order theorem prover, and then starts the first-order theorem prover on them. Note that writing FO-like clauses on the blackboard is by far not as time consuming as generating higher-order proof objects. As soon as either LEO or the first-order prover finds a refutation, the second process reports LEO's proof or partial proof to the blackboard, that is, it instantiates argument (2). Once all four arguments of our inference rule are instantiated, the rule becomes applicable and its application closes the open proof goal in the central proof object. That is, the open goal is proved by the cooperation between LEO and a first-order theorem prover. When computing applicability of the inference rule, the second and the fourth process concurrently spawn processes running LEO or a first-order prover on a different set of FO-like clauses. Thus, when actually applying the inference rule, all these instances of provers working on the same open subgoal are stopped.

While in the previous approach with multiple inference rules the cooperation between LEO and BLIKSEM was modelled at the upper layer of the $\Omega$ANTS architecture, our new approach models their cooperation by exploiting the lower



layer of the $\Omega$ANTS blackboard architecture. This is not an ad hoc solution, but rather, it demonstrates $\Omega$ANTS's flexibility in modelling the integration of cooperative reasoning systems.

Our approach to the cooperation between a higher-order and a first-order theorem prover has many advantages. The main one is that the communication is restricted to the transmission of clauses, and thus it avoids any intermediate translation into the language of the central proof object. This significantly reduces the communication overhead and makes effective proving of more involved theorems feasible.

## 4 Constructing a Combined Proof Object

A disadvantage of our approach is that we cannot easily translate and integrate the two proof objects produced by LEO and BLIKSEM into the central proof object maintained by $\Omega$ANTS. This has been possible in our previous approach with multiple inference rules. Thus, we developed a simple and pragmatic solution to the problem:

- The main idea is to replay the proof on the upper level of the $\Omega$ANTS architecture (using the multiple inference rule modelling) once a proof attempt was successful (with a single inference rule modelling) on the lower level.
- We can essentially reconstruct all the information from the blackboard that we need in order to replay the proof. For this remember that the rule LEO-BLIKSEM is only applicable if all parameters of the rule are instantiated, that is, the respective parameter instantiation information is available on the blackboard for each successful cooperative proof attempt. Respective instantiation information generated from a successful cooperative proof attempt for our running example SET171+3, for instance, is:

$$\textbf{HO-Goal} := \forall B, C, D. C \cup (B \cap D) = (C \cup B) \cap (C \cup D)$$

$$\textbf{Leo-partial-proof} := \dots a \text{ } HO \text{ } resolution \text{ } proof \text{ } object \text{ } \Delta \text{ } \dots$$

$$\textbf{FO-clauses} := \begin{array}{ll} (7) & [Bx]^F \\ (8) & [Bx]^T \vee [Cx]^T \\ (9) & [Bx]^T \vee [Dx]^T \\ (10) & [Cx]^F \vee [Dx]^F \end{array}$$

$$\textbf{FO-proof} := \dots a \text{ } FO \text{ } resolution \text{ } proof \text{ } object \text{ } \Gamma \text{ } \dots$$

$$\textbf{Leo-parameters} := \dots the \text{ } flags \text{ } chosen \text{ } for \text{ } the \text{ } \text{LEO} \text{ } call \dots$$

$$\textbf{Bliksem-parameters} := \dots the \text{ } flags \text{ } chosen \text{ } for \text{ } the \text{ } \text{BLIKSEM} \text{ } call \dots$$

- For finding joint proofs efficiently in our experiment we called BLIKSEM in the fastest mode. In this case the generated FO-proof object is typically very sparse, i.e. contains only very little information for proof reconstruction and transformation.



- When the above suggestion of a successful joint proof attempt is selected for application in $\Omega$ANTS, the initially open (sub-)goal $\forall B, C, D.C \cup (B \cap D) = (C \cup B) \cap (C \cup D)$ is closed and the new justification of this proof node becomes 'LEO-BLIKSEM' augmented with the above parameter instantiation information:

$$\overline{\forall B, C, D.C \cup (B \cap D) = (C \cup B) \cap (C \cup D)} \text{ LEO-BLIKSEM } \textit{(above param. inst.)}$$

- Expansion of this node then replaces the (sub-)proof object by the following (sub-)proof object employing the multiple inference rule modelling of the cooperative proof attempt:

$$\frac{\overline{neg\text{-}FO\text{-}clauses} \text{ BLIKSEM } \textit{(modified Bliksem-param. instantiation)}}{\forall B, C, D. \ldots = \ldots} \text{ LEO-\textit{with-partial-result} } \textit{(Leo-param. instantiation)}$$

where '*neg-FO-clauses*' is computed from the instantiation of the parameter **FO-clauses** as

$$\neg(\neg(Bx) \land (Bx \lor Cx) \land (Bx \lor Dx) \land (\neg(Cx) \lor \neg(Dx))$$

- The idea is to support verification of this (sub-)proof by subsequent proof node expansion, i.e., to investigate the contributions of both reasoning systems separately. For the expansion of BLIKSEM, a translation of the previously generated proof into a proper proof-object is not an option if we called BLIKSEM in the fastest mode since the delivered first-order proof object may be too sparse. Therefore, the expansion of this proof node simply calls BLIKSEM again but now within a different mode (determined by the slightly changed *modified Bliksem-param. instantiation*) which ensures the generation of detailed first-order proof objects.
- For the translation of this regenerated, detailed first-order proof object into an $\Omega$ANTS proof object we employ the TRAMP system [18]. This enables us to verify the (sub-)proof of BLIKSEM after its translation into an $\Omega$ANTS proof object.
- Generally, we could also replace the second call to BLIKSEM by a call to any other first-order proof system that is supported by TRAMP's generic proof transformation mechanism (and which is as strong as BLIKSEM).

## 5 Conclusion

In this paper we have discussed the difference between two forms of modelling cooperating proof systems within $\Omega$ANTS: the multiple inference rule approach and the single inference rule approach. In previous experiments the latter has been shown as highly efficient and it has outperformed state-of-the-art first-order specialist reasoners on 45 examples on sets, relations and functions; cf. [9]. The drawback so far, however, was that no joint proof object could be generated. In



this paper we have reported how we have solved this problem by simply mapping the single inference rule modelling back to the multiple inference rule modelling.

Related to our approach is the TECHS system [13], which realises a cooperation between a set of heterogeneous first-order theorem provers. Similarly to our approach, partial results in TECHS are exchanged between the different theorem provers in form of clauses. The main difference to the work of Denzinger *et al.* (and other related architectures like [14]) is that our system bridges between higher-order and first-order automated theorem proving. Also, unlike in TECHS, we provide a declarative specification framework for modelling external systems as cooperating, concurrent processes that can be (re-)configured at run-time. Related is also the work of Hurd [16] which realises a generic interface between HOL and first-order theorem provers. It is similar to the solution previously achieved by TRAMP [18] in OMEGA, which serves as a basis for the sound integration of ATPs into $\Omega$ANTS. Both approaches pass essentially first-order clauses to first-order theorem provers and then translate their results back into HOL resp. OMEGA. Some further related work on the cooperation of Isabelle with VAMPIRE is presented in [19]. The main difference of our work to the related systems is that while our system calls first-order provers from within higher-order proof search, this is not the case for [16, 18, 19].

Future work is to investigate how far our approach scales up to more complex problems and more advanced mathematical theories. In less trivial settings as discussed in this paper, we will face the problem of selecting and adding relevant lemmata to avoid immediate reduction to first principles and to appropriately instantiate set variables. Relevant related work for this setting is Bishop's approach to *selectively expand definitions* as presented in [10] and Brown's PhD thesis on *set comprehension in Church's type theory* [11].

# References


1. P. Andrews. *An Introduction to mathematical logic and Type Theory: To Truth through Proof.* Number 27 in Applied Logic Series. Kluwer, 2002.
2. C. Benzmüller. *Equality and Extensionality in Higher-Order Theorem Proving.* PhD thesis, Universität des Saarlandes, Germany, 1999.
3. C. Benzmüller. Extensional higher-order paramodulation and RUE-resolution. *Proc. of CADE-16, LNAI* 1632, p. 399–413. Springer, 1999.
4. C. Benzmüller. Comparing approaches to resolution based higher-order theorem proving. *Synthese*, 133(1-2):203–235, 2002.
5. C. Benzmüller, M. Jamnik, M. Kerber, and V. Sorge. Experiments with an Agent-Oriented Reasoning System. *Proc. of KI 2001, LNAI* 2174, p.409–424. Springer, 2001.
6. C. Benzmüller and M. Kohlhase. LEO – a higher-order theorem prover. *Proc. of CADE-15, LNAI* 1421. Springer, 1998.
7. C. Benzmüller and V. Sorge. A Blackboard Architecture for Guiding Interactive Proofs. *Proc. of AIMSA'98, LNAI* 1480, p. 102–114. Springer, 1998.
8. C. Benzmüller and V. Sorge. $\Omega$ANTS – An open approach at combining Interactive and Automated Theorem Proving. *Proc. of Calculemus-2000.* AK Peters, 2001.
9. C. Benzmüller, V. Sorge, M. Jamnik, and M. Kerber. Can a Higher-Order and a First-Order Theorem Prover Cooperate? Proc. LPAR'04, LNAI 3452, Montevideo, Uruguay. Springer, 2005.





10. M. Bishop and P. Andrews. Selectively instantiating definitions. *Proc. of CADE-15, LNAI* 1421. Springer, 1998.

11. C. E. Brown. *Set Comprehension in Church's Type Theory.* PhD thesis, Dept. of Mathematical Sciences, Carnegie Mellon University, USA, 2004.

12. H. de Nivelle. *The Bliksem Theorem Prover, Version 1.12.* Max-Planck-Institut, Saarbrücken, Germany, 1999. http://www.mpi-sb.mpg.de/ bliksem/manual.ps.

13. J. Denzinger and D. Fuchs. Cooperation of Heterogeneous Provers. *Proc. IJCAI-16,* p. 10–15. Morgan Kaufmann, 1999.

14. M. Fisher and A. Ireland. Multi-agent proof-planning. CADE-15 Workshop "Using AI methods in Deduction", 1998.

15. H. Ganzinger and J. Stuber. Superposition with equivalence reasoning and delayed clause normal form transformation. *Proc. of CADE-19, LNAI* 2741. Springer, 2003.

16. J. Hurd. An LCF-style interface between HOL and first-order logic. *Automated Deduction — CADE-18, LNAI* 2392, p. 134–138. Springer, 2002.

17. M. Kerber. *On the Representation of Mathematical Concepts and their Translation into First Order Logic.* PhD thesis, Universität Kaiserslautern, Germany, 1992.

18. A. Meier. TRAMP: Transformation of Machine-Found Proofs into Natural Deduction Proofs at the Assertion Level. *Proc. of CADE-17, LNAI* 1831. Springer, 2000.

19. J. Meng and L. C. Paulson. Experiments on supporting interactive proof using resolution. *Proc. of IJCAR 2004, LNCS* 3097, p. 372–384. Springer, 2004.

20. R. Nieuwenhuis, Th. Hillenbrand, A. Riazanov, and A. Voronkov. On the evaluation of indexing techniques for theorem proving. *Proc. of IJCAR-01, LNAI* 2083, p. 257–271. Springer, 2001.

21. D. Pastre. Muscadet2.3 : A knowledge-based theorem prover based on natural deduction. *Proc. of IJCAR-01, LNAI* 2083, p. 685–689. Springer, 2001.

22. A. Riazanov and A. Voronkov. Vampire 1.1 (system description). *Proc. of IJCAR-01, LNAI* 2083, p. 376–380. Springer, 2001.

23. V. Sorge. *OANTS: A Blackboard Architecture for the Integration of Reasoning Techniques into Proof Planning.* PhD thesis, Universität des Saarlandes, Germany, 2001.

24. G. Stenz and A. Wolf. E-SETHEO: An Automated[3] Theorem Prover – System Abstract. *Proc. of the TABLEAUX'2000, LNAI* 1847, p. 436–440. Springer, 2000.

25. G. Sutcliffe and C. Suttner. The TPTP Problem Library: CNF Release v1.2.1. *Journal of Automated Reasoning,* 21(2):177–203, 1998.




# Benchmarks for Higher-Order Automated Reasoning


Chad E. Brown

Universität des Saarlandes, Saarbrücken, Germany, cebrown@ags.uni-sb.de


For a higher-order system to be successful it should support users performing tasks both large and small. Large tasks include interactive construction of large theories, including storing definitions, theorems and proofs. Small tasks include using automation to fill in small gaps in proofs. Consider the following theorem:

**(C)** If $f$ is an $n$-tuple of complex numbers and $f_i = 0$, then the product $f_1 \cdots f_n$ of the $n$-tuple is 0.

In order to even state this theorem, one must first have already defined the complex numbers, $n$-tuples of complex numbers and multiplication of such $n$-tuples. If one defines the complex numbers using pairs of reals, defines the reals using Dedekind cuts, and so on, then it is unrealistic to expect a system to automatically prove this theorem. On the other hand, suppose we include the following as a hypothesis:

**(A)** Any $n$-tuple of complex numbers has product 0 iff there exists some $j$ between 1 and $n$ such that $f_j = 0$.

Proving the first from the second (i.e., $[\mathbf{A} \supset \mathbf{C}]$) is a minor exercise in logic and is precisely the sort of gap automation should be able to fill. Such problems are not trivial, however, since an automated system might consider any number of irrelevant possibilities during proof search. This is especially true once one begins expanding definitions.

The example above comes from Jutting's translation of Landau's *Grundlagen der Analysis* [6] into Automath [5]. This formalization was recovered and restored by Wiedijk [7]. We are now in the process of porting the definitions and theorems from the Automath signature into Church's Type Theory, a form of higher-order logic based on simple type theory [2, 4]. However, we are not translating the Automath proof objects to Church's Type Theory. Consequently, one obtains thousands of unproven theorems such as **C**.

For each such theorem **C**, one can attempt to prove **C** in isolation (possibly making use of axioms such as description, choice, extensionality or infinity). For most theorems this is unrealistic since the gap between the axioms and the theorem is simply too wide. On the other hand, one can take all axioms and previously proven theorems $\mathbf{A}_1, \dots \mathbf{A}_m$ and try to prove **C** follows from the conjunction of $\mathbf{A}_1, \dots \mathbf{A}_m$. This is generally unrealistic for two reasons. Firstly, the formula becomes too large for automated search once $m$ becomes large. Secondly, if some $\mathbf{A}_i$ contains type variables, then one must find a way to instantiate these



type variables during the search for a proof. A third alternative is the most realistic. We provide precisely the relevant axioms and previously proven theorems, with the correct type variable instantiations, and try to prove $\mathbf{C}$ follows. In general, of course, we cannot know which of the $\mathbf{A}_i$'s are relevant. However, for the *Grundlagen* theorems, we can extract this information from the Automath proof terms. Using this information, we obtain thousands of theorems of the form

$$[\mathbf{A}'_{i_1} \wedge \cdots \wedge \mathbf{A}'_{i_k}] \supset \mathbf{C}$$

where $\mathbf{A}'_{i_1}$ is $\mathbf{A}_{i_1}$ with types instantiated appropriately.

Most such theorems correspond to a step given by a single line in the Automath code. For this reason, one can expect most of the theorems to have reasonably short proofs. On the other hand, Automath has a stronger type system than simple type theory, so one step in Automath may correspond to many steps in the simply typed version. (Intuitively, what was type checking in Automath becomes deduction in Church's type theory.)

This corpus of theorems can be used to empirically test the automated facilities of a higher-order reasoning system. The hope is that such a corpus can provide reasonable, practical benchmarks for judging the effectiveness and efficiency of systems and procedures. We report on the initial results of applying the theorem prover TPS [1,3].

# References


1. Peter B. Andrews, Matthew Bishop, Sunil Issar, Dan Nesmith, Frank Pfenning, and Hongwei Xi. TPS: A theorem proving system for classical type theory. *Journal of Automated Reasoning*, 16:321–353, 1996.
2. Peter B. Andrews. *An Introduction to Mathematical Logic and Type Theory: To Truth Through Proof*. Kluwer Academic Publishers, second edition, 2002.
3. Chad E. Brown. *Set Comprehension in Church's Type Theory*. PhD thesis, Department of Mathematical Sciences, Carnegie Mellon University, 2004.
4. Alonzo Church. A Formulation of the Simple Theory of Types. *Journal of Symbolic Logic*, 5:56–68, 1940.
5. L. S. Jutting. *Checking Landau's "Grundlagen" in the AUTOMATH system*. PhD thesis, Eindhoven Univ., Math. Centre, Amsterdam, 1979.
6. E. Landau. *Grundlagen der Analysis*. Leizig, 1930.
7. Freek Wiedijk. A new implementation of Automath. *J. Autom. Reasoning*, 29(3-4):365–387, 2002.




# Co-Synthesis of New Complex Selection Algorithms and their Human Comprehensible XML Documentation


Jutta Eusterbrock

JEusterbrock@seamless-solutions.de



**Abstract.** In this paper, an approach for program and algorithm synthesis within a higher-order framework is presented that allows us to generate new algorithm structures together with readable documentation which enable users to conveniently inspect the results of this synthesis process. The synthesis approach is based on feature graphs as a higher-order data type for specifying synthesis types like proof terms, theorems and document types. The document synthesis method uses constructive reasoning about a user-defined knowledge base to synthesise documents from metaobjects which are composed of diagrams, text, and references to articles, theorems and algorithms. The data format used for reasoning is a variant of XML syntax and thus enables the organisation of comprehensive knowledge in XML repositories and the conversion of document terms into LaTeX, PDF or XHTML documents which can be displayed by browsers.

This framework has been applied for the automated synthesis of several new selection algorithms and their documentation. As one new result, an algorithm that proves selecting the 4th element of 24 elements needs at most 34 comparisons was synthesised using 50 seconds CPU time on an AMD Athlon 3200. The correctness of the synthesised algorithm was manually checked. It improves the known upper bounds for the specific selection problems. The running time for synthesis is several orders of magnitude more efficient than comparative approaches.


## 1 Introduction

The idea of automatically discovering solutions to mathematical conjectures or proving the non-existence of solutions which are beyond human comprehension is intriguing. In recent years, computers have been used for a number of famous algorithmic problems to facilitate analysis based upon a detailed problem specific formalisation. For instance, C. Lam proved that a projective plane of order 10 does not exist [Lam91]. As the "proof" took several years of computer search (the equivalent of 2000 hours on a Cray-1), it is considered to be the most time-intensive computer assisted single proof. Recently, G. Gonthier, a mathematician who works at Microsoft Research in Cambridge, England, used the CoQ system (cf. [Dev05]) to verify the computer supported proof of the famous Four Color



Theorem, which was originally proven in 1976 by Appel and Haken who used computer programs to check a very large number of cases.

However, for a mathematician it is unsatisfying to know that there exists a solution or no solution for a problem, because thousands, hundreds of thousands or millions of states have been explored by a theorem prover whose rules have been verified, and at the same time, not be able to comprehend the tedious machine-generated proofs and not be able to draw conclusions from the automated proof.

In previous work [Eus92b] on automated algorithm synthesis, we designed and implemented a metalevel methodology and system to assist algorithm synthesis and the proof of lower complexity bounds. The system evaluated the search process and derived new knowledge which was abstracted and added to a dynamically growing knowledge base. The system was applied to assist the proof of a number-theoretic conjecture for the selection problem. It was possible to prove the values for very small $n$ within a few minutes on a Sparc Workstation. Surprisingly, the system synthesised an algorithm which constructed a counter example to a set of rules which should be verified by the system [Eus92a]. As a side effect, an isolated counter-example for a published lower bound for the selection problem was constructed. The computation took several days and the machine generated proofs were cumbersome because the nested proof graphs are hard to comprehend. However, treating proofs as graphs enables various forms for generating explanations. In [EN96], a visualisation component was implemented in order to assist the exploration of huge graphs. Techniques like zooming, organising of graphs on various hierarchy levels, folding and unfolding techniques enabled the interactive exploration of large graphs. However, it is assumed that presenting algorithms by documents which are similar to scientific or textbook descriptions, making use of different formats and establishing links to published results is cognitively more appropriate for proof presentation than uniform means such as visualisation techniques. Meanwile, XML-based data formats have emerged as a standard for data encoding and exchange. XML documents can be comparatively easily converted into web pages, LaTeX or PDF document files which facilitate the display of natural language, graphics, mathematical symbols and references in an appropriate form.

In this paper, it is illustrated how XML based documentation for automatically synthesised algorithms and programs can be generated from specifications and proof terms and further processed by XML tools is analysed. Synthesised theorems and programs should be presented at the right level of granularity such that experts can check the automatically synthesised programs in the same way as they verify a proof in a publication. The key to a solution is abstraction. It is achieved by a higher-order formalisation of abstract synthesis objects and document components and a higher-order approach towards automated synthesis which raises the level of abstraction (cf. [Kre93]). A method is devised that transforms program specifications and synthesised algorithms into documents which include graphics, text, and references. The method is based upon general correspondences between synthesis types and document types. User-defined



context-specific rules as how to decompose the proof graph, when to generate lemmata and what kind of additional visual information to present at various stages can be provided. The method has been applied to automatically synthesise documents for new complex selection algorithms which were automatically synthesised.

The paper is organised as follows. In section 2, the selection problem is introduced. Section 3 gives an overview of the synthesis framework. In section 4, the synthesis types and the building blocks of the metatheory for the encoding of mathematical knowledge are defined. The document synthesis method is exlained in section 5. Section 6 summarises some experimental results. Section 7 concludes this paper. Appendix A contains the automatically synthesised document for the automatically constructed new selection algorithm.

## 2 The Selection Problem

The selection problem is the problem of finding the i-th largest element, given a set of n distinct unordered numbers, $1 < i < n$. The special case $i = \lceil n/2 \rceil$ is the median problem. The worst-case, minimum number of comparisons is denoted by $V_i(n)$. The problem goes back to Rev. C. L. Dodgson's (aka Lewis Carroll) essay on how prizes were awarded unfairly in tennis tournaments (see Knuth [8:5.33]). In the classic book *The Art of Computer Programming, Volume 2, Sorting and Searching* [Knu73b], D. Knuth introduces the problem and states the combinatorial bounds for $n \leq 10$. In [Eus85], the present author constructed the formula $H_i(n)$

$$H_i(n) = n - i + \sum_{l=1}^{i-1}(\lceil lg(\frac{n-i+2}{i-l+3})\rceil + 2).\tag{1}$$

It was shown in [Eus85] that the numbers $H_i(n)$ unify the published results for the worst-case behaviour of selection-algorithms as follows. For admissible combinations of $i, n$ the numbers $H_i(n)$:

- match the exact numbers $V_i(n)$ for $i = 1, 2, 3$ [Kis64,Knu73b,Aig82];
- are equal to or less than the lower bounds [Kis64,Yao74,FG79,Aig82,MP82,BJ85];
- are equal or greater than the upper bounds
  [Kis64,HS69,BFP⁺73,FR75,SPP76,Yap76,Aig82,RH84]

known at that time. Furthermore, using the numbers $H_i(n)$, novel combinatorial algorithms for small values of $i, n$ were constructed which prove the upper bounds for small values of $i, n$:

$$V_i(n) \leq H_i(n), \text{ iff } i \leq 4, n \leq 14 \text{ and } i \leq 5, n \leq 12.\tag{2}$$

The approximate behaviour for the medians is given by the formula below

$$H_{n/2}(n) \approx 2.5n - 3\lceil lg(n+4)\rceil + 5\tag{3}$$



The author stated the hypothesis $V_i(n) = H_i(n)$ for all $i, n$. For all known upper and lower bounds either $V_i(n) \geq L_i(n)$ or $V_i(n) \leq U_i(n)$ with one exception which will be described later on. The hypothetical formula for the median coincides with the conjecture of Yao and the conjecture of Paterson $V_{n/2} \approx 2.4094n$. In [Eus92b] we designed a system in order to assist the refinement of rules. Using this system, it was possible to prove the values for very small $n$ within a few minutes on a Sun Workstation. Surprisingly the system synthesised an algorithm which proves $V_3(22) < H_3(22)$ and thus constructed a counter example to a set of rules which should be verified by the system [Eus92a]. The computation took several days.

Computerised searches for the selection problem were subsequently performed by [GKP96] and [Oks05] which use alpha-beta search, a transposition table and some optimisations. Oksanen's system constructs decision graphs which in summary suggest $V_i(n) \leq H_i(n)$, iff $i \leq 6, n \leq 14$. However, it is also claimed that $V_5(12) \geq 19 = H_5(12) - 1$ while in [Eus85] an algorithm is presented that proves $V_5(12) \leq 18$ and thus contradicts the automatically proven statement. The constructed decision graphs presented in [Oks05] are too large to comprehend and in the case $i = 7, n = 14$ consist of more than 600 nodes.

The process for constructing algorithms or lower bounds for selection problems is similar to minimum-comparison sorting. M. Peczarski has devised a system and analysed the optimal lower bounds $S(n)$ for sorting $n$ elements, $n = 13, 14, 22$ based on an algorithm for counting the linear extensions of partial orders. The proof $S(22) > 70$ took 1740 hours on a computer with a 650 MHz processor.

# 3  The Higher-order Synthesis Framework

In this paper, a reformalisation and reimplementation of our knowledge-based synthesis framework, called SEAMLESS (cf. [Eus95]) is analysed. The synthesis system consists of verified generic methods which take a specification as input and generate metalevel proofs by metalevel reasoning about domain-specific knowledge which may be interpreted as programs. The synthesis system evaluates the search processes, generalises the case solutions which result from successful and failed proof attempts and stores them for further reuse in the knowledge base. A documentation synthesis component has been added. The document synthesis component transforms specifications, derived theorems and synthesised proofs into XML documents including graphical visualisations, references and text. The XML documents can be converted by freely available tools into human-readable documents in multiple formats such as XHTML, LaTeX or PDF. The resulting system structure is shown in Figure 1.

In order to achieve an integrated formal framework for the logic-based co-synthesis of proofs and their documentation, principles of a higher-order logic of program synthesis are applied (cf. [Kre93]). The core for the integration of the different types of knowledge and the correct design of the synthesis methods is a metatheory or, in other terms, an ontology. Types and higher-order predicates to



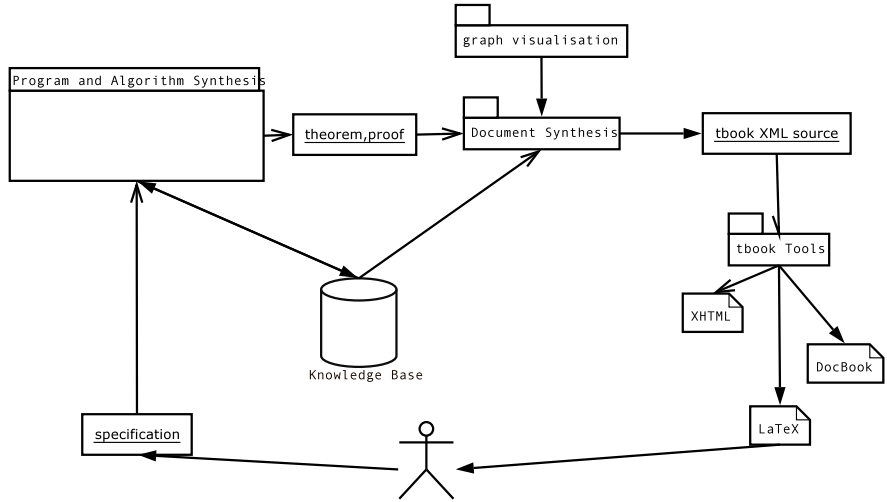

**Fig. 1.** Scenario for Knowledge-based Synthesis

represent the structure of algorithm design knowledge fragments and properties or relationships between them have been defined. Figure 2 lists the major types of the SEAMLESS framework.

Basic types      *Bool, Integer, Constants, Vars, String*
Object logic     *Atom, Clause, Algebraic Expression, Constraint*
Synthesis types  *Precondition, Postcondition, Program, Proof*

**Fig. 2.** Types for Program Synthesis

In this paper, the notation of [Kre93] is adopted and used in a semi-formal way. A definition *New Object Type ≡ Composition of Defined Object Types* defines a new object in terms of already existing object types. Meta theorems are written in the form $Goal \Leftarrow Subgoal_1 \wedge \ldots \wedge Subgoal_r$ or they are stated as facts.

The Floyd-Hoare logic [Hoa69] is used for specifying the semantics of imperative programs and to associate logical specifications with programs. In Floyd-Hoare's logic, triples of the form $\{Pre\}Prog\{Post\}$ state that if program $Prog$ starts in an input state satisfying $Pre$ then if and when $Prog$ halts, it does so in a state satisfying $Post$. Programs are sequences of statements. Hoare provided a set of logical rules in order to reason about the correctness of computer programs. It is well-known that the Floyd-Hoare rules and axioms can be embedded in higher order logic and become derived rules. In the SEAMLESS framework, a statement is a variable assignment, procedure or conditional. Loop statements together with their specifications can be added to the knowledge base, however,



their automatic synthesis is currently not supported as it is based on non-trivial mathematical induction. Consequently, the key conceptual building elements of the SEAMLESS theory include the abstractions precondition, postcondition, proofs which may be interpreted as programs, and, moreover, range constraints for a cost function as part of the postcondition. The metatheory is defined in terms of generic predicates and formal axioms for them which provide correctness axioms for the suitable domain theories. For example, the correctness axiom for the higher-order predicate $Know$ states that a proof for the validity of a Hoare triple is known.

$$Know(Pre, Post, Prog, True) \equiv \vdash \{Pre\} Prog \{Post\}$$

Domain-specific design knowledge can be provided by definitions for the open generic higher-order predicates, if the correctness axioms are satisfied. The SEAMLESS knowledge base of the system entails theorems which are relations among truth values, specifications and programs. They are stated as higher-order theorems, once types for the corresponding abstractions and the semantics of the higher-order predicates have been defined. In this application scenario, the knowledge bases contain published domain specific theorems about the complexity of selection problems, as summarised in section 2, and various related algorithms encoded as arguments of the higher-order predicate $Know$. The knowledge base of the synthesis system also comprises theorems whose proof is only given by a bibliographic reference to the corresponding document.

Generic synthesis methods have been derived from the generic predicates as metatheorems. The proofs-as-programs paradigm is adapted to the Hoare logic for the purpose of extending it to synthesising imperative programs. A synthesis method is a set of rules including metavariables for programs which are instantiated while proving the synthesis task. A very basic synthesis strategy is to retrieve solutions from the knowledge base which is specified below.

$$Synthesis(Pre, Prog, Post, Bool) \Leftarrow Know(Pre, Prog, Post, Bool)$$

Experience has shown that theorems as they are published in the literature are often not directly applicable for solving a problem specification. Equivalences, generalisations, and reductions are considered to establish semantic relations between specifications and the theorems in the literature to obtain proofs. These relations can be modelled by corresponding higher-order predicates.

## 4   Graphterms as a Higher-order Datastructure and XML

The extensible Markup Language (XML) has emerged as a quasi-standard for knowledge exchange, document processing and Web applications. XML is a metalanguage used to create generalised markup languages. XML annotations facilitate the retrieval of document fragments based upon their semantic annotations. The focus in this paper is on a more convenient and efficient data structure than arbitrary terms for the organisation of the knowledge fragments. The data structures are optimised in order



- to facilitate the automatic synthesis of human comprehensible XML-based documents;
- to elicit the interplay between design knowledge - which has been published in scientific documents and Web resources - and the corresponding encoding as theorem in the knowledge base of a reasoning system;
- to tackle the huge search complexities.

Linear (serial) term representations, named *graphterms*, were devised as a data structure for the encoding of labelled DAGs and feature graphs [Eus97]. Graphterms (cf. [Eus01,Eus97]) are used as the core data structure to encode formalised knowledge fragments in SEAMLESS.

**Definition 1** *Suppose that there are given an infinite set of variables, a set of features, and a set of constants. From features, constants attribute variables, and variables, terms are constructed, if features are seen equivalently as binary predicates that must be interpreted as functional relations. Let $f$ be a feature symbol, $X$ a variable. A feature graphterm is an expression $f(X, Graphtermlist)$, where Graphtermlist is either the empty list $[]$, or denotes a list of constants, variables and feature graphterms.*

To ensure that graphterms are directed acyclic graphs, further axioms constrain the valid terms. A graphterm algebra, that is term rewriting operations that implement graph operations and canonical forms for classes of isomorphic objects was constructed (cf. [Eus97]). Composed types can be defined using feature graphs and the objects of a type are instantiated feature graphs. The following assertion defines a type $Spec$

$$Spec([Id = No], [Pre([], [Pre]), Post([], [Post])]) \Leftarrow$$
$$Conjunction\_of\_Atom(Pre) \wedge Conjunction\_of\_Atom(Post).$$

Each attribute variable of a feature graph may be instantiated by a thuslist of attributes. In the examples above, specifications are assigned the attribute identifier. Attributes don't change the logical semantics of the terms, however, they may be used to design more efficient synthesis methods. It is possible to attach hash values to a graphterm by means of attributes. Objects may be substituted by references to them. An algorithm reference can be an automatically generated counter, e.g., *Alg12345*. Then it refers to an automatically generated object in the knowledge base or it refers to published algorithms, e.g., *Kislitsyn,Aigner* which were constructed outside the synthesis system. The use of references facilitates sharing of terms. It decreases the size of the knowledge base and makes relationships more obvious. The encoding of knowledge is demonstrated by the higher-order formula 4.

$$Know([], Pre([], [poset]), Prog([Id = id, Refid = Kislitsyn], []),$$
$$Post([], [Select(i, poset), Bound(r..r)], True) \Leftarrow$$
$$i = 2 \wedge Forest(poset) \wedge CostK(poset) = r. \qquad (4)$$



The advantage of using graphterms for encoding higher-order formulas is that feature graphterms directly correspond to XML Document Type Definitions (DTDs) and the instantiated ground terms are syntactic variants of XML (cf. [Eus01]). This facilitates the storing and maintenance of structured design knowledge fragments in XML repositories or XML databases, their retrieval based upon semantic annotations, and at the same time enables formal reasoning about them.

## 5 Higher-order XML-based Document Synthesis

To an increasing degree widely used XML Document Type Definitions like DocBook are being used for the structuring and mark-up of software documentation and scientific publications. Common document structuring elements are, for example, figure, graphics, theorem, proof or enumeration. These XML documents then can be transformed into different target formats like LaTeX, XHTML or PDF and provided with a professional layout using freely available tools. Document synthesis is a process that automatically generates a complete structured document. In this section, how to synthesise documents, given a program specification and its proof will be examined.

In order to enable constructive synthesis of documentation, a subset of the **t**book DTD is used to formalise the types of the theory for document synthesis. The **t**book DTD was chosen because it is an XML file format that is suitable for scientific texts, but it is also as simple and small as possible, uses similar names to LaTeX and it accepts MathML's presentation and contents markup. The **t**book tools for XML authoring (cf. [Bro05]) may be used to transform the XML document into XHTML, DocBook or LaTeX documents. The **t**book document types are used in the SEAMLESS knowledge base to model document structure. Type definitions are encoded by non-ground facts using feature graphs as knowledge representation format. The simplified document type is sketched in Figure 3. Automatic document synthesis is imple-

| Document | ≡ | Sequence of Header and Body |
|----------|---|------------------------------|
| Body | ≡ | Sequence of Theorems or Lemmata and their Proofs |
| Theorem, Lemma | ≡ | Sequence of Statements or Enumeration of Items |
| Proof | ≡ | Sequence of Statements or Enumeration of Items |
| Item, Statement | ≡ | Natural language sentence, Figure, Reference, Algebraic expression |

**Fig. 3.** Simplified Document Type

mented by the generic method *Doc_synthesis* which when invoked by a goal ⇐ *Doc_synthesis([spec,proof],docterm)*, given a pair *[spec,proof]*, causes the instantiation of the metavariable `docterm` by a document term that describes the document structure. Scripts are provided that convert document terms of



type *document* into the corresponding XML syntax. The resulting documents can be processed by the **t**book tools. The synthesised method is formalised by higher-order predicates which relate programming types and document types. A method which creates the basic document structure is outlined in Figure 4 and need to be augmented by the specifications for the methods *Doc_syn_theorem*, *Doc_syn_proof*, *Doc_syn_lemmata*:

- *Doc_syn_theorem* synthesises theorem content from formal specifications.
- *Doc_syn_proof* generates the proof content and a set of proof cases which shall be treated as lemmata.
- *Doc_syn_lemmata* generates the presentation for the proof cases.

$$Doc\_syn([spec, proof], doc) \Leftarrow \quad Doc\_header(header) \wedge$$
$$Doc\_body([spec, proof], body) \wedge$$
$$doc = Document(att, [header, body]).$$
$$Doc\_body([spec, proof], body) \Leftarrow Doc\_theorem(spec, theo) \wedge$$
$$Doc\_proof(proof, doc\_proof, subcases) \wedge$$
$$Doc\_syn\_lemmata(subCases, seq\_lem\_proof)) \wedge$$
$$body = Body(att, [theo, doc\_proof, seq\_lem\_proof].$$

**Fig. 4.** Skeleton of the document synthesis method

The above synthesis skeleton prescribes the main structure of the document to be synthesised. Specifications are mapped into theorems and programs into proofs. However, the content of these elements needs to be determined in more detail. It requires the partition of the specifications and their proofs into manageable parts, each part should be presented adequately, avoiding low-level details, but compiling all the information which is necessary for the comprehension of a proof part and required to keep the coherence between the parts. Each of the 3 methods *Doc_syn_theorem, Doc_syn_proof, Doc_syn_lemmata* is realised as a divide-and-conquer strategy. The specifications of the generic methods consist of metarules which define how to decompose the program synthesis fragments and how to compose the document fragments. Synthesis involves various decisions which are based on context-specific knowledge:

- how to decompose the program into subprograms which are represented as one major step;
- when to use graph visualisations or textual explanations for proof parts;
- which parts of the proof graph to split and to treat separately as lemmata.

The knowledge base of the document synthesiser can be augmented by heuristic layout knowledge, e.g., "acceptable" sizes for decision trees. The knowledge base entails parameterised text templates which are associated with corresponding objects of the program synthesis theory. The graph visualisation tool graphviz [Res05] is called on demand to generate graph layouts in the desired format.



## 6    Experimental Results

A previous version of SEAMLESS included generic methods for synthesis and machine learning which were devised independently of the data structure used for knowledge encoding. Domain specific knowledge was provided through an intermediate component as definitions for open generic predicates. The system was used to prove $V_3(22) \leq 28$ which constructed a counter example for a published theorem. Although the time for constructing the algorithm that has verified $V_3(22) \leq 28$ was not exactly measured, the computation had taken several days on a Sun Solaris workstation.

The system has been partially reimplemented by reusing the formally correct generic methods, however, changing the implementation of the data structures and the knowledge representation format. The knowledge representation schema and graph-based data structure which have been devised in this paper have been used to structure and implement the specialised domain knowledge about the selection problem. Indexing techniques for graphs have been introduced to tackle the huge search complexities structure. The results of these changes are

1. a radical improvement of the search complexities;
2. more abstract well-defined metalevel proofs with a higher degree of reuse which is reflected by references and links to objects and theorems;
3. a clearer distinction between informal or formal knowledge which has been derived outside of the system and automatically derived knowledge.

This implementation was used to re-synthesise algorithms for small n and it could be automatically proven that $V_i(n) \leq H_i(n), 1 \leq i \leq 6, 1 \leq n \leq 14$. In addition, the implementation was used to automatically synthesise various new complex algorithms which prove $V_4(21) \leq 30, V_4(23) \leq 33, V_4(24) \leq 34$, $V_4(25) \leq 35$ and $V_4(26) \leq 26$ and confirm the number-theoretic hypothesis for the selection problem.

The re-implementation was experimentally analysed on an Athlon 3200 64 bit computer with 1 GB memory and on a 1.3 GHZ Intel Centrino laptop with 512 MB memory. In this experimental application scenario, Yap Prolog turned out to be the fastest Prolog compared against Sicstus, GNU, B and SWI Prolog. In Figure 6 some complexity indicators for automated synthesis are collected: 1) a small example whose solution is described in the classic book by Knuth; 2) a more complex example; 3) the previously discovered selection algorithm which took days and now can be handled within seconds; 4) much more complex problems whose solutions improve the known upper bounds and confirm the stated hypothesis. The test environment was as follows: a) Centrino + Yap Prolog, b) Centrino + SWI Prolog, c) Athlon 3200 + Yap Prolog, d) Athlon 3200 + SWI Prolog. Some experimental results are summarised in Figure 5. The CPU time is measured in seconds. During the synthesis process, for each case solution, a canonical form is computed that represents the solution up to isomorphism. The abstracted case solutions can be reused and adapted. The number of generated case solutions for isomorphism classes is given in the last column of the table. A missing entry means that a solution couldn't be constructed within a couple of



hours. It should be noted that these indicators are snapshots of an implementation in progress. Each minor modification may alter the search space. This can decrease or increase the complexities to solve a problem drastically.

| $Spec$ | a | b | c | d | #examined isomorphism classes |
|---|---|---|---|---|---|
| $V_4(7) \leq 10$ | 0.026 | 0.05 | 0.014 | 0.01 | 68 |
| $V_3(22) \leq 28$ | 1.87 | 8.6 | 0.84 | 2.93 | 3,062 |
| $V_4(14) \leq 21$ | 9.0 | 56.0 | 3.3 | 18.1 | 12,122 |
| $V_4(21) \leq 30$ | 52.0 | 996.4 | 19.3 | 321.2 | 61,859 |
| $V_4(23) \leq 33$ | 288.9 | 16,239.7 | 107.1 | 5,210.2 | 277,178 |
| $V_4(24) \leq 34$ | 136.8 | 6,163.3 | 50.4 | 1,951.6 | 127,572 |
| $V_4(25) \leq 35$ | 390.4 | – | 146.5 | 13,034.3 | 362,458 |
| $V_4(26) \leq 36$ | – | – | 525.1 | – | 1,048,665 |

**Fig. 5.** Complexity indicators for correct search

In all experimental analysed cases, it can be experimentally verified that reuse of derived knowledge from previous proof searches when solving new problems improves the search complexities by up to 25%. Although the system is able to examine around 1,500,000 isomorphic states in less than one half hour and store them in 1 GB memory, the complexities of the selection problem fairly soon exceed the capabilities of the system. The experiments also showed that when using the computers to full capacity, odd runtime errors occur, e.g., uninstantiated variables, which cannot attributed to programming errors.

The reuse and knowledge-based synthesis approach generates proof graphs whose size has been decreased enormously compared to the uniform approaches. However, taking the constructed selection algorithms, in most of the more complex cases, the sizes of the generated decision graphs are still unsatisfying, because they cannot be manually checked with reasonable effort.

The generic synthesis method is based upon a 3-valued logic. Heuristic knowledge marked with the truth value $Maybe$ can be provided. In order to decrease the sizes of the proof graphs various incorrect lower bounds were experimentally added to the knowledge base. The simplest form of an incorrect lower bound is a depth restriction. The incorrect lower bounds restrict the search space. Employing these heuristics, an unsuccesful search may yield the result $Maybe$. Using heuristic search a solution for $V_4(24) \leq 34$ could be constructed within 6.5 secs, having examined 13695 isomorphism classes and having used Yap Prolog on the AMD Athlon 3200. The size of this proof graph was substantially decreased.

The generated graph is still too large to be easily comprehensible, therefore it is embedded in a system-generated documentation which provides a mathematically skilled reader background information about the proofs, such as cases which can be reduced to problems known from the literature, or are solved using the dedicated lemmata. The automatically synthesised documentation of this automatically synthesised algorithm is presented in appendix A.



A major advantage of combining higher-order program synthesis and document synthesis is that the generated proof terms describe programs at the algorithmic abstraction level. There is no need to raise the level of abstraction or to integrate a planning module into the document synthesis component. The synthesised document is a direct mapping from specifications and proofterms. An improved knowledge base of the system directly causes an improved documentation. Comprehensibility depends on the size of the automatically synthesised proof graphs. For small $i, n$ the automated co-synthesis of algorithms and their documentation achieves human-comprehensible documents. For complex problems, synthesis of proof terms of reasonable sizes is an experimental endeavour.

## 7  Conclusions and Related Work

In [Kre93], it has been stated "We believe that a *formalization of the metatheory of programming* is one of the most important steps towards the development of program synthesisers which are flexible and reliable and whose derivations are both formally correct and comprehensible for human programmers." As a step towards this objective, this paper has experimentally analysed the interplay between mathematical documents and knowledge-based algorithm synthesis based upon a metatheory for structuring problem solving knowledge. Structuring elements are identified to formalise algorithm design knowledge which is implicit in scientific documents and include abstractions like pre- and postconditions based upon the Floyd-Hoare program logics. Types, higher-order predicates and synthesis methods have been defined. The resulting framework is more fine-grained than related approaches for program synthesis which are based upon higher-order logic or proof planning [Kre93,IS04,RicCA]. The SEAMLESS framework also aims at the formalisation of heuristic problem solving strategies and resource restrictions. The framework devised has been applied to structure comprehensive knowledge relating to the selection problem and encode it for use in a knowledge base. The knowledge base has been used for the synthesis of complexity bounded algorithms using generic methods that were previously devised. Extended graphterms are used as the core data structure for knowledge encoding. They provide the key for the efficiency of the implementation as they support various optimisation techniques. The huge search complexities are tackled by a combination of indexing techniques, isomorphism abstraction, machine learning and knowledge reuse through references. It was possible to synthesise several new complex selection algorithms which improve known bounds and confirm our hypothesis.

Based upon the higher-order synthesis framework, a method that synthesises human-comprehensible XML-based documents from specifications and constructed proofs has been devised and implemented. The synthesised documents include diagrams and references to reused knowledge. The method has been applied for the automatic synthesis of documentation for automatically synthesised algorithms. It allows logic-based synthesis of documents on a higher abstraction level than a number of current low-level XML-based approaches which aim at the exchange or presentation of arbitrary mathematical documents, Website synthe-



sis or program documentation and merely exploit the correspondences between XML data and logical terms, e.g., [BDHG,LR03,SR01]. The presented framework distinguishes itself from related approaches (cf. [Pec04,GKP96,Oks05]) in which computer have been used for the complexity analysis of unsolved mathematical problems. The results synthesised by SEAMLESS are constructive, comprehensible and can be manually checked for correctness, provided that the sizes of the generated proof graphs are reasonable. In one case, it was possible to synthesise a new complex algorithm of reasonable length, and, thus the synthesised documentation is comprehensible (cf. Appendix A).

Our future work will include experimental work on automated abstraction to decrease the size of the generated algorithms and to get more abstract definitions for series of proof steps or comprehensive decision trees. The objective is to discover rules which describe new correct algorithms for infinitely many $n$.

## References


[Aig82]   M. Aigner. Selecting the top three elements. *Discrete Applied Mathematics*, 4:247–267, 1982.

[BDHG]   Dietmar A. Seipel, Bernd D. Heumesser and Ulrich Güntzer. An expert system for the flexible processing of XML-based mathematical knowledge in a Prolog-environment.

[BFP+73]   M. Blum, R.W. Floyd, V. Pratt, R.L. Rivest, and R.E. Tarjan. Time bounds for selection. *J. Comp. Syst. Sci.*, 7:448–461, 1973.

[BJ85]   S.W. Bent and J.W. John. Finding the median requires 2n comparisons. In *Proc. 17th ACM Symposium Theory of Computing*, pages 213–216, 1985.

[Bro05]   Torsten Bronger. The tbook system for XML authoring. http://tbookdtd.sourceforge.net, 25.9.2005.

[Dev05]   Keith Devlin. Last doubts removed about the proof of the four color theorem. http://www.maa.org/devlin/devlin_01_05.html, January 2005.

[EN96]   J. Eusterbrock and M. Nicolaides. The visualization of constructive proofs by compositional graph layout: A world-wide web interface. *Proc. CADE Visual Reasoning Workshop, Rutgers University*, 1996.

[Eus85]   J. Eusterbrock. Ein rekursiver Ansatz zur Bestimmung der Anzahl von Vergleichen bei kombinatorischen Selektionsproblemen. Diplomarbeit, Universität Dortmund, 1985.

[Eus92a]   J. Eusterbrock. Errata to "Selecting the top three elements" by M. Aigner: A Result of a computer assisted proof search. *Discrete Applied Mathematics*, 41:131–137, 1992.

[Eus92b]   J. Eusterbrock. *Wissensbasierte Verfahren zur Synthese mathematischer Beweise: Eine kombinatorische Anwendung*, volume 10 of *DISKI*, 1992.

[Eus95]   J. Eusterbrock. SEAMLESS: Knowledge based evolutionary system synthesis. *ERCIM News*, 23, October 1995.

[Eus97]   J. Eusterbrock. Canonical term representations of isomorphic transitive DAGs for efficient knowledge-based reasoning. In *Proceedings of the International KRUSE Symposium, Knowledge Retrieval, Use and Storage for Efficiency*, pages 235–249, 1997.

[Eus01]   J. Eusterbrock. Knowledge mediation in the world-wide web based upon labelled dags with attached constraints. *Electronic Transactions on Artificial Intelligence*, 5:"http://www.ida.liu.se/ext/epa/ej/etai/2001/D", 2001.





[FG79]    F. Fussenegger and H.N. Gabow. A counting approach to lower bounds for selection problems. *J. Assoc. Comput. Mach.*, 26:227–238, 1979.

[FR75]    R.W. Floyd and R.L. Rivest. Expected time bounds for selection. *Comm. ACM*, 18:165–172, 1975.

[GKP96]   William Gasarch, Wayne Kelly, and William Pugh. Finding the ith largest of n for small i,n. *SIGACT News*, 27(2):88–96, 1996.

[Hoa69]   C.A.R. Hoare. An axiomatic basis for computer programming. *CACM*, 12(10):576–581, 1969.

[HS69]    A. Hadian and M. Sobel. Selecting the t-th largest using binary errorless comparisons. In P. Erdös, A. Renyi, and V.T. Sos, editors, *Combinatorial Theory and its Applications II*, pages 585–600. North Holland, 1969.

[IS04]    A. Ireland and J. Stark. Combining proof plans with partial order planning for imperative program synthesis. *Journal of Automated Software Engineering*, Accepted for publication, 2004.

[Kis64]   S. S. Kislitsyn. On the selection of the k-th element of an ordered set by pairwise comparisons. *Sib. Mat. Z.*, 5:557–564, 1964.

[Knu73b]  D.E. Knuth. *Sorting and Searching*, volume 3 of *The Art of Computer Programming*. Addison Wesley, Reading, MA, 1973.

[Kre93]   Chr. Kreitz. Metasynthesis - deriving programs that develop programs. Habilitationsschrift, Technische Hochschule Darmstadt, 1992.

[Lam91]   Clement Lam. The search for a finite projective plane of order 10. *American Mathematical Monthly*, 98:305–318, 1991.

[LR03]    S. Leung and D. Robertson. Automated website synthesis. http://www.ukuug.org/events/linux2003/papers/leung.pdf, 2003.

[MP82]    J.I. Munro and P.V. Poblete. A lower bound for determining the median. Technical report, University of Waterloo Research Report CS-82-21, 1982.

[Oks05]   Kenneth Oksanen. Selecting the ith largest of n elements. http://www.cs.hut.fi/ cessu/selection/, 2005.

[Pec04]   Marcin Peczarski. New results in minimum comparison sorting. *Algorithmica*, 40:133–145, 2004.

[Res05]   AT&T Research. Graphviz - graph visualization software. http://www.graphviz.org/, 25.9.2005.

[RH84]    P.V. Ramanan and L. Hyafil. New algorithms for selection. *J. of Alg.*, 1:557–578, 1984.

[RicCA]   C.S Richardson, J.D. Proof planning and program synthesis: a survey. In *Logic-Based Program Synthesis: State-of-the-Art & Future Trends, AAAI 2002 Spring Symposium*, March 25-27, 2002, Stanford University, CA.

[SPP76]   A. Schönhage, M. Paterson, and N. Pippenger. Finding the median. *J. Comp. System Sci.*, 13:184–199, 1976.

[SR01]    Johann Schumann and Peter Robinson. [] or success is not enough: Current technology and future directions in proof presentation. http://www.cs.bham.ac.uk/ mmk/events/ijcar01-future/, 2001.

[Yao74]   F.F. Yao. On lower bounds for selection problems. Technical report, TR-121, MIT, Project Mac, Cambridge, Mass., 1974.

[Yap76]   C.K. Yap. New upper bounds for selection. *Comm. ACM*, 19:501–508, 1976.


# A   Appendix





# Automatically Synthesised Selection Algorithm *

## © Jutta Eusterbrock

**theorem 1.** $V_4(24) \leq 34$

*Proof.* Let KEYS be a totally ordered set, |KEYS| =24. The 4 -th largest element of KEYS is computed by Algorithm 1. The computation takes in the worst-case at most 34 comparisons.

**algorithm 1.** *1. Partition the set KEYS into disjoint subsets |K1|=16,|K2|=8. Determine the maxima of K1,K2 by setting-up balanced tournaments. The resulting poset is isomorphic to the poset as shown in Figure 1. For setting-up the balanced tournaments 22 comparisons are needed.*

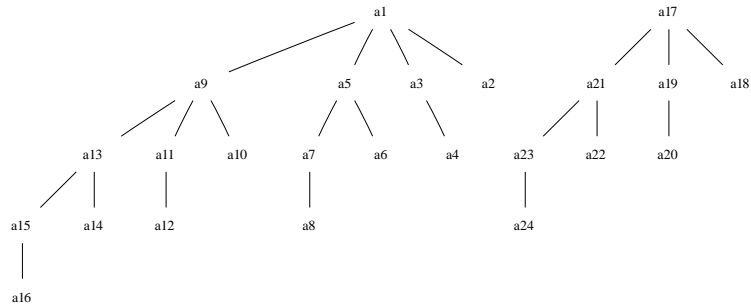

**Figure 1:** Balanced Tournaments of Set Partition

*2. Perform the comparisons according to Figure 2. The nodes in the decision graph represent:*

*(a) Comparisons $X : Y$, shown as circles. The left child node represents the case $X > Y$ and the right one the case $X < Y$.*

*(b) Subgraph place holders, shown by references to a diamond which correspond to solutions which can be obtained by instantiating theorems and algorithms*

---

*This is - apart from this footnote - an automatically synthesised documentation of a new, automatically synthesised algorithm *Select*(4, 24) with the worst-case complexity $V_4(24) \leq 34$. Based on the proof structure derived during the synthesis, the system selected, instantiated, and composed appropriate text templates to generate the explanatory text and document structure. The graphs were visualised by the graphviz system. The proof was checked for correctness by the present author.





*known from published literature. The cases 1, 27, 48, 49, 739, 743, 744, 873, 874, 7772, 7773, 7823, 7824, 7839, 7840 are instances of published theorems.*

(c) *Place holders for calls to simplification functions, especially isomorphism functions, which are denoted by boxes. There are two type of function calls. Firstly, solutions for the simplified subcases are represented by a subgraph which is then referenced. This concerns case 7844. Secondly, in some situations more detailed explanations for the simplified subcases, e.g, lemmata, have been generated. In the given decision graph, the cases 876 and 7775 are handled separately.*

3. *Retrieve the set of re-used algorithms and theorems:*

- *Case 876 is described as follows. Let P be the partially ordered set as visualised in Figure 3. It needs to be proven $V_3(P) \leq 8$. The specification is up-to-isomorphism handled by Lemma 1.*

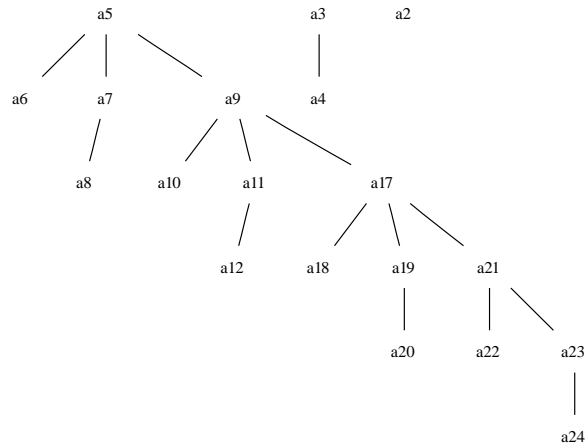

**Figure 3:** Poset P

- *Case 7775 is described as follows. Let P be the partially ordered set as visualised in Figure 4. It needs to be proven $V_3(P) \leq 8$. The specification is up-to-isomorphism handled by Lemma 2.*





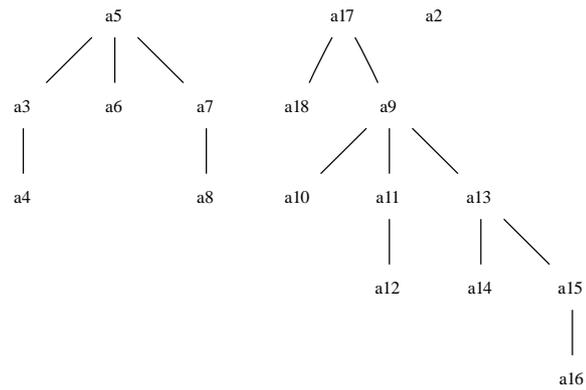

**Figure 4:** Poset P

□

**lemma 1.** *Let P be a partially ordered set as visualised in Figure 5. The 3 largest element of P can be computed by at most 8 comparisons.*

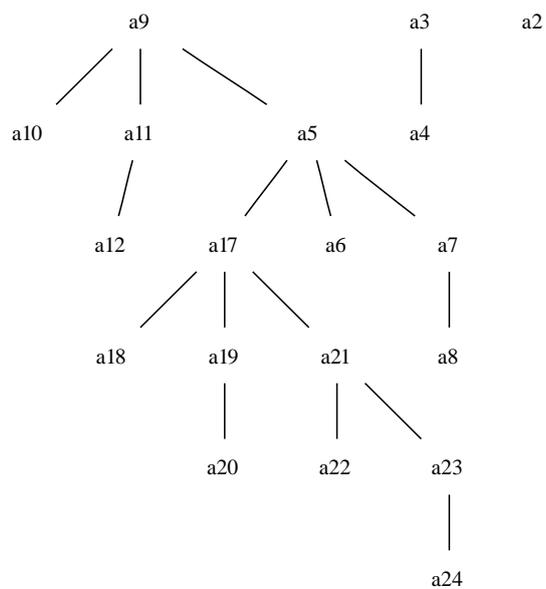

**Figure 5:** Poset P

*Proof.*





**algorithm 2.**    *1.  Compare a9 and a3.*

(a) *a9 > a3.  Let P be the corresponding partially ordered set.  P satisfies the precondition of the algorithm Kislitsyn (1964, kislitsyn).  Hence, selecting the 2-nd largest element takes at most f(5,14.0)=7 comparisons.*

(b) *a9 < a3.  Let P be the corresponding partially ordered set.  P satisfies the precondition of the algorithm Kislitsyn (1964, kislitsyn).  Hence, selecting the 2-nd largest element takes at most f(3,10.0)=5 comparisons.*

<div align="right">□</div>

**lemma 2.**  *Let P be a partially ordered set as visualised in Figure 6.  The 3 largest element of P can be computed by at most 8 comparisons.*

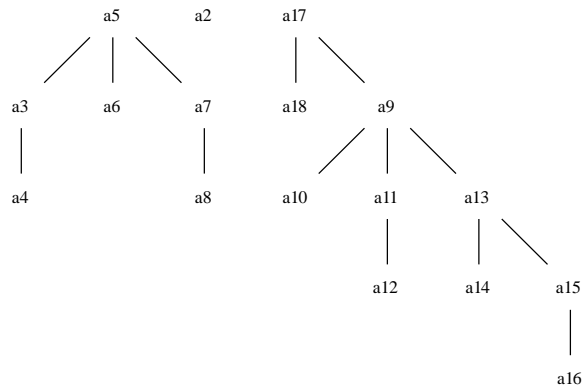

**Figure 6:** Poset P

*Proof.*

**algorithm 3.**    *1.  Compare a5 and a17.*

(a) *a5 > a17.  Let P be the corresponding partially ordered set.  P satisfies the precondition of the algorithm Kislitsyn (1964, kislitsyn).  Hence, selecting the 2-nd largest element takes at most f(5,10.0)=7 comparisons.*

(b) *a5 < a17.  Let P be the corresponding partially ordered set.  P satisfies the precondition of the algorithm Kislitsyn (1964, kislitsyn).  Hence, selecting the 2-nd largest element takes at most f(4,18.0)=7 comparisons.*

<div align="right">□</div>

# References

Kislitsyn, S. S., 1964: *On the selection of the k-th element of an ordered set by pairwise comparisons.* Sib Mat Z, **5**, pp. 557–564. 4



# Mixing Finite Success and Finite Failure
# in an Automated Prover

Alwen Tiu[1], Gopalan Nadathur[2], and Dale Miller[3]

[1] INRIA Lorraine/LORIA
[2] Digital Technology Center and Dept of CS, University of Minnesota
[3] INRIA & LIX, École Polytechnique

**Abstract.** The operational semantics and typing judgements of modern programming and specification languages are often defined using relations and proof systems. In simple settings, logic programming languages can be used to provide rather direct and natural interpreters for such operational semantics. More complex features of specifications such as names and their bindings, proof rules with negative premises, and the exhaustive enumeration of state spaces, all pose significant challenges to conventional logic programming systems. In this paper, we describe a simple architecture for the implementation of deduction systems that allows a specification to interleave between finite success and finite failure. The implementation techniques for this prover are largely common ones from higher-order logic programming, i.e., logic variables, (higher-order pattern) unification, backtracking (using stream-based computation), and abstract syntax based on simply typed $\lambda$-terms. We present a particular instance of this prover's architecture and its prototype implementation, Level 0/1, based on the dual interpretation of (finite) success and finite failure in proof search. We show how Level 0/1 provides a high-level and declarative implementation of model checking and bisimulation checking for the (finite) $\pi$-calculus.

## 1 Introduction

The operational semantics and typing judgements of modern programming and specification languages are often defined using relations and proof systems, e.g., in the style of Plotkin's structural operational semantics. In simple settings, higher-order logic programming languages, such as $\lambda$Prolog and Twelf, can be used to provide rather direct and natural interpreters for operational semantics. However, such logic programming languages can provide little more than animation of semantic descriptions: in particular, reasoning about specified languages has to be done outside the system. For instance, checking bisimulation in process calculi needs analyzing all the transition paths a process can potentially go through. To add to the complication, modern language specifications often make use of complex features such as variable bindings and the notion of *names* (as in the $\pi$-calculus [MPW92]), which interferes in a non-trivial way with case analyses. These case analyses cannot be done directly inside the logic



programming system, not in a purely logical way at least, even though they are simply enumerations of answer substitutions. In this paper, we describe an extension to logic programming with logically sound features which allow us to do some modest automated reasoning about specifications of operational semantics. This extension is more conceptual than technical, that is, the implementation of the extended logic programming language uses only implementation techniques that are common to logic programming, i.e., logic variables, higher-order pattern unification, backtracking (using stream-based computation) and abstract syntax based on typed $\lambda$-calculus.

The implementation described in this paper is based on the logic $FO\lambda^{\Delta\nabla}$ [MT03], which is a logic based on a subset of Church's Simple Theory of Types but extended with fixed points and the $\nabla$ quantifier. In $FO\lambda^{\Delta\nabla}$ quantification over propositions is not allowed but quantifiers can otherwise range over variables of higher-types. Thus the terms of the logic can be simply typed terms, which can be used to encode the $\lambda$-tree syntax of encoded objects in an operational semantics specification. This style of encoding is a variant of *higher-order abstract syntax* in which meta-level $\lambda$-abstractions are used to encode object-level variable binding. The quantifier $\nabla$ is first introduced in [MT03] to help encode the notion of "generic judgment" that occurs commonly when reasoning with $\lambda$-tree syntax.

The logical extension to allow fixed points is done through a proof theoretical notion of *definitions* [SH93,Eri91,Gir92,Stä94,MM00]. In a logic with definitions, an atomic proposition may be defined by another formula (which may contain the atomic proposition itself). Proof search for a defined atomic formula is done by unfolding the definition of the formula. A provable formula like $\forall X.pX \supset qX$, where $p$ and $q$ are some defined predicates, expresses the fact that for every term $t$ for which there is a successful computation (proof) of $pt$, there is a computation (proof) of $qt$. Towards establishing the truth of this formula, if the computation tree associated with $p$ is finite, we can effectively enumerate all its computation paths and check the provability of $qt$ for each path. Note that if the computation tree for $p$ is empty ($pt$ is not provable for any $t$) then $\forall X.pX \supset qX$ is (vacuously) true. In other words, *failure* in proof search for $pX$ entails *success* in proof search for $pX \supset qX$. The analogy with *negation-as-failure* in logic programming is obvious: if we take $qX$ to be $\bot$ (false), then provability of $pX \supset \bot$ corresponds to success in proof search for $not(pX)$ in logic programming. This relation between negation-as-failure in logic programming and negation in logic with definitions has been observed in [HSH91,Gir92]. In the implementation of $FO\lambda^{\Delta\nabla}$, the above observation leads to a neutral view on proof search: If proof search for a goal $A$ returns a non-empty set of answer substitutions, then we have found a proof of $A$. On the other hand, if proof search for $A$ returns an empty answer set, then we have found a proof for $\neg A$. Answer substitutions can thus be interpreted in a dual way depending on the context of proof search; see Section 3 for more details.

The rest of the paper is organized as follows. In Section 2, we give an overview of the logic $FO\lambda^{\Delta\nabla}$. Section 3 describes an implementation of a fragment of



$FO\lambda^{\Delta\nabla}$, the Level-0/1 prover, which is based on a dual interpretation of failure/success in proof search. Section 4 discusses the treatment of variables in the Level-0/1 prover, in particular, it discusses the issues concerning the interaction between *eigenvariables* and logic variables. Section 5 specifically contrasts the expressiveness of Level-0/1 over what is available in $\lambda$Prolog. Section 6 gives a specification of the operational semantics for the $\pi$-calculus and shows how Level-0/1 can turn that specification naturally into a checker for (open) bisimulation. Section 7 provides a specification of modal logic for the $\pi$-calculus, which the Level-0/1 prover can use to do model checking for that process calculus. These two specifications involving the $\pi$-calculus illustrate the use of the $\nabla$-quantifier to help capture various restrictions of names in the $\pi$-calculus. Section 8 discusses the components of proof search implementation and outlines a general implementation architecture for $FO\lambda^{\Delta\nabla}$. Section 9 discusses future work. An extended version of this paper is available on the web, containing more examples and more detailed comparison with logic programming.

## 2 Overview of the logic $FO\lambda^{\Delta\nabla}$

The logic $FO\lambda^{\Delta\nabla}$ [MT03] (pronounced "fold-nabla") is presented using a sequent calculus that is an extension of Gentzen's system LJ [Gen69] for intuitionistic logic. The first extension to LJ is to allow terms to be simply typed $\lambda$-terms and to allow quantification to be over all types not involving the predicate type (in Church's notation [Chu40], the types of quantified variables do not contain $o$). A *sequent* is an expression of the form $B_1, \ldots, B_n \vdash B_0$ where $B_0, \ldots, B_n$ are formulas and the elongated turnstile $\vdash$ is the sequent arrow. To the left of the turnstile is a multiset: thus repeated occurrences of a formula are allowed. If the formulas $B_0, \ldots, B_n$ contain free variables, they are considered universally quantified outside the sequent, in the sense that if the above sequent is provable then every instance of it is also provable. In proof theoretical terms, such free variables are called *eigenvariables*. Eigenvariable can be used to encode the dynamics of abstraction in the operational semantics of various languages. However, for reasoning about certain uses of abstraction, notably the notion of *name restriction* in $\pi$-calculus, eigenvariables do not capture faithfully the intended meaning of such abstractions. To address this problem, in the logic $FO\lambda^{\Delta\nabla}$ sequents are extended with a new notion of "local scope" for proof-level bound variables (see [MT03] for motivations and examples). In particular, sequents in $FO\lambda^{\Delta\nabla}$ are of the form

$$\Sigma \,;\, \sigma_1 \rhd B_1, \ldots, \sigma_n \rhd B_n \vdash \sigma_0 \rhd B_0$$

where $\Sigma$ is a *global signature*, i.e., the set of eigenvariables whose scope is over the whole sequent, and $\sigma_i$ is a *local signature*, i.e., a list of variables scoped over $B_i$. We shall consider sequents to be binding structures in the sense that the signatures, both the global and local ones, are abstractions over their respective scopes. The variables in $\Sigma$ and $\sigma_i$ will admit $\alpha$-conversion by systematically changing the names of variables in signatures as well as those in their scope,



$$\frac{\Sigma, \sigma \vdash t : \gamma \qquad \Sigma\,;\, \sigma \triangleright B[t/x], \Gamma \vdash \mathcal{C}}{\Sigma\,;\, \sigma \triangleright \forall_\gamma x.B, \Gamma \vdash \mathcal{C}} \ \forall\mathcal{L} \qquad \frac{\Sigma, h\,;\, \Gamma \vdash \sigma \triangleright B[(h\ \sigma)/x]}{\Sigma\,;\, \Gamma \vdash \sigma \triangleright \forall x.B} \ \forall\mathcal{R}$$

$$\frac{\Sigma, h\,;\, \sigma \triangleright B[(h\ \sigma)/x], \Gamma \vdash \mathcal{C}}{\Sigma\,;\, \sigma \triangleright \exists x.B, \Gamma \vdash \mathcal{C}} \ \exists\mathcal{L} \qquad \frac{\Sigma, \sigma \vdash t : \gamma \qquad \Sigma\,;\, \Gamma \vdash \sigma \triangleright B[t/x]}{\Sigma\,;\, \Gamma \vdash \sigma \triangleright \exists_\gamma x.B} \ \exists\mathcal{R}$$

$$\frac{\Sigma\,;\, (\sigma, y) \triangleright B[y/x], \Gamma \vdash \mathcal{C}}{\Sigma\,;\, \sigma \triangleright \nabla x\ B, \Gamma \vdash \mathcal{C}} \ \nabla\mathcal{L} \qquad \frac{\Sigma\,;\, \Gamma \vdash (\sigma, y) \triangleright B[y/x]}{\Sigma\,;\, \Gamma \vdash \sigma \triangleright \nabla x\ B} \ \nabla\mathcal{R}$$

**Fig. 1.** The introduction rules for quantifiers in $FO\lambda^{\Delta\nabla}$.

following the usual convention of the $\lambda$-calculus. The meaning of eigenvariables is as before, only that now instantiation of eigenvariables has to be capture-avoiding, with respect to the local signatures. The variables in local signatures act as locally scoped *generic constants*, that is, they do not vary in proofs since they will not be instantiated. The expression $\sigma \triangleright B$ is called a *generic judgment* or simply a *judgment*. We use script letters $\mathcal{A}$, $\mathcal{B}$, etc., to denote judgments. We write simply $B$ instead of $\sigma \triangleright B$ if the signature $\sigma$ is empty.

The logical constants of $FO\lambda^{\Delta\nabla}$ are $\forall_\gamma$ (universal quantifier), $\exists_\gamma$ (existential quantifier), $\nabla_\gamma$ (nabla quantification), $\wedge$ (conjunction), $\vee$ (disjunction), $\supset$ (implication), $\top$ (true) and $\bot$ (false). The subscript for the three quantifiers is the type of the variable they are intended to bind: in particular, $\gamma$ can range over any type not containing the predicate type. Usually this type subscript is suppressed. The inference rules for the three quantifiers of $FO\lambda^{\Delta\nabla}$ are given in Figure 1. The introduction rules for propositional connectives are straightforward generalization of LJ: in particular, local signatures are distributed over the subformulas of the main formula (reading the rules bottom-up). The complete set of rules for $FO\lambda^{\Delta\nabla}$ is given in Figure 10 at the end of this paper.

During the search for proofs (reading rules bottom up), the right-introduction rule for $\forall$ and the left-introduction rule for $\exists$ place new variables into the global signature: the left and right introduction rules for $\nabla$ place new variables into the local signature. In the $\forall\mathcal{R}$ and $\exists\mathcal{L}$ rules, raising [Mil92] is used when replacing the bound variable $x$, which can range over the variables in both the global signature and the local signature $\sigma$, with the variable $h$ that can only range over variables in the global signature: so as not to miss substitution terms, the variable $x$ is replaced by the term $(h\,x_1 \ldots x_n)$, which we shall write simply as $(h\,\sigma)$, where $\sigma$ is the list $x_1, \ldots, x_n$ ($h$ must not be free in the lower sequent of these rules). In $\forall\mathcal{L}$ and $\exists\mathcal{R}$, the term $t$ can have free variables from both $\Sigma$ and $\sigma$. This is presented in the rule by the typing judgment $\Sigma, \sigma \vdash t : \gamma$. The $\nabla\mathcal{L}$ and $\nabla\mathcal{R}$ rules have the proviso that $y$ is not free in $\nabla x\ B$.

Besides these introduction rules for logical constants, $FO\lambda^{\Delta\nabla}$ additionally allows the introduction of atomic judgments, that is, judgments of the form $\sigma \triangleright A$ where $A$ is an atomic formula. To each atomic judgment, $\mathcal{A}$, we associate a judgment $\mathcal{B}$ called the *definition* of $\mathcal{A}$. The introduction rule for the judgment $\mathcal{A}$ is in effect done by replacing $\mathcal{A}$ with $\mathcal{B}$ during proof search. This notion of



definitions is an extension of work by Schroeder-Heister [SH93], Eriksson [Eri91], Girard [Gir92], Stärk [Stä94] and McDowell and Miller [MM00] and allows for modest reasoning about the fixed points of definitions.

**Definition 1.** *A* definition clause *is written* $\forall \bar{x}[p\,\bar{t} \stackrel{\triangle}{=} B]$, *where $p$ is a predicate constant, every free variable of the formula $B$ is also free in at least one term in the list $\bar{t}$ of terms, and all variables free in $p\,\bar{t}$ are contained in the list $\bar{x}$ of variables. The atomic formula $p\,\bar{t}$ is called the* head *of the clause, and the formula $B$ is called the* body. *The symbol $\stackrel{\triangle}{=}$ is used simply to indicate a definitional clause: it is not a logical connective.*

*Let $\forall_{\gamma_1} x_1 \ldots \forall_{\gamma_n} x_n . H \stackrel{\triangle}{=} B$ be a definition clause. Let $y_1, \ldots, y_m$ be a list of variables of types $\alpha_1, \ldots, \alpha_m$, respectively. The* raised definition clause *of $H$ with respect to the signature $\{y_1 : \alpha_1, \ldots, y_m : \alpha_m\}$ is defined as*

$$\forall h_1 \ldots \forall h_n . \bar{y} \triangleright H\theta \stackrel{\triangle}{=} \bar{y} \triangleright B\theta$$

*where $\theta$ is the substitution $[(h_1\,\bar{y})/x_1, \ldots, (h_n\,\bar{y})/x_n]$ and $h_i$ is of type $\alpha_1 \to \ldots \to \alpha_m \to \gamma_i$. A* definition *is a set of definition clauses together with their raised clauses.*

The introduction rules for a defined judgment are displayed below. When applying the introduction rules, we shall omit the outer quantifiers in a definition clause and assume implicitly that the free variables in the definition clause are distinct from other variables in the sequent.

$$\frac{\{\Sigma\theta\,;\,\mathcal{B}\theta, \Gamma\theta \vdash \mathcal{C}\theta \mid \theta \in CSU(\mathcal{A}, \mathcal{H}) \text{ for some clause } \mathcal{H} \stackrel{\triangle}{=} \mathcal{B}\}}{\Sigma\,;\,\mathcal{A}, \Gamma \vdash \mathcal{C}} \; def\mathcal{L}$$

$$\frac{\Sigma\,;\,\Gamma \vdash \mathcal{B}\theta}{\Sigma\,;\,\Gamma \vdash \mathcal{A}} \; def R, \quad \text{where } \mathcal{H} \stackrel{\triangle}{=} \mathcal{B} \text{ is a definition clause and } \mathcal{H}\theta = \mathcal{A}$$

In the above rules, we apply substitutions to judgments. The result of applying a substitution $\theta$ to a generic judgment $x_1, \ldots, x_n \triangleright B$, written as $(x_1, \ldots, x_n \triangleright B)\theta$, is $y_1, \ldots, y_m \triangleright B'$, if $(\lambda x_1 \ldots \lambda x_n.B)\theta$ is equal (modulo $\lambda$-conversion) to $\lambda y_1 \ldots \lambda y_m.B'$. If $\Gamma$ is a multiset of generic judgments, then $\Gamma\theta$ is the multiset $\{J\theta \mid J \in \Gamma\}$. In the *def$\mathcal{L}$* rule, we use the notion of *complete set of unifiers* (CSU) [Hue75]. We denote by $CSU(\mathcal{A}, \mathcal{H})$ a complete set of unifiers for the pair $(\mathcal{A}, \mathcal{H})$, that is, for any unifier $\theta$ of $\mathcal{A}$ and $\mathcal{H}$, there is a unifier $\rho \in CSU(\mathcal{A}, \mathcal{H})$ such that $\theta = \rho \circ \theta'$ for some substitution $\theta'$. Since we allow higher-order terms in definitions, in certain cases there are no finite CSU's for a given unification problem. Thus, in the fully general case, *def$\mathcal{L}$* may have an infinite number of premises [MM00]. In all the applications of *def$\mathcal{L}$* in this paper, however, the terms involved in unification are those of *higher-order patterns* [Mil91,Nip93], that is, terms in which variables are applied only to distinct bound variables. Since higher-order pattern unification is decidable and unary (i.e., the most general unifiers exist if the unification is solvable), the set $CSU(\mathcal{A}, \mathcal{H})$ in this



case can be treated as being either empty or containing a single substitution which is the most general unifier. In this restricted setting, $def\mathcal{L}$ will have a finite number of premises (assuming as we shall that definitions are based on the raising of only a *finite* number of clauses). The signature $\Sigma\theta$ in $def\mathcal{L}$ denotes the signature obtained from $\Sigma$ by removing the variables in the domain of $\theta$ and adding the variables in the range of $\theta$. In the $def\mathcal{L}$ rule, reading the rule bottom-up, eigenvariables can be instantiated in the premise, while in the $def R$ rule, eigenvariables are not instantiated. The set that is the premise of the $def\mathcal{L}$ rule means that that rule instance has a premise for every member of that set: if that set is empty, then the premise is considered proved.

One might find the following analogy with logic programming helpful: if a definition is viewed as a logic program, then the $def R$ rule captures backchaining and the $def\mathcal{L}$ rule corresponds to *case analysis* on all possible ways an atomic judgment could be proved. The latter is a distinguishing feature between the implementation of $FO\lambda^{\Delta\nabla}$ discussed in Section 3 and logic programming. For instance, given the definition

$$\{pa \overset{\triangle}{=} \top, \quad pb \overset{\triangle}{=} \top, \quad qa \overset{\triangle}{=} \top, \quad qb \overset{\triangle}{=} \top, \quad qc \overset{\triangle}{=} \top\},$$

one can prove $\forall x.px \supset qx$: for all successful "computation" of $p$, there is a successful computation for $q$. Notice that by encoding logic programs as definitions, one can effectively encode *negation-as-failure* in logic programming using $def\mathcal{L}$ [HSH91], e.g., for the above program (definition), the goal $not(pc)$ in logic programming is encoded as the formula $pc \supset \bot$.

Two properties of $FO\lambda^{\Delta\nabla}$ are particularly important to note here. First, if a certain stratification of predicates within definitions is made (so that there is no circularity in defining predicates through negations) then cut-elimination and consistency can be proved [MT05,Tiu04]. Second, many inference rules are known to be *invertible*, in the sense that they can always be applied without the need for backtracking. Those rules include $def\mathcal{L}$, $\nabla\mathcal{L}$, $\nabla\mathcal{R}$, $\exists\mathcal{L}$, $\forall\mathcal{R}$, the right introduction rules for $\wedge$ and $\supset$, and the left introduction rules for $\wedge$ and $\vee$ [Tiu04]. The invertibility of these rules motivates the choice of the fragment of $FO\lambda^{\Delta\nabla}$ on which the Level-0/1 prover works.

## 3 Mixing success and failure in a prover

We now give an overview of an implementation of proof search for a fragment of $FO\lambda^{\Delta\nabla}$. This implementation, called *Level 0/1 prover*, is based on the dual interpretation of finite success and *finite failure* in proof search. In particular, the finite failure in proving a goal $\exists x.G$ should give us a proof of $\neg(\exists x.G)$ and vice versa. We experiment with a simple class of formulas which exhibits this duality. We first assume that all predicate symbols are classified as belonging to either level-0 or level-1 (via some mapping of predicates to $\{0,1\}$). Next consider



the following classes of formulas:

Level 0: $G := \top \mid \bot \mid A \mid G \wedge G \mid G \vee G \mid \exists x.G \mid \nabla x.G$
Level 1: $D := \top \mid \bot \mid A \mid D \wedge D \mid D \vee D \mid \exists x.D \mid \nabla x.D \mid \forall x.D \mid G \supset D$
atomic: $A := p\, t_1 \ldots t_n$

Here, atomic formulas $A$ in level 0 formulas must have predicates that have been assigned to level 0. Atomic formulas in level 1 formulas can have predicates of either level 0 and 1. Each definition clause $p\bar{t} \stackrel{\triangle}{=} B$ must be *stratified*, i.e., if $p$ is a level-0 predicate then $B$ should belong to the class level-0, otherwise if $p$ is a level-1 predicate then $B$ can be a level-0 or level-1 formula. In the current implementation, stratification checking and type checking are not implemented, so that we can experiment with a wider range of definitions than those for which the meta-theory is fully developed.

Notice that in the Level-1 formula, the use of implication is restricted to the form $G \supset D$ where $G$ is a Level-0 formula. Therefore, nested implication like $(A \supset B) \supset C$ is not allowed. The Level-0/1 prover actually consists of two separate subprovers, one for each class of formulas. Implementation of proof search for level-0 formula follows the standard logic-programming implementation for Horn clauses: it is actually the subset of $\lambda$Prolog based on Horn clauses but allowing also $\nabla$ quantification in the body of clauses. In this prover, existential quantifiers are instantiated with logic variables, $\nabla$-quantifiers are instantiated with scoped (local) constants (which have to be distinguished from eigenvariables), and *def R* is implemented via backchaining. For level-1 formulas, the non-standard case is when the goal is an implication, e.g., $G \supset D$. Proof search strategy for this case derives from the following observation: the left-introduction rules for level-0 formulas are all invertible rules, and hence can always be applied first. Proof search for an implicational goal $G \supset D$ therefore proceeds as follows:

**Step 1** Run the level-0 prover with the goal $G$, treating any level-1 eigenvariables as level-0 logic variables.

**Step 2** Collect all answer substitutions produced by Step 1 into a lazy stream of substitutions and for each substitution $\theta$ in this stream, proceed with proving $G\theta$. For example, if Step 1 fails, then this stream is empty and this step succeeds immediately.

In Step 1, we impose a restriction: the formula $G$ must not contain any occurrences of level-1 logic variables. If this restriction is violated, a runtime exception is returned and proof search is aborted. We shall return to this technical restriction in Section 4. This restriction on the occurrence of logic variable has not posed a problem for a number of applications, e.g., checking bisimulation and satisfiability of modal logic formulas for the $\pi$-calculus.

We claim the following soundness theorem for the provers architecture above: If Level-0/1 is given a definition and a goal formula and it successfully claims to have a proof of that goal (that is, the system terminates without a runtime error), then that goal follows from the definition also in the $FO\lambda^{\Delta\nabla}$ logic.



*Concrete syntax* The concrete syntax for Level 0/1 prover follows the syntax of
λProlog. The concrete syntax for logical connectives are as follows:

| | | | |
|---|---|---|---|
| ⊤ | `true` | ⊥ | `false` |
| ∧ | `&` (ampersand) or `,` (comma) | ∨ | `;` (semi-colon) |
| ∀ | `pi` | ∃ | `sigma` |
| ∇ | `nabla` | ⊃ | `=>` |

The λ-abstraction is represented in the concrete syntax using an infix back-
slash, with the body of a λ-abstraction is goes as far to the right as possible,
consistent with the existing parentheses: for example, $\lambda x \lambda f.fx$ can be written
as `(x\f\ f x)`. The order of precedence for the connectives is as follows (in
decreasing order): $\wedge, \vee, \supset, \{\forall, \exists, \nabla\}$. Follow the convention started by Church
[Chu40], the bound variable associated to a quantifier is actually a λ-abstraction:
for example, the logical expression $\forall x[p(x) \supset q(x)] \wedge p(a)$ can be encoded as the
`(pi x\ p x => q x) & (p a)`. Non-logical constants, such as 'not' (negation-
as-failure) and '!' (Prolog cut), are not implemented, while we do allow the
non-logical constant `print` for printing terms. Finally, we note that the percent
sign `%` starts a comment line.

The symbol $\triangleq$ separating the head and the body of a definition clause is writ-
ten as ':=' in the concrete syntax. For example, the familiar 'append' predicate
for lists can be represented as the following definition.

```
append nil L L.
append (cons X L1) L2 (cons X L3) := append L1 L2 L3.
```

As in λProlog, we use '.' (dot) to indicate the end of a formula. Identifiers
starting with a capital letter denote variables and those starting with lower-case
letter denote constants. Variables in a definition clause are implicitly quantified
outside the clause (the scope of such quantification is over the clause, so there
is no accidental mixing of variables across different clauses). A definition clause
with the body 'true' is abbreviated with the 'true' removed, e.g., the first clause
of append above is actually an abbreviation of `append nil L L := true`.

## 4 Eigenvariables, logic variables and ∇

The three quantifiers, ∀, ∃ and ∇, give rise to three kinds of variables dur-
ing proof search: eigenvariables, logic variables and "variables" generated by ∇.
Their characteristics are as follows: logic variables are genuine variables, in that
they can be instantiated during proof search. Eigenvariables are subject to in-
stantiation only in proving negative goals, while in positive goals they are treated
as scoped constants. Variables generated by ∇ are never instantiated and are
usually represented by λ-abstractions. Eigenvariables and logic variables share
similar data structures, and explicit raising is used to encode their dependency
on ∇-variables. The interaction between eigenvariables and logic variables is



more subtle. Consider the case where both eigenvariables and logic variables are present in a negative goal, for example, consider proving the goal

$$\forall x.\exists y.(px \wedge py \wedge x = y \supset \bot),$$

where $p$ is defined as $\{pa \stackrel{\triangle}{=} \top, pb \stackrel{\triangle}{=} \top, pc \stackrel{\triangle}{=} \top\}$. In proof search for this formula, we are asked to produce for each $x$, a $y$ such that $x$ and $y$ are distinct. This is no longer a unification problem in the usual sense, since we seek to cause a failure in unification, instead of success. This type of problem is generally referred to as *complement problems* or *disunification* [LC89], and its solution is not unique in general, even for the first-order case, e.g., in the above disunification problem, if $x$ is instantiated to $a$ then $y$ can be instantiated with either $b$ or $c$. In the higher-order case [MP03] the problem is considerably more difficult, and, hence, in the current implementation, we disallow occurrences of logic variables in negative goals.

In Figure 2, we show a sample session in Level 0/1 prover which highlights the differences between eigenvariables, logic variables, and $\nabla$-variables. The unification problem in the first two goals can be seen as the unification problem $\lambda x.x = \lambda x.(Mx)$. Notice that there is no difference between $\forall$ and $\nabla$ if the goal is level-0 (i.e., there is no implication in the goal). A non-level 0 goal is given in the third example. Here the unification fails (hence the goal succeeds) because $x$ is bound in the scope of where $M$ is bound. It is similar to the unification problem $\lambda x.x = \lambda x.M$. Here substitution must be capture-avoiding, therefore $M$ cannot be instantiated with $x$. However, if we switch the order of quantifier or using application-term (as in $(fx)$ in the fourth goal) the unification succeeds. In the last goal, we are trying to prove implicational goal with logic variables, and the system returns an exception.

## 5   Comparison with λProlog

Setting aside the $\nabla$ quantifier, one might think that the proof search behavior for $\forall$ and $\supset$ connectives in $FO\lambda^{\Delta\nabla}$ can be approximated in λProlog with negation-as-failure. As we outline below, only in some weak settings can λProlog naturally capture the deduction intended in $FO\lambda^{\Delta\nabla}$.

The $\supset$ connective, for instance, might be defined in λProlog as

```
imp A B :- not(A, not(B)).
```

If proof search for `A` terminates with failure, then the goal `imp A B` succeeds. Otherwise, for each answer substitution for `A`, if `B` fails then the whole goal fail, otherwise the `not(B)` fails and hence `imp A B` succeeds. For ground terms `A` and `B` (thus, containing no eigenvariables), this coincides with the operational reading of `A => B` in Level 0/1 prover. The story is not so simple, however, if there are occurrences of eigenvariables in `A` or `B`.

One can sort of see intuitively why the inclusion of eigenvariables in `A` or `B` would cause problem: the eigenvariables in λProlog play a single role as scoped



```
?- nabla x\ x = (M x).
Yes
M = x1\x1
Find another? [y/n] y
No.
?- pi x\ x = (M x).
Yes
M = x1\x1
Find another? [y/n] y
No.
?- pi M\ nabla x\ x = M => false.
Yes
Find another? [y/n] y
No.
?- pi f\ nabla x\ x = f x => print "unification succeeded".
unification succeeded
Yes
?- nabla x\ pi y\ x = y => print "unification succeeded".
unification succeeded
Yes
?-  nabla x\ x = (M x) => false.
Error: non-pure term found in implicational goal.
```

**Fig. 2.** A session in Level 0/1 prover.

constant, while in Level 0/1 they have dual roles, as constants and as variables to be instantiated. However, there is one trick to deal with this, that is, suppose we are to prove $\forall x.Ax \supset Bx$, instead of the straightforward encoding of $\forall$ as pi, we may use sigma instead:

$$\texttt{sigma x\textbackslash not (A x, not (B x)).}$$

Here the execution of the goal forces the instantiation of the (supposed to be) 'eigenvariable'. The real problem appears when eigenvariables may assume two roles at the same time. Consider the goal

$$\forall x \forall y.x = a \supset y = b$$

where $a$ and $b$ are constants. Assuming nothing about the domain of quantification, this goal is not provable. Now, the possible encodings into $\lambda$Prolog is to use either sigma or pi to encode the quantifier. Using the former, we get

$$\texttt{sigma x\textbackslash sigma y\textbackslash not (x = a, not(y = b)).}$$

This goal is provable, hence it is not the right encoding. If instead we use pi to encode $\forall$, we get

$$\texttt{pi x\textbackslash pi y\textbackslash not (x = a, not (y = b)).}$$



This goal also succeeds, since x here will become an eigenvariable and hence it is not unifiable with a. Of course, one cannot rule out other more complicated encodings, e.g., treating ∀ as pi in one place and as sigma in others, but it is doubtful that there will be an encoding scheme which can be generalized to arbitrary cases.

## 6    Example: the $\pi$-calculus and bisimulation

An implementation of one-step transitions and strong bisimulation for the $\pi$-calculus [MPW92] are given in this section. More details on the adequacy of the encodings presented in this section can be found in [TM04,Tiu04]. We consider only finite $\pi$-calculus, that is, the fragment of $\pi$-calculus without recursion or replication. The syntax of processes is defined as follows

$$\mathtt{P} ::= 0 \ \mid \ \bar{x}y.\mathtt{P} \ \mid \ x(y).\mathtt{P} \ \mid \ \tau.\mathtt{P} \ \mid \ (x)\mathtt{P} \ \mid \ [x=y]\mathtt{P} \ \mid \ \mathtt{P|Q} \ \mid \ \mathtt{P+Q}.$$

We use the notation P, Q, R, S and T to denote processes. Names are denoted by lower case letters, e.g., $a, b, c, d, x, y, z$. The occurrence of $y$ in the process $x(y).\mathtt{P}$ and $(y)\mathtt{P}$ is a binding occurrence, with P as its scope. The set of free names in P is denoted by fn(P), the set of bound names is denoted by bn(P). We write n(P) for the set fn(P) ∪ bn(P). We consider processes to be syntactically equivalent up to renaming of bound names. The operator $+$ denotes the choice operator: a process $P + Q$ can behave either like $P$ or $Q$. The operator | denotes parallel composition: the process $P|Q$ consists of subprocesses $P$ and $Q$ running in parallel. The process $[x = y]P$ behaves like $P$ if $x$ is equal to $y$. The process $x(y).P$ can input a name through $x$, which is then bound to $y$. The process $\bar{x}y.P$ can output the name $y$ through the channel $x$. Communication takes place between two processes running in parallel through the exchanges of messages (names) on the same channel (another name). The restriction operator (), e.g., in $(x)P$, restricts the scope of the name $x$ to $P$.

One-step transition in the $\pi$-calculus is denoted by $\mathtt{P} \overset{\alpha}{\longrightarrow} \mathtt{Q}$, where P and Q are processes and $\alpha$ is an action. The kinds of actions are *the silent action $\tau$, the free input action $xy$, the free output action $\bar{x}y$, the bound input action $x(y)$* and *the bound output action $\bar{x}(y)$*. Since we are working with the *late* transition semantics [MPW92], we shall not be concerned with the free input action. The name $y$ in $x(y)$ and $\bar{x}(y)$ is a binding occurrence. Just like we did with processes, we use fn($\alpha$), bn($\alpha$) and n($\alpha$) to denote free names, bound names, and names in $\alpha$. An action with a binding occurrences of a name is a *bound action*, otherwise it is a *free action*.

We encode the syntax of process expressions using $\lambda$-tree syntax as follows. We shall require three primitive syntactic categories: *n* for names, *p* for processes, and *a* for actions, and the constructors corresponding to the operators in $\pi$-calculus. The translation from $\pi$-calculus processes and transition judgments to $\lambda$-tree syntax is given in Figure 3. Figure 4 shows some example processes in $\lambda$-tree syntax. The definition clauses corresponding to the operational semantics of



$$
\begin{array}{lll}
\text{z} : p & \text{in} : n \to (n \to p) \to p & \text{out, match} : n \to n \to p \to p \\
\text{plus} : p \to p \to p & \text{par} : p \to p \to p & \text{taup} : p \to p \\
\text{nu} : (n \to p) \to p & \text{tau} : a & \text{up} : n \to n \to a \\
\text{dn} : n \to n \to a & \text{one} : p \to a \to p \to o & \text{onep} : p \to (n \to a) \to (n \to p) \to o
\end{array}
$$

$$
\begin{array}{ll}
[\![0]\!] = \text{z} & [\![x = y]\mathbf{P}]\!] = \text{match x y } [\![\mathbf{P}]\!] \\
[\![\bar{x}y.\mathbf{P}]\!] = \text{out x y } [\![\mathbf{P}]\!] & [\![x(y).\mathbf{P}]\!] = \text{in x } \lambda y.[\![\mathbf{P}]\!] \\
[\![\mathbf{P} + \mathbf{Q}]\!] = \text{plus } [\![\mathbf{P}]\!]\ [\![\mathbf{Q}]\!] & [\![\mathbf{P}|\mathbf{Q}]\!] = \text{par } [\![\mathbf{P}]\!]\ [\![\mathbf{Q}]\!] \\
[\![\tau.\mathbf{P}]\!] = \text{taup } [\![\mathbf{P}]\!] & [\![(x)\mathbf{P}]\!] = \text{nu } \lambda x.[\![\mathbf{P}]\!] \\
[\![\mathbf{P} \xrightarrow{\tau} \mathbf{Q}]\!] = \text{one } [\![\mathbf{P}]\!]\ \text{tau } [\![\mathbf{Q}]\!] & [\![\mathbf{P} \xrightarrow{\bar{x}y} \mathbf{Q}]\!] = \text{one } [\![\mathbf{P}]\!]\ (\text{up x y}) \ [\![\mathbf{Q}]\!] \\
[\![\mathbf{P} \xrightarrow{x(y)} \mathbf{Q}]\!] = \text{onep } [\![\mathbf{P}]\!]\ (\text{dn x}) \ (\lambda y [\![\mathbf{Q}]\!]) & [\![\mathbf{P} \xrightarrow{\bar{x}(y)} \mathbf{Q}]\!] = \text{onep } [\![\mathbf{P}]\!]\ (\text{up x}) \ (\lambda y [\![\mathbf{Q}]\!])
\end{array}
$$

**Fig. 3.** Encoding the $\pi$-calculus syntax with $\lambda$-tree syntax.

```
example 0 (nu x\ match x a (taup z)).
example 1 (par (in x y\z) (out x a z)).
example 2 (in x u\ (plus (taup (taup z)) (taup z))).
example 3 (in x u\ (plus (taup (taup z))
          (plus (taup z) (taup (match u y (taup z)))))).
example 4 (taup z).
example 5 (nu x\ (par (in x y\z) (out x a z))).
example 6 (in x u\ nu y\ ((plus (taup (taup z))
           (plus (taup z) (taup (match u y (taup z))))))).
```

**Fig. 4.** Several examples processes written in Level-0/1 syntax.

$\pi$-calculus are given in Figure 5. The original specification of the late semantics of $\pi$-calculus can be found in [MPW92]. We note that various side conditions on names and their scopes in the inference rules in the original specification are not present in the encoding in Figure 5 since these are handled directly by the use of $\lambda$-tree syntax and the $FO\lambda^{\Delta\nabla}$ logic.

We consider some simple examples involving one-step transitions, using the example processes in Figure 4. We can, for instance, check whether a process is stuck, i.e., no transition is possible from the given process. Consider example 0 in Figure 4 which corresponds to the process $(x)[x = a]\tau.0$. This process clearly cannot make any transition since the name $x$ has to be distinct with respect to the free names in the process. This is specified as follows

```
?- example 0 P, (pi A\pi Q\ one  P A Q => false),
                (pi A\pi Q\ onep P A Q => false).
Yes
```

Recall that we distinguish between bound-action transition and free-action transition, and hence there are two kinds of transitions to be verified.



```
onep (in X M) (dn X) M.                    % bound input
one  (out X Y P) (up X Y) P.               % free output
one  (taup P) tau P.                       % tau
one  (match X X P) A Q := one P  A Q.      % match prefix
onep (match X X P) A M := onep P A M.
one  (plus P Q) A R := one P A R.          % sum
one  (plus P Q) A R := one Q A R.
onep (plus P Q) A M := onep P A M.
onep (plus P Q) A M := onep Q A M.
one  (par P Q) A (par P1 Q) := one P A P1.  % par
one  (par P Q) A (par P Q1) := one Q A Q1.
onep (par P Q) A (x\par (M x) Q) := onep P A M.
onep (par P Q) A (x\par (N x)) := onep Q A N.
% restriction
one  (nu x\P x) A (nu x\Q x) :=  nabla x\ one (P x) A (Q x).
onep (nu x\P x) A (y\ nu x\Q x y) := nabla x\ onep (P x) A (y\ Q x y).
% open
onep (nu y\M y) (up X) N := nabla y\ one (M y) (up X y) (N y).
% close
one (par P Q) tau (nu y\ par (M y) (N y)) :=
   sigma X\ onep P (dn X) M & onep Q (up X) N.
one (par P Q) tau (nu y\ par (M y) (N y)) :=
   sigma X\ onep P (up X) M & onep Q (dn X) N.
% comm
one (par P Q) tau (par R T) := sigma X\ sigma Y\ sigma M\
   onep P (dn X) M & one Q (up X Y) T & (R = (M Y)).
one (par P Q) tau (par R T) := sigma X\ sigma Y\ sigma M\
   onep Q (dn X) M & one P (up X Y) R & (T = (M Y)).
```

**Fig. 5.** Definition of one-step transitions of finite late $\pi$-calculus

```
bisim P Q :=
   (pi A\ pi P1\ one P A P1 => sigma Q1\ one Q A Q1 & bisim P1 Q1) &
   (pi X\ pi M\  onep P (dn X) M => sigma N\ onep Q (dn X) N &
                                     pi w\ bisim (M w) (N w)) &
   (pi X\ pi M\  onep P (up X) M => sigma N\ onep Q (up X) N &
                                     nabla w\ bisim (M w) (N w)) &
   (pi A\ pi Q1\ one Q A Q1 => sigma P1\ one P A P1 & bisim Q1 P1) &
   (pi X\ pi N\  onep Q (dn X) N => sigma M\ onep P (dn X) M &
                                     pi w\ bisim (N w) (M w)) &
   (pi X\ pi N\  onep Q (up X) N => sigma M\ onep P (up X) M &
                                     nabla w\ bisim (N w) (M w)).
```

**Fig. 6.** Definition of open bisimulation



We now consider a notion of equivalence between processes, called *bisimulation*. It is formally defined as follows: a relation $\mathcal{R}$ is a bisimulation, if it is a symmetric relation such that for every $(\mathtt{P},\mathtt{Q}) \in \mathcal{R}$,

1. if $\mathtt{P} \xrightarrow{\alpha} \mathtt{P}'$ and $\alpha$ is a free action, then there is $\mathtt{Q}'$ such that $\mathtt{Q} \xrightarrow{\alpha} \mathtt{Q}'$ and $(\mathtt{P}',\mathtt{Q}') \in \mathcal{R}$,

2. if $\mathtt{P} \xrightarrow{x(z)} \mathtt{P}'$ and $z \notin \mathrm{n}(\mathtt{P},\mathtt{Q})$ then there is $\mathtt{Q}'$ such that $\mathtt{Q} \xrightarrow{x(z)} \mathtt{Q}'$ and for every name $y$, $(\mathtt{P}'[y/z],\mathtt{Q}'[y/z]) \in \mathcal{R}$,

3. if $\mathtt{P} \xrightarrow{\bar{x}(z)} \mathtt{P}'$ and $z \notin \mathrm{n}(\mathtt{P},\mathtt{Q})$ then there is $\mathtt{Q}'$ such that $\mathtt{Q} \xrightarrow{\bar{x}(z)} \mathtt{Q}'$ and $(\mathtt{P}',\mathtt{Q}') \in \mathcal{R}$.

Two processes $\mathtt{P}$ and $\mathtt{Q}$ are *strongly bisimilar* if there is a bisimulation $\mathcal{R}$ such that $(\mathtt{P},\mathtt{Q}) \in \mathcal{R}$. The above definition is also called *late bisimulation* in the literature.

Consider the definition of the $\mathtt{bisim}$ predicate Figure 6 that is inspired by the above definition. Notice that the difference between bound-input and bound-output actions is captured by the use of $\forall$ and $\nabla$ quantifiers. This definition provides a sound encoding of late bisimulation, meaning that if bisim $P\ Q$ is provable then $P$ and $Q$ are late-bisimilar. This encoding turns out to sound and complete for *open bisimulation* [San96], a finer bisimulation relation than late bisimulation (see [TM04] for details of the encoding and adequacy results). The following example, taken from [San96], illustrates the incompleteness with respect to late bisimulation.

$$P = x(u).(\tau.\tau.0 + \tau.0), \qquad Q = x(u).(\tau.\tau.0 + \tau.0 + \tau.[u = y]\tau.0).$$

This example fails because to prove their bisimilarity, one needs to do case analysis on the input name $u$ above, i.e., whether it is equal to $y$ or not, and since our current prover implements intuitionistic logic, this case split based on the excluded middle is not available. However, if we restrict the scope of $y$ so that it appears inside the scope of $u$, then $[u = y]$ is trivially false. In this case, the processes would be $x(u).(\tau.\tau.0 + \tau.0)$ and $x(u).(y)(\tau.\tau.0 + \tau.0 + \tau.[u = y]\tau.0)$, which correspond to example 3 and 6 in Figure 4. They can be proved bisimilar.

```
?- example 2 P, example 6 Q, bisim P Q.
Yes
```

One should compare the above declarative specification and its implementation of symbolic bisimulation checking with that found in, say, [BN96].

## 7 Example: modal logics for π-calculus

We now consider the modal logics for π-calculus introduced in [MPW93]. In order not to confuse meta-level ($FO\lambda^{\Delta\nabla}$) formulas (or connectives) with the formulas (connectives) of modal logics under consideration, we shall refer to the latter as object formulas (respectively, object connectives). We shall work only with object formulas which are in negation normal form, i.e., negation appears only



top : $o'$,      bot : $o'$,         and : $o' \to o' \to o'$,     or : $o' \to o' \to o'$
boxMatch : $n \to n \to o' \to o'$,    diaMatch : $n \to n \to o' \to o'$,
boxAct : $a \to o' \to o'$,          diaAct : $a \to o' \to o'$,
boxInL : $n \to (n \to o') \to o'$,    diaInL : $n \to (n \to o') \to o'$
boxOut : $n \to (n \to o') \to o'$,    diaOut : $n \to (n \to o') \to o'$
sat : $p \to o' \to o$.

$\llbracket \text{true} \rrbracket = \text{top}$   $\llbracket \text{false} \rrbracket = \text{bot}$
$\llbracket \mathtt{A} \wedge \mathtt{B} \rrbracket = \text{and } \llbracket \mathtt{A} \rrbracket \ \llbracket \mathtt{B} \rrbracket$   $\llbracket \mathtt{A} \vee \mathtt{B} \rrbracket = \text{or } \llbracket \mathtt{A} \rrbracket \ \llbracket \mathtt{B} \rrbracket$
$\llbracket [x = y]\mathtt{A} \rrbracket = \text{boxMatch x y } \llbracket \mathtt{A} \rrbracket$   $\llbracket \langle x = y\rangle\mathtt{A} \rrbracket = \text{diaMatch x y } \llbracket \mathtt{A} \rrbracket$
$\llbracket \langle \alpha \rangle \mathtt{A} \rrbracket = \text{diaAct } \alpha \ \llbracket \mathtt{A} \rrbracket$   $\llbracket [\alpha]\mathtt{A} \rrbracket = \text{boxAct } \alpha \ \llbracket \mathtt{A} \rrbracket$
$\llbracket \langle x(y)\rangle^L \mathtt{A} \rrbracket = \text{diaInL x } (\lambda y\llbracket \mathtt{A} \rrbracket)$   $\llbracket [x(y)]^L \mathtt{A} \rrbracket = \text{boxInL x } (\lambda y\llbracket \mathtt{A} \rrbracket)$
$\llbracket \langle \bar{x}(y)\rangle\mathtt{A} \rrbracket = \text{diaOut x } (\lambda y\llbracket \mathtt{A} \rrbracket)$   $\llbracket [\bar{x}(y)]\mathtt{A} \rrbracket = \text{boxOut x } (\lambda y\llbracket \mathtt{A} \rrbracket)$
$\llbracket \mathtt{P} \models \mathtt{A} \rrbracket = \text{sat } \llbracket \mathtt{P} \rrbracket \ \llbracket \mathtt{A} \rrbracket$

**Fig. 7.** Translation from modal formula to $\lambda$-tree syntax.

at the level of atomic object formulas. As a consequence, we introduce explicitly each dual pair of the object connectives. Note that since the only atomic object formulas are either true or false, we will not need negation as a connective (since $\neg\text{true} \equiv \text{false}$ and $\neg\text{false} \equiv \text{true}$). The syntax of the object formulas is given by

$$\mathtt{A} ::= \text{true} \mid \text{false} \mid \mathtt{A} \wedge \mathtt{A} \mid \mathtt{A} \vee \mathtt{A} \mid [x = z]\mathtt{A} \mid \langle x = z\rangle\mathtt{A}$$
$$\mid \langle \alpha \rangle\mathtt{A} \mid [\alpha]\mathtt{A} \mid \langle \bar{x}(y)\rangle\mathtt{A} \mid [\bar{x}(y)]\mathtt{A} \mid \langle x(y)\rangle^L\mathtt{A} \mid [x(y)]^L\mathtt{A}$$

Here, $\alpha$ denotes a free action, i.e., it is either $\tau$ or $\bar{x}y$. The modalities $[x(y)]^L$ and $\langle x(y)\rangle^L$ are the *late bound-input* modalities, and $\langle \bar{x}(y)\rangle$ and $[\bar{x}(y)]$ are the bound output modalities. There are other variants of input and output modalities considered in [MPW93] which we do not represent here. For the complete encoding of the modal logics, we refer the interested readers to [Tiu05]. In each of the formulas (and their dual 'boxed'-formulas) $\langle \bar{x}(y)\rangle\mathtt{A}$ and $\langle x(y)\rangle^L\mathtt{A}$, the occurrence of $y$ in parentheses is a binding occurrence whose scope is $\mathtt{A}$. Object formulas are considered equivalent up to renaming of bound variables. We shall be concerned with checking whether a process $\mathtt{P}$ satisfies a given modal formula $\mathtt{A}$. This satisfiability judgment is written as $\mathtt{P} \models \mathtt{A}$. The translation from modal formulas and judgments to $\lambda$-tree syntax is given in Figure 7.

The satisfiability relation for the modal logic is encoded as the definition clauses in Figure 8. For the original specification, we refer the interested readers to [MPW93]. The definition in Figure 8 is not complete, in the sense that there are true assertion of the modal logic which are not provable using this definition alone. For instance, the modal judgment

$$x(y).x(z).0 \models \langle x(y)\rangle^L\langle x(z)\rangle^L(\langle x = z\rangle\text{true} \vee [x = z]\text{false})$$

which basically says that two names are either equal or not equal, is valid, but its encoding in $FO\lambda^{\Delta\nabla}$ is not provable since the meta logic is intuitionistic. A complete encoding of the modal logic is given in [Tiu05] by explicitly introducing axioms for the excluded-middle on name equality, namely, $\forall x \forall y [x = y \vee x \neq y]$.



```
sat P top.
sat P (and A B) := sat P A, sat P B.
sat P (or  A B) := sat P A; sat P B.
sat P (boxMatch X Y A) := (X = Y) => sat P A.
sat P (diaMatch X Y A) := (X = Y), sat P A.
sat P (boxAct X A) := pi P1\ one P X P1 => sat P1 A.
sat P (diaAct X A) := sigma P1\ one P X P1, sat P1 A.
sat P (boxOut X A) := pi Q\ onep P (up X) Q => nabla y\ sat (Q y) (A y).
sat P (diaOut X A) := sigma Q\ onep P (up X) Q, nabla y\ sat (Q y)(A y).
sat P (boxInL X A) := pi Q\ onep P (dn X) Q => sigma y\ sat (Q y) (A y).
sat P (diaInL X A) := sigma Q\ onep P (dn X) Q, pi y\ sat (Q y) (A y).
```

**Fig. 8.** Specification of a modal logic for $\pi$-calculus.

The definition in Figure 8 serves also as a model checker for $\pi$-calculus. For instance, consider the processes 2 and 6 given by in Figure 4. We have seen that the two processes are bisimilar. A characterization theorem given in [MPW93] states that (late) bisimilar processes satisfy the same set of modal formulas. We consider a particular case here. The modal formula

$$\langle x(y) \rangle^L (\langle \tau \rangle \langle \tau \rangle \text{true} \vee \langle \tau \rangle \text{true})$$

naturally corresponds to the process 2. In the concrete syntax, this formula is written as follows

```
assert (diaInL x (y\ or (diaAct tau (diaAct tau top))
        (diaAct tau top))).
```

We show that both processes 2 and 6 satisfy this formula.

```
?- assert A, example 2 P, example 6 Q, sat P A, sat Q A.
Yes
```

## 8 Components of proof search implementation

Implementation of proof search for $FO\lambda^{\Delta\nabla}$ is based on a few simple key components: $\lambda$-tree syntax, i.e., data structures for representing objects containing binding, higher-order pattern unification, and stream-based computation. The first two are implemented using the *suspension calculus* [NW98], an explicit substitution notation that allows computations over $\lambda$-terms to be realized flexibly and efficiently; further details of the implementation used may be found in [NL05]. We explain the last component briefly. We use streams to store answer substitutions, which are computed lazily, i.e., only when they are queried. The data type for stream in the ML language is shown in Figure 9. Here the type `ustream` is a polymorphic stream. The element of a stream is represented as the data type `cell`, which can be a *delayed cell* or a *forced cell*. A delayed cell stores



an unevaluated expression, and its evaluation is triggered by the call to the function `getcell`. A forced cell is an element which is already a value. Elements of a stream are initially created as delayed cells. Note that since an element of a stream can also be a (cell of) stream, we can encode different computation paths using streams of streams. This feature is used, in a particular case, to encode the notion of backtracking in logic programming.

```
datatype 'a cell = delayedcell of unit -> 'a | forcedcell of 'a
type      'a elm = 'a cell
datatype 'a ustream = empty | ustream of 'a * ('a ustream elm ref)
fun getcell(t as ref(delayedcell t')) =
        let val v = t'() in (t := (forcedcell v); v) end
  | getcell(ref (forcedcell v)) = v
fun mkcell t = ref(delayedcell t)
```

**Fig. 9.** The stream datatype in ML.

A stream of substitutions for a given goal stores all answer substitutions for the goal. In logic programming, such answer substitutions can be queried one by one by users. Often we are interested in properties that hold for *all* answer substitutions. For instance, in bisimulation checking for transition systems, as we have seen in the $\pi$-calculus example, one needs to enumerate all possible successors of a process and check bisimilarity for each successor. In some other examples, information on failed proof search attempts could be of interest as well, e.g., generating counter-model in model checking. This motivates the choice of implementation architecture for $FO\lambda^{\Delta\nabla}$: various fragments of $FO\lambda^{\Delta\nabla}$ are implemented as (specialized) automated provers which interact with one another. For the current implementation, interaction between provers are restricted to exchanging streams of answer substitutions. A particular arrangement of the interaction between provers that we found quite useful is what we call a $\forall\exists$-interaction. In its simplest form, this consists of two provers, as exemplified in the Level-0/1 prover. Recall that in Level-0/1 prover, a proof search session consists of Level-1 calling the Level-0 prover, extracting all answer substitutions, and for each answer substitutions, repeating the calling cycle until the goals are proved. At the implementation level, one can generalize the provers beyond two levels using the same implementation architecture. For instance, one can imagine implementing a "Level-2 prover" which extracts answers from a Level-1 prover and perform some computations on them. Using the example of $\pi$-calculus, a Level-2 prover would, for instance, allow for proving goals like "$P$ and $Q$ are not bisimilar". This would be implemented by simply calling Level-1 on this goal and declare a success if Level-1 fails.



## 9  Future work

The current prover implements a fairly restricted fragment of the logic $FO\lambda^{\Delta\nabla}$.
We consider extending it to richer fragments to include features like, among
others, induction and co-induction proof rules (see, e.g.,[Tiu04]) and arbitrary
stratified definition (i.e., to allow more nesting of implications in goals). Of
course, with induction and co-induction proofs, there is in general no complete
automated proof search. We are considering implementing a *circular proof* search
to automatically generates the (co)inductive invariants. Works along this line has
been studied in, e.g., [SD03]. This extended feature would allow us, for example,
to reason about bisimulation of non-terminating processes. Another possible
extension is inspired by an on going work on giving a game semantics for proof
search, based on the duality of success and failure in proof search. Our particular
proof search strategy for Level-0/1 prover turns out to correspond to certain $\forall\exists$-
and $\exists\forall$-strategies in the game semantics in [MS05]. The game semantics studied
there also applies to richer fragments of logics. It would be interesting to see if
these richer fragments can be implemented as well using a similar architecture
as in Level-0/1 prover.

We also plan to use more advance techniques to improve the current im-
plementation such as using tabling to store and reuse subproofs. The use of
tabled deduction in higher-order logic programming has been studied in [Pie03].
It seems that the techniques studied there are applicable to our implementation,
to the Level-0 prover at least, since it is a subset of $\lambda$Prolog. Another possi-
ble extension would be a more flexible restriction on the occurrence of logic
variables. The current prover cannot yet handle the case where there is a case
analysis involving both eigenvariables and logic variables. Study on a notion of
higher-order pattern disunification [MP03] would be needed to attack this prob-
lem at a general level. However, we are still exploring examples and applications
which would justify this additional complication to proof search. We also plan
to study more examples on encoding process calculi and the related notions of
bisimulations.

*Acknowledgements.* Support has been obtained for this work from the following
sources: from INRIA through the "Equipes Associées" Slimmer, from the ACI
grants GEOCAL and Rossignol and from the NSF Grant CCR-0429572 that also
includes support for Slimmer.

## References

[BN96]    Michele Boreale and Rocco De Nicola.  A symbolic semantics for the $\pi$-
          calculus. *Information and Computation*, 126(1):34–52, April 1996.
[Chu40]   Alonzo Church. A formulation of the simple theory of types. *J. of Symbolic
          Logic*, 5:56–68, 1940.
[Eri91]   Lars-Henrik Eriksson.  A finitary version of the calculus of partial induc-
          tive definitions.  In L.-H. Eriksson, L. Hallnäs, and P. Schroeder-Heister,
          editors, *Proc. of the Second International Workshop on Extensions to Logic
          Programming,* volume 596 of *LNAI,* pages 89–134. Springer-Verlag, 1991.



$$\frac{}{\Sigma;\ \sigma \triangleright B, \Gamma \vdash \sigma \triangleright B}\ init \qquad \frac{\Sigma;\ \Delta \vdash \mathcal{B} \quad \Sigma;\ \mathcal{B}, \Gamma \vdash \mathcal{C}}{\Sigma;\ \Delta, \Gamma \vdash \mathcal{C}}\ cut$$

$$\frac{\Sigma;\ \sigma \triangleright B, \sigma \triangleright C, \Gamma \vdash \mathcal{D}}{\Sigma;\ \sigma \triangleright B \wedge C, \Gamma \vdash \mathcal{D}}\ \wedge\mathcal{L} \qquad \frac{\Sigma;\ \Gamma \vdash \sigma \triangleright B \quad \Sigma;\ \Gamma \vdash \sigma \triangleright C}{\Sigma;\ \Gamma \vdash \sigma \triangleright B \wedge C}\ \wedge\mathcal{R}$$

$$\frac{\Sigma;\ \sigma \triangleright B, \Gamma \vdash \mathcal{D} \quad \Sigma;\ \sigma \triangleright C, \Gamma \vdash \mathcal{D}}{\Sigma;\ \sigma \triangleright B \vee C, \Gamma \vdash \mathcal{D}}\ \vee\mathcal{L} \qquad \frac{\Sigma;\ \Gamma \vdash \sigma \triangleright B}{\Sigma;\ \Gamma \vdash \sigma \triangleright B \vee C}\ \vee\mathcal{R}$$

$$\frac{}{\Sigma;\ \sigma \triangleright \bot, \Gamma \vdash \mathcal{B}}\ \bot\mathcal{L} \qquad \frac{\Sigma;\ \Gamma \vdash \sigma \triangleright C}{\Sigma;\ \Gamma \vdash \sigma \triangleright B \vee C}\ \vee\mathcal{R}$$

$$\frac{\Sigma;\ \Gamma \vdash \sigma \triangleright B \quad \Sigma;\ \sigma \triangleright C, \Gamma \vdash \mathcal{D}}{\Sigma;\ \sigma \triangleright B \supset C, \Gamma \vdash \mathcal{D}}\ \supset\mathcal{L} \qquad \frac{\Sigma;\ \sigma \triangleright B, \Gamma \vdash \sigma \triangleright C}{\Sigma;\ \Gamma \vdash \sigma \triangleright B \supset C}\ \supset\mathcal{R}$$

$$\frac{\Sigma, \sigma \vdash t : \gamma \quad \Sigma;\ \sigma \triangleright B[t/x], \Gamma \vdash \mathcal{C}}{\Sigma;\ \sigma \triangleright \forall_\gamma x.B, \Gamma \vdash \mathcal{C}}\ \forall\mathcal{L} \qquad \frac{\Sigma, h;\ \Gamma \vdash \sigma \triangleright B[(h\ \sigma)/x]}{\Sigma;\ \Gamma \vdash \sigma \triangleright \forall x.B}\ \forall\mathcal{R}$$

$$\frac{\Sigma, h;\ \sigma \triangleright B[(h\ \sigma)/x], \Gamma \vdash \mathcal{C}}{\Sigma;\ \sigma \triangleright \exists x.B, \Gamma \vdash \mathcal{C}}\ \exists\mathcal{L} \qquad \frac{\Sigma, \sigma \vdash t : \gamma \quad \Sigma;\ \Gamma \vdash \sigma \triangleright B[t/x]}{\Sigma;\ \Gamma \vdash \sigma \triangleright \exists_\gamma x.B}\ \exists\mathcal{R}$$

$$\frac{\Sigma;\ (\sigma, y) \triangleright B[y/x], \Gamma \vdash \mathcal{C}}{\Sigma;\ \sigma \triangleright \nabla x\ B, \Gamma \vdash \mathcal{C}}\ \nabla\mathcal{L} \qquad \frac{\Sigma;\ \Gamma \vdash (\sigma, y) \triangleright B[y/x]}{\Sigma;\ \Gamma \vdash \sigma \triangleright \nabla x\ B}\ \nabla\mathcal{R}$$

$$\frac{\Sigma;\ \mathcal{B}, \mathcal{B}, \Gamma \vdash \mathcal{C}}{\Sigma;\ \mathcal{B}, \Gamma \vdash \mathcal{C}}\ c\mathcal{L} \qquad \frac{\Sigma;\ \Gamma \vdash \mathcal{C}}{\Sigma;\ \mathcal{B}, \Gamma \vdash \mathcal{C}}\ w\mathcal{L} \qquad \frac{}{\Sigma;\ \Gamma \vdash \sigma \triangleright \top}\ \top\mathcal{R}$$

**Fig. 10.** The core rules of $FO\lambda^{\Delta\nabla}$.


[Gen69]   Gerhard Gentzen. Investigations into logical deductions. In M. E. Szabo, editor, *The Collected Papers of Gerhard Gentzen*, pages 68–131. North-Holland Publishing Co., Amsterdam, 1969.

[Gir92]   Jean-Yves Girard. A fixpoint theorem in linear logic. Email to the linear@cs.stanford.edu mailing list, February 1992.

[HSH91]   Lars Hallnäs and Peter Schroeder-Heister. A proof-theoretic approach to logic programming. II. Programs as definitions. *Journal of Logic and Computation*, 1(5):635–660, October 1991.

[Hue75]   Gérard Huet. A unification algorithm for typed λ-calculus. *Theoretical Computer Science*, 1:27–57, 1975.

[LC89]    Pierre Lescanne and Hubert Comon. Equational problems and disunification. *Journal of Symbolic Computation*, 3 and 4:371–426, 1989.

[Mil91]   Dale Miller. A logic programming language with lambda-abstraction, function variables, and simple unification. *Journal of Logic and Computation*, 1(4):497–536, 1991.

[Mil92]   Dale Miller. Unification under a mixed prefix. *J. of Symbolic Computation*, 14(4):321–358, 1992.

[MM00]    Raymond McDowell and Dale Miller. Cut-elimination for a logic with definitions and induction. *Theoretical Computer Science*, 232:91–119, 2000.

[MP03]    Alberto Momigliano and Frank Pfenning. Higher-order pattern complement and the strict λ-calculus. *ACM Trans. Comput. Logic*, 4(4):493–529, 2003.

[MPW92]   Robin Milner, Joachim Parrow, and David Walker. A calculus of mobile processes, Part II. *Information and Computation*, pages 41–77, 1992.





[MPW93]  Robin Milner, Joachim Parrow, and David Walker. Modal logics for mobile processes. *Theoretical Computer Science*, 114(1):149–171, 1993.

[MS05]  Dale Miller and Alexis Saurin. A game semantics for proof search: Preliminary results. In *Proceedings of the Mathematical Foundations of Programming Semantics (MFPS)*, 2005.

[MT03]  Dale Miller and Alwen Tiu. A proof theory for generic judgments: An extended abstract. In *LICS 2003*, pages 118–127. IEEE, June 2003.

[MT05]  Dale Miller and Alwen Tiu. A proof theory for generic judgments. *ACM Transactions on Computational Logic*, 6(4), October 2005.

[Nip93]  Tobias Nipkow. Functional unification of higher-order patterns. In M. Vardi, editor, *LICS93*, pages 64–74. IEEE, June 1993.

[NL05]  Gopalan Nadathur and Natalie Linnell. Practical higher-order pattern unification with on-the-fly raising. In *ICLP 2005: 21st International Logic Programming Conference*, volume 3668 of *LNCS*, pages 371–386, Sitges, Spain, October 2005. Springer.

[NW98]  Gopalan Nadathur and Debra Sue Wilson. A notation for lambda terms: A generalization of environments. *Theoretical Computer Science*, 198(1-2):49–98, 1998.

[Pie03]  Brigitte Pientka. *Tabled Higher-Order Logic Programming*. PhD thesis, Carnegie Mellon University, December 2003.

[San96]  Davide Sangiorgi. A theory of bisimulation for the π-calculus. *Acta Informatica*, 33(1):69–97, 1996.

[SD03]  Christoph Sprenger and Mads Dam. On the structure of inductive reasoning: Circular and tree-shaped proofs in the μ-calculus. In A.D. Gordon, editor, *Proceedings, Foundations of Software Science and Computational Structures (FOSSACS), Warsaw, Poland*, volume 2620 of *LNCS*, pages 425–440. Springer-Verlag, 2003.

[SH93]  Peter Schroeder-Heister. Rules of definitional reflection. In M. Vardi, editor, *Eighth Annual Symposium on Logic in Computer Science*, pages 222–232. IEEE Computer Society Press, June 1993.

[Stä94]  R. F. Stärk. Cut-property and negation as failure. *International Journal of Foundations of Computer Science*, 5(2):129–164, 1994.

[Tiu04]  Alwen Tiu. *A Logical Framework for Reasoning about Logical Specifications*. PhD thesis, Pennsylvania State University, May 2004.

[Tiu05]  Alwen Tiu. Model checking for π-calculus using proof search. In Martín Abadi and Luca de Alfaro, editors, *CONCUR*, volume 3653 of *Lecture Notes in Computer Science*, pages 36–50. Springer, 2005.

[TM04]  Alwen Tiu and Dale Miller. A proof search specification of the π-calculus. In *3rd Workshop on the Foundations of Global Ubiquitous Computing*, 2004.




# Otter-lambda

Michael Beeson[1]

San José State University, San José, Calif.
beeson@cs.sjsu.edu,
www.cs.sjsu.edu/faculty/beeson

**Abstract.** Otter-lambda is a theorem-prover based on an untyped logic with lambda calculus, called Lambda Logic. Otter-lambda is built on Otter, so it uses resolution proof search, supplemented by demodulation and paramodulation for equality reasoning, but it also uses a new algorithm, lambda unification, for instantiating variables for functions or predicates. The underlying logic of Otter-lambda is lambda logic, an untyped logic combining lambda calculus and first-order logic. The use of lambda unification allows Otter-lambda to prove some theorems usually thought of as "higher-order". There are theoretical questions about lambda logic and its relation to first-order and higher-order logic, and theoretical questions about lambda unification and its relation to higher-order unification, but the demonstration will focus on the practical capabilities of Otter-lambda. Specifically, several proofs in algebra and number theory will be discussed, with special focus on the use of Otter-lambda in connection with mathematical induction. Otter-lambda has had some successes in this area, since lambda logic can state the general induction schema (with a variable for a predicate), and lambda unification can sometimes find the appropriate instance(s) of induction for a particular problem, even when nested multiple inductions are required. Once that it is done, the full resources of Otter are available to carry out the base case and the induction step, with lambda-unification still available if another induction is needed. Some examples are carried out directly from Peano's axioms, such as the commutativity of multiplication. Some involve algebra, for example, there are no nilpotents in an integral domain. Others are carried out with the aid of external simplification by MathXpert, for example, a proof by induction on $n$ of Bernoulli's inequality $1 + nx \le (1 + x)^n$ if $x > -1$.



# Tps: A Theorem Proving System for Church's Type Theory


Chad E. Brown

Universität des Saarlandes, Saarbrücken, Germany, cebrown@ags.uni-sb.de


Tps [1, 2] is a theorem proving system providing support for automated and interactive proving in fragments of Church's Type Theory [3, 5]. Tps has been developed by Peter Andrews and several of his students at Carnegie Mellon University. Tps users can interactively construct natural deduction proofs of theorems in Church's Type Theory. Users can also ask Tps to prove theorems automatically using a variety of modes. (A mode is a collection of flag settings.) Depending on the mode, the search procedure will attempt to find a proof in elementary type theory or extensional type theory, with various restrictions on the search space [4]. Proofs found automatically by Tps are translated to natural deduction proofs. Users can also interactively construct part of a natural deduction proof, then ask Tps to automatically fill in certain gaps.

## References


1. Peter B. Andrews, Matthew Bishop, Sunil Issar, Dan Nesmith, Frank Pfenning, and Hongwei Xi. TPS: A theorem proving system for classical type theory. *Journal of Automated Reasoning*, 16:321–353, 1996.
2. Peter B. Andrews, Matthew Bishop, and Chad E. Brown. System description: TPS: A theorem proving system for type theory. In McAllester [6], pages 164–169.
3. Peter B. Andrews. *An Introduction to Mathematical Logic and Type Theory: To Truth Through Proof.* Kluwer Academic Publishers, second edition, 2002.
4. Chad E. Brown. *Set Comprehension in Church's Type Theory.* PhD thesis, Department of Mathematical Sciences, Carnegie Mellon University, 2004.
5. Alonzo Church. A formulation of the simple theory of types. *Journal of Symbolic Logic*, 5:56–68, 1940.
6. David McAllester, editor. *Proceedings of the 17th International Conference on Automated Deduction*, volume 1831 of *Lecture Notes in Artificial Intelligence*, Pittsburgh, PA, USA, 2000. Springer-Verlag.






# System Description: The Metis Proof Tactic

Joe Hurd*

Computing Laboratory
University of Oxford,
`joe.hurd@comlab.ox.ac.uk`

The Metis proof tactic for the HOL4 theorem prover [1] proves higher order logic goals using a first order proof calculus. It is implemented in Standard ML, supporting a tight integration with the rest of the HOL4 theorem prover, and is required to respond within a few seconds to be useful during interactive proof.

The steps of its operation are as follows:

1. The initial higher order logic goal is negated and converted by proof to conjunctive normal form. Definitional conjunctive normal form is used to avoid exponential blow-up (occasionally encountered in practice).
2. A suitable logical interface $(m, t)$ is selected, consisting of a mapping $m$ from higher order logic formulas to first order clauses, and also a translation $t$ that lifts first order refutations to higher order logic proofs [2]. The conjuncts are mapped to first order logic clauses.
3. A refutation of the clauses is found, where the search is performed using the ordered paramodulation calculus [3].
4. The first order refutation is lifted to a higher order logic proof of the normalized goal, completing the proof of the initial goal.

The Metis proof tactic is effective on many classes of higher order logic goal, particularly those that require a combination of deductive and equality reasoning. It automatically selects an interface for the goal, first trying a fast one that discards type information, and then a more robust one that includes type information in the first order clauses. A syntactic check detects whether the goal contains higher order features such as quantification over functions, and if so an interface is selected that maps higher order logic function application to a first order function symbol (i.e., $f(x)$ is mapped to $\mathsf{app}(f, x)$). This automatic interface selection makes Metis more efficient and also gives it a wider coverage than might be expected of a first order proof tactic. As an indication of its popularity, the string `METIS_TAC` occurs 1,822 times in the 243,636 lines of the HOL4 sources, and work is ongoing to port it to other higher order logic theorem provers.

## References

1. M. J. C. Gordon and T. F. Melham, editors. *Introduction to HOL (A theorem-proving environment for higher order logic).* Cambridge University Press, 1993.

---

* Supported by a Junior Research Fellowship at Magdalen College, Oxford.





2. Joe Hurd. An LCF-style interface between HOL and first-order logic. In Andrei Voronkov, editor, *Proceedings of the 18th International Conference on Automated Deduction (CADE-18)*, volume 2392 of *Lecture Notes in Artificial Intelligence*, pages 134–138, Copenhagen, Denmark, July 2002. Springer.
3. R. Nieuwenhuis and A. Rubio. Paramodulation-based theorem proving. In A. Robinson and A. Voronkov, editors, *Handbook of Automated Reasoning*, volume I, chapter 7, pages 371–443. Elsevier Science, 2001.